\documentclass[final]{article}
\usepackage[utf8]{inputenc}
\usepackage[english]{babel}

\usepackage{comment} 

\usepackage{amssymb}   
\usepackage{amsthm}    
\usepackage{thmtools}  
\usepackage{mathtools} 
\usepackage{mathrsfs}  
\usepackage{enumitem}  

\usepackage{graphicx}  
\usepackage{adjustbox}
\graphicspath{{figurer/}}
\usepackage{booktabs, multirow}  
\usepackage{babel}     
\usepackage{csquotes}  
\usepackage{textcomp}  
\usepackage{listings}  
\usepackage{caption}
\usepackage{subcaption}

\usepackage{accents}
\newcommand{\thickbar}{} 
\DeclareRobustCommand*\thickbar[1]{\accentset{\rule{.35em}{.65pt}}{#1}}

\usepackage{mathscinet}
\usepackage[backend    = biber,
            sortcites  = false,
            sorting    = nyt,
            giveninits = false,
            uniquename = false,
            doi        = false,
            isbn       = false,
            url        = false,
            maxcitenames = 2,
            style      = authoryear-comp]{biblatex}
\DeclareNameAlias{sortname}{family-given}
\DeclareNameAlias{default}{family-given}
\DeclareFieldFormat[article]{volume}{\bibstring{jourvol}\addnbspace#1}
\DeclareFieldFormat[article]{number}{\bibstring{number}\addnbspace#1}
\renewbibmacro*{volume+number+eid}
{
    \printfield{volume}
    \setunit{\addcomma\space}
    \printfield{number}
    \setunit{\addcomma\space}
    \printfield{eid}
}
\usepackage{appendix}

\usepackage{xargs}                      
\usepackage[pdftex,dvipsnames,table]{xcolor}  
\usepackage[colorinlistoftodos,prependcaption,textsize=tiny]{todonotes}
\newcommandx{\unsure}[2][1=]{\todo[linecolor=red,backgroundcolor=red!25,bordercolor=red,#1]{#2}}
\newcommandx{\change}[2][1=]{\todo[linecolor=blue,backgroundcolor=blue!25,bordercolor=blue,#1]{#2}}
\newcommandx{\info}[2][1=]{\todo[linecolor=OliveGreen,backgroundcolor=OliveGreen!25,bordercolor=OliveGreen,#1]{#2}}
\newcommandx{\improvement}[2][1=]{\todo[linecolor=Plum,backgroundcolor=Plum!25,bordercolor=Plum,#1]{#2}}
\newcommandx{\thiswillnotshow}[2][1=]{\todo[disable,#1]{#2}}
\definecolor{col_ind}{HTML}{E81A20}
\definecolor{col_emp}{HTML}{FF7F00}
\definecolor{col_par}{HTML}{FFDF0F}
\definecolor{col_gen}{HTML}{33A02C}
\definecolor{col_sep}{HTML}{A6CEE3}
\definecolor{col_sur}{HTML}{1F78B4}

\usepackage{varioref}
\usepackage[pdfusetitle]{hyperref}
\usepackage[nameinlink, noabbrev]{cleveref}
\Crefname{chapter}{Chapter}{Chapters}
\Crefname{section}{Section}{Sections}

\newcommand{\set}[1]{\left\lbrace #1 \right\rbrace}
\newcommand{\p}[1]{\left( #1 \right)}
\newcommand{\abs}[1]{\left| #1 \right|}%
\newcommand{\tm}[1]{\texttt{#1}}%

\newcommand{\diff}{\mathop{}\!\mathrm{d}}

\DeclareMathOperator{\Var}{Var}

\DeclarePairedDelimiterX{\infdivx}[2]{(}{)}{%
  #1\;\delimsize\|\;#2%
}

\newcommand{\N}{\mathbb{N}}   
\newcommand{\R}{\mathbb{R}}   
\newcommand{\E}{\mathbb{E}}   

\newcommand{\x}{\boldsymbol{x}}

\newcommand{\z}{\boldsymbol{z}}
\newcommand{\0}{\boldsymbol{0}}

\newcommand{\xsb}{{\boldsymbol{x}_{\thickbar{\mathcal{S}}}}}

\newcommand{\xs}{{\boldsymbol{x}_{\mathcal{S}}}}
\newcommand{\xss}{{\boldsymbol{x}_{\mathcal{S}}^*}}
\newcommand{\sbb}{{\thickbar{\mathcal{S}}}}
\newcommand{\s}{{\mathcal{S}}}
\newcommand{\M}{{\mathcal{M}}}
\newcommand{\pow}{{\mathcal{P}}}
\newcommand{\btheta}{{\boldsymbol{\theta}}}
\newcommand{\bpsi}{{\boldsymbol{\psi}}}
\newcommand{\bphi}{{\boldsymbol{\phi}}}
\newcommand{\bmu}{{\boldsymbol{\mu}}}

\newcommand{\bbeta}{{\boldsymbol{\beta}}}

\newcommand{\bSigma}{{\boldsymbol{\Sigma}}}

\newcommand{\vaeac}{$\texttt{VAEAC}$}

\newcommand{\vaeacf}{\texttt{VAEAC-f}}
\newcommand{\vaeacfdir}{\texttt{VAEAC-f-dir}}
\newcommand{\vaeacfindir}{\texttt{VAEAC-f-indir}}
\newcommand{\tu}{\textunderscore}

\newcommand{\empirical}{\texttt{empirical}}
\newcommand{\parametric}{\texttt{parametric}}
\newcommand{\generative}{\texttt{generative}}
\newcommand{\separate}{\texttt{separate regression}}
\newcommand{\surrogate}{\texttt{surrogate regression}}
\newcommand{\separatereg}{\texttt{separate regression}}
\newcommand{\surrogatereg}{\texttt{surrogate regression}}
\newcommand{\ctree}{\texttt{ctree}}

\newcommand{\Gaussian}{\texttt{Gaussian}}
\newcommand{\GH}{\texttt{GH}}
\newcommand{\Burr}{\texttt{Burr}}
\newcommand{\copula}{\texttt{copula}}
\newcommand{\independence}{{\texttt{independence}}}

\newcommand{\Rlang}{\textsc{R}}
\newcommand{\lm}{\text{lm}}
\newcommand{\gam}{\text{gam}}
\newcommand{\no}{\text{no}}
\newcommand{\some}{\text{some}}
\newcommand{\more}{\text{more}}
\newcommand{\many}{\text{many}}
\newcommand{\numerous}{\text{numerous}}
\newcommand{\one}{\text{one}}
\newcommand{\two}{\text{two}}
\newcommand{\three}{\text{three}}
\newcommand{\five}{\text{five}}
\newcommand{\all}{\text{all}}
\DeclareMathOperator*{\argmin}{arg\,min}

\usepackage{xcolor}

\usepackage[bottom]{footmisc}
\usepackage{makecell} 

\newcommand\undermat[2]{%
  \makebox[0pt][l]{$\smash{\underbrace{\phantom{%
    \begin{matrix}#2\end{matrix}}}_{\text{$#1$}}}$}#2}

\usepackage{siunitx}
\renewrobustcmd{\bfseries}{\fontseries{b}\selectfont}
\renewrobustcmd{\boldmath}{}
\newrobustcmd{\B}{\bfseries}

\usepackage[margin=1.40in]{geometry}

\usepackage{authblk}

\title{A Comparative Study of Methods for Estimating Conditional Shapley Values and When to Use Them}
\author[1]{Lars Henry Berge Olsen\thanks{E-mail address to corresponding author: \href{mailto:lholsen@math.uio.no}{\texttt{lholsen@math.uio.no}}.}}
\author[1]{Ingrid Kristine Glad}
\author[2]{\\Martin Jullum}
\author[2]{Kjersti Aas}
\affil[1]{Department of Mathematics, University of Oslo}
\affil[2]{Norwegian Computing Center}
\date{\today}

\begin{document}

\maketitle

\begin{abstract}
Shapley values originated in cooperative game theory but are extensively used today as a model-agnostic explanation framework to explain predictions made by complex machine learning models in the industry and academia. There are several algorithmic approaches for computing different versions of Shapley value explanations. Here, we focus on conditional Shapley values for predictive models fitted to tabular data. Estimating precise conditional Shapley values is difficult as they require the estimation of non-trivial conditional expectations. In this article, we develop new methods, extend earlier proposed approaches, and systematize the new refined and existing methods into different method classes for comparison and evaluation. The method classes use either Monte Carlo integration or regression to model the conditional expectations. We conduct extensive simulation studies to evaluate how precisely the different method classes estimate the conditional expectations, and thereby the conditional Shapley values, for different setups. We also apply the methods to several real-world data experiments and provide recommendations for when to use the different method classes and approaches. Roughly speaking, we recommend using parametric methods when we can specify the data distribution almost correctly, as they generally produce the most accurate Shapley value explanations. When the distribution is unknown, both generative methods and regression models with a similar form as the underlying predictive model are good and stable options. Regression-based methods are often slow to train but produce the Shapley value explanations quickly once trained. The vice versa is true for Monte Carlo-based methods, making the different methods appropriate in different practical situations.



\end{abstract}

\section{Introduction}
\label{Introduction}
Complex machine learning (ML) models are extensively applied to solve supervised learning problems in many different fields and settings; cancer prognosis \parencite{KOUROU20158}, credit scoring \parencite{KVAMME2018207}, impact sensitivity of energetic crystals \parencite{lansford2022building}, and money laundering detection \parencite{jullum2020detecting}. The ML methods are often very complex, as they contain thousands, millions, or even billions of tuneable model parameters. Thus, understanding the complete underlying decision making process of the ML algorithms are infeasible (for us humans). The use of ML methods is based on them having the potential to generate more accurate predictions than established statistical models, but this may come at the expense of model interpretability, as discussed by \textcite{johansson2011trade, Guo2019AnIM, luo2019balancing}. \textcite{rudin2019stop} conjectures that equally accurate but interpretable models exist across domains even though they might be hard to find. 

The lack of understanding of how the input features of the ML model influence the model's output is a major drawback. Hence, to remedy the absence of interpretation, the fields of explainable artificial intelligence (XAI) and interpretable machine learning (IML) have become active research fields in recent years \parencite{adadi2018peeking, molnar2019, covert2021explaining}. There has been developed a wide variety of explanation frameworks which extract the hidden knowledge about the underlying data structure captured by the black-box model, making the model's decision-making process more transparent. This is essential for, e.g., medical researchers who apply an intricate ML model to obtain well-performing predictions but who simultaneously also aim to discover important risk factors. The \textit{Right to Explanation} legislation in the European Union's General Data Protection Regulation (GDPR) has also been a driving factor \parencite{regulation2016}.



One of the most commonly used explanation frameworks in XAI is \textit{Shapely values}, which is an explanation methodology with a strong mathematical foundation and unique theoretical properties from the field of cooperative game theory \parencite{shapley1953value}. Shapley values are most commonly used as a \textit{model-agnostic} explanation framework for individual predictions, that is, for \textit{local explanations}. Model-agnostic means that Shapley values do not rely on model internals and can be used to compare and explain any ML model trained on the same supervised learning problem. Local explanation means that Shapley values explain the local model behavior for an individual observation and not the global model behavior across all data instances. The methodology has also been used to provide \textit{global explanations}, see, e.g., \textcite{owen2014sobol, covert2020understanding, frye_shapley-based_2020,  giudici2021shapley}.  See \textcite{molnar2019} for an overview and detailed introduction to other explanation frameworks.

Shapley values originated in cooperative game theory, but have been reintroduced as a framework for model explanation by \textcite{strumbelj2010efficient, strumbelj2014explaining, lundberg2017unified}. Originally, Shapley values described a possible solution concept of how to fairly allocate the payout of a game among the players based on their contribution to the overall cooperation/payout. The solution concept is based on several desirable axioms, for which the Shapley values are the unique solution. When applying Shapley values as an explanation framework, we treat the features as the ``players", the machine learning model as the ``game", and the corresponding prediction as the ``payout". 

There are several ways to define the game, which yields different types of Shapley values. For local explanations, the two main types are \textit{interventional} and \textit{conditional} Shapley values\footnote{They are also called \textit{marginal} and \textit{observational} Shapley values, respectively.}, and there is an ongoing debate about when to use them \parencite{chen2020true, kumar2020problems, chen2022algorithms}. Briefly stated, the interventional version does not take dependencies between the features into consideration, while the conditional version does. A disadvantage of the conditional Shapley values, compared to the interventional counterpart, is that they require the estimation/modeling of non-trivial conditional expectations. Throughout this article, we mean conditional Shapley values when we discuss Shapley values, if not otherwise specified.

\begin{figure}[!t]
    \centering
    \centerline{\includegraphics[width=1\textwidth]{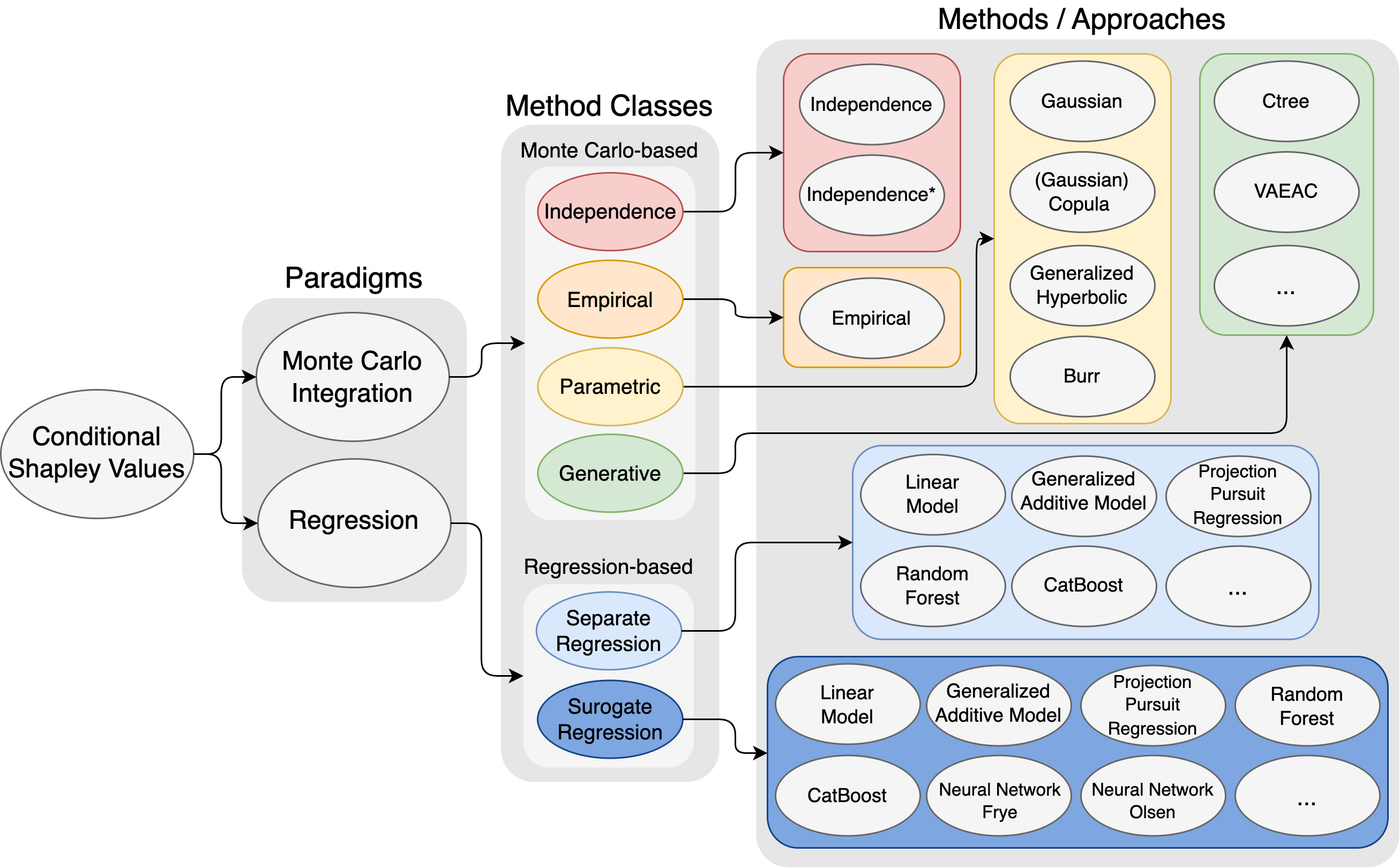}}
    \caption{{\small Schematic overview of the paradigms, method classes, and methods/approaches used in this article to compute the conditional Shapley value explanations.}}
    \label{fig:SchematicOverviewMethodsMainText}
\end{figure}

There is a vast amount of literature on different approaches for estimating Shapley values \parencite{strumbelj2009explaining,lundberg2017unified,lundberg2017consistent,redelmeier2020,williamson2020efficient,aas2019explaining,aas2021explaining,frye_shapley-based_2020,covert2021explaining,Olsen2022}. These methods can be grouped into different method classes based on their characteristics, that is, if they (implicitly) assume feature independence or use empirical estimates, parametric assumptions, generative methods, and/or regression models; see \Cref{fig:SchematicOverviewMethodsMainText}. To the best of our knowledge, there exist no thorough and methodological comparison between all the method classes and approaches. \textcite[Section 6]{chen2022algorithms} states that ``[conditional Shapley values] constitutes an important future research direction that would benefit from new methods or systematic evaluations of existing approaches".



In this article, we both investigate existing methods, introduce several new approaches, and conduct extensive simulation studies starting from a very simple set-up with an interpretable model as a sanity check and gradually increase the complexity of the predictive model. We also investigate the effect the data distribution, with varying levels of dependence and the training sample size, has on the estimation of the conditional expectations using the different methods. Finally, we also conduct experiments on real-world data sets from the UCI Machine Learning Repository. In the numerical simulation studies, the parametric methods, which correctly (or nearly correctly) assume the data distribution, generate the most accurate Shapley values. However, if the data distribution is unknown, such as for most real-world data sets, our experiments show that using either a generative method or a regression model with the same form as the predictive model is the best option. In addition to accuracy, we also investigate the computation times of the methods. Based on our findings, we present recommendations for when to use the different method classes and approaches.


In \Cref{sec:ShapleyValues}, we give an overview of Shapley values' origin and use as a model-agnostic explanation framework. The existing and novel methods for estimating the Shapley value explanations are described in \Cref{sec:ConditionalShapleyValuesApproaches}. In \Cref{sec:Numerical_Simulation_Studies}, we present the simulation studies and discuss the corresponding results. We conduct experiments on real-world data sets in \Cref{sec:real_world_data}. Recommendations for when to use the different methods and a conclusion are given in \Cref{sec:Recommendations,sec:Conclusion}, respectively. In the \nameref{Appendix}, we provide more approaches, implementation details, and additional simulation studies.

\section{Shapley Values}
\label{sec:ShapleyValues}
In this section, we first briefly describe Shapley values in cooperative game theory, before we elaborate on their use for model explanations.

\subsection{Shapley Values in Cooperative Game Theory}
\label{subsec:ShapleyValuesGameTheory}

Shapley values are a solution concept of how to divide the payout of a cooperative game $v:\mathcal{P}(\M) \mapsto \R$ based on four axioms \parencite{shapley1953value}. The game is played by $M$ players where $\M = \{1,2,\dots,M\}$ denotes the set of all players and $\mathcal{P}(\M)$ is the power set, that is, the set of all subsets of $\M$. We call $v(\s)$ the \textit{contribution function}\footnote{The $v(\s)$ is also called the \textit{reward function} and \textit{characteristic function} in the literature.} and it maps a subset of players $\s \in \pow(\M)$, also called a coalition, to a real number representing their contribution in the game $v$. The Shapley values $\phi_j = \phi_j(v)$ assigned to each player $j$, for $j = 1, \dots, M$, uniquely satisfy the following properties:
\begin{enumerate}[align=left, itemsep=2pt, topsep=6pt] 
    \item [\textbf{Efficiency}:] They sum to the value of the grand coalition $\M$ over the empty set $\emptyset$, that is, $\sum_{j=1}^M \phi_j = v(\M) - v(\emptyset)$.
    \item [\textbf{Symmetry}:] Two equally contributing players $j$ and $k$, that is,  $v(\s \cup \{j\}) = v(\s \cup \{k\})$ for all $\s$, receive equal payouts $\phi_j = \phi_k$.
    \item [\textbf{Dummy}:] A non-contributing player $j$, that is, $v(\s) = v(\s \cup \{j\})$ for all $\s$, receives $\phi_j = 0$.
    \item [\textbf{Linearity}:] A linear combination of $n$ games $\{v_1, \dots, v_n\}$, that is, $v(\s) = \sum_{k=1}^nc_kv_k(\s)$, has Shapley values given by $\phi_j(v) = \sum_{k=1}^nc_k\phi_j(v_k)$.
\end{enumerate}
    
\textcite{shapley1953value} showed that the values $\phi_j$ which satisfy these axioms are given by
\begin{align}
    \label{eq:ShapleyValuesDef}
    \phi_j = \sum_{\mathcal{S} \in \pow(\mathcal{M} \backslash \{j\})} \frac{|\mathcal{S}|!(M-|\mathcal{S}|-1)!}{M!}\left(v(\mathcal{S} \cup \{j\}) - v(S) \right),
\end{align}
where $|\mathcal{S}|$ is the number of players in coalition $\s$. The number of terms in \eqref{eq:ShapleyValuesDef} is $2^{M}$, hence, the complexity grows exponentially with the number of players $M$. Each Shapley value is a weighted average of the player’s marginal contribution to each coalition $\mathcal{S}$. 

\subsection{Shapley Values in Model Explanation}
\label{subsec:ShapleyValuesExplainability}
We consider the setting of supervised learning where we aim to explain a predictive model $f(\boldsymbol{x})$ trained on $\mathcal{X} = \{\boldsymbol{x}^{[i]}, y^{[i]}\}_{i = 1}^{N_\text{train}}$, where $\boldsymbol{x}^{[i]}$ is an $M$-dimensional feature vector, $y^{[i]}$ is a univariate response, and $N_\text{train}$ is the number of training observations. The prediction $\hat{y} = f(\boldsymbol{x})$, for a specific feature vector $\boldsymbol{x} = \boldsymbol{x}^*$, is explained using Shapley values as a model-agnostic explanation framework \parencite{strumbelj2010efficient, strumbelj2014explaining, lundberg2017unified}. The fairness aspect of Shapley values in the model explanation setting is discussed in, for example, \textcite{chen2020true, Fryer2021ShapleyVF, aas2019explaining}. 

In the Shapley value framework, the predictive model $f$ (indirectly) replaces the cooperative game and the $M$-dimensional feature vector replaces the $M$ players. The Shapley value $\phi_j$ describes the importance of the $j$th feature in the prediction $f(\boldsymbol{x}^*) = \phi_0^* + \sum_{j=1}^M\phi_j^*$, where $\phi_0 = \E \left[f(\boldsymbol{x})\right]$. That is, the sum of the Shapley values explains the difference between the prediction $f(\boldsymbol{x}^*)$ and the global average prediction. 

To calculate \eqref{eq:ShapleyValuesDef}, we need to define an appropriate contribution function $v(\mathcal{S}) = v(\mathcal{S}, \x^*)$ which should resemble the value of $f(\boldsymbol{x}^*)$ when only the features in coalition $\mathcal{S}$ are known. We use the contribution function proposed by \textcite{lundberg2017unified}, namely the expected response of $f(\boldsymbol{x})$ conditioned on the features in $\mathcal{S}$ taking on the values $\boldsymbol{x}_\mathcal{S}^*$. That is, 
\begin{align}
    \label{eq:ContributionFunc}
    \begin{split}
        v(\mathcal{S}) 
        &=
        \E\left[ f(\boldsymbol{x}) | \boldsymbol{x}_{\mathcal{S}} = \boldsymbol{x}_{\mathcal{S}}^* \right] 
        =
        \E\left[ f(\boldsymbol{x}_{\thickbar{\mathcal{S}}}, \boldsymbol{x}_{\mathcal{S}}) | \boldsymbol{x}_{\mathcal{S}} = \boldsymbol{x}_{\mathcal{S}}^* \right] 
        = 
        \int f(\boldsymbol{x}_{\thickbar{\mathcal{S}}}, \boldsymbol{x}_{\mathcal{S}}^*) p(\boldsymbol{x}_{\thickbar{\mathcal{S}}} | \boldsymbol{x}_{\mathcal{S}} = \boldsymbol{x}_{\mathcal{S}}^*) \diff \boldsymbol{x}_{\thickbar{\mathcal{S}}},
    \end{split}
\end{align}
where $\boldsymbol{x}_{\mathcal{S}} = \{x_j:j \in \mathcal{S}\}$ denotes the features in subset $\mathcal{S}$, $\boldsymbol{x}_{\thickbar{\mathcal{S}}} = \{x_j:j \in \thickbar{\mathcal{S}}\}$ denotes the features outside $\mathcal{S}$, that is, $\thickbar{\mathcal{S}} = \mathcal{M}\backslash\mathcal{S}$, and $p(\boldsymbol{x}_{\thickbar{\mathcal{S}}} | \boldsymbol{x}_{\mathcal{S}} = \boldsymbol{x}_{\mathcal{S}}^*)$ is the conditional density of $\xsb$ given $\xs = \xss$. The conditional expectation summarizes the whole probability distribution, it is the most common estimator in prediction applications, and it is also the minimizer of the commonly used squared error loss function \parencite{aas2019explaining}. 
Note that the last equality of \eqref{eq:ContributionFunc} only holds for continuous features. If there are any discrete or categorical features, the integral should be replaced by sums for these features. Hence, $p(\boldsymbol{x}_{\thickbar{\mathcal{S}}} | \boldsymbol{x}_{\mathcal{S}} = \boldsymbol{x}_{\mathcal{S}}^*)$ is then no longer continuous. 


The contribution function in \eqref{eq:ContributionFunc} is also used by, for example, \textcite{covert2020understanding, aas2019explaining, aas2021explaining, frye_shapley-based_2020, Olsen2022}. 
\textcite{covert2021explaining} argue that the conditional approach in \eqref{eq:ContributionFunc} is the only approach that is consistent with standard probability axioms. Computing \eqref{eq:ContributionFunc} is not straightforward for a general data distribution and model. Assuming independent features, or having $f$ be linear, simplifies the computations \parencite{lundberg2017unified, aas2019explaining}, but these assumptions do not hold in general. 

To compute the Shapley values in \eqref{eq:ShapleyValuesDef}, we need to compute the contribution function $v(s)$ in \eqref{eq:ContributionFunc} for all $\s \in \pow(\M)$, except for the edge cases $\mathcal{S} \in \set{\emptyset, \M}$. For $\s = \M$, we have that $\xs = \x^*$ and $v(\mathcal{M}) = f(\x^*)$ by definition. For $\s = \emptyset$, we have by definition that $\phi_0 = v(\emptyset) = \E[f(\x)]$, where the average training response is a commonly used estimate \parencite{aas2019explaining}. We denote the non-trivial coalitions by $\pow^*(\M) = \pow(\M) \backslash \{\emptyset, \M\}$. The Shapley values $\bphi^* = \{\phi_j^*\}_{j=0}^M$ for the prediction $f(\x^*)$ are computed as the solution of a weighted least squares problem \parencite{charnes1988extremal,lundberg2017unified,aas2019explaining}.

In \Cref{subsubsec:ShapleyValuesExplainability:MonteCarlo,subsubsec:ShapleyValuesExplainability:Regression}, we describe two prominent paradigms for estimating the contribution function $v(\mathcal{S})$ for all $\s \in \pow^*(\M)$, namely, Monte Carlo integration and regression. 

\subsubsection{Monte Carlo Integration}
\label{subsubsec:ShapleyValuesExplainability:MonteCarlo}
One way to estimate the contribution function $v(\mathcal{S})$ is by using Monte Carlo integration. I.e.,
\begin{align}
    \label{eq:KerSHAPConditionalFunction}
    v(\mathcal{S})  
    =
    v(\mathcal{S}, \x^*)  
    =
    \E\left[ f(\boldsymbol{x}_{\thickbar{\mathcal{S}}}, \boldsymbol{x}_{\mathcal{S}}) | \boldsymbol{x}_{\mathcal{S}} = \boldsymbol{x}_{\mathcal{S}}^* \right] 
    \approx
    \frac{1}{K} \sum_{k=1}^K f(\boldsymbol{x}_{\thickbar{\mathcal{S}}}^{(k)}, \boldsymbol{x}_{\mathcal{S}}^*) 
    =    
    \hat{v}(\mathcal{S}),
\end{align}
where $f$ is the predictive model, $\boldsymbol{x}_{\thickbar{\mathcal{S}}}^{(k)} \sim p(\boldsymbol{x}_{\thickbar{\mathcal{S}}} | \boldsymbol{x}_{\mathcal{S}} = \boldsymbol{x}_{\mathcal{S}}^*)$, for $k=1,2,\dots,K$, and $K$ is the number of Monte Carlo samples. We insert $\hat{v}(\mathcal{S})$ into \eqref{eq:ShapleyValuesDef} to estimate the Shapley values. To obtain accurate conditional Shapley values we need to generate Monte Carlo samples which follow the true conditional distribution of the data. This distribution is in general not known and needs to be estimated based on the training data. In \Cref{subsec:ConditionalShapleyValues:independence,subsec:ConditionalShapleyValues:empirical,subsec:ConditionalShapleyValues:ParametricAssumption,subsec:ConditionalShapleyValues:GenerativeModel}, we describe different method classes for generating the conditional samples $\boldsymbol{x}_{\thickbar{\mathcal{S}}}^{(k)} \sim p(\boldsymbol{x}_{\thickbar{\mathcal{S}}} | \boldsymbol{x}_{\mathcal{S}} = \boldsymbol{x}_{\mathcal{S}}^*)$.

\subsubsection{Regression}
\label{subsubsec:ShapleyValuesExplainability:Regression}
As stated above, the conditional expectation \eqref{eq:ContributionFunc} is the minimizer of the mean squared error loss function. That is, 
\begin{equation}
\label{eq:ContributionFunc:Regression}
    v(\s) = v(\s, \x^*) = \E\left[ f(\xsb, \xs) | \xs = \xss \right] = \argmin_c \E\left[(f(\xsb, \xs) - c)^2 | \xs = \xss\right].
\end{equation}
Thus, any regression model $g_\s(\xs)$ which is fitted with the mean squared error loss function as the objective function will approximate \eqref{eq:ContributionFunc:Regression}, obtaining an alternative estimator $\hat{v}(\s)$. The accuracy of the approximation will depend on the form of the predictive model $f(\x)$, the flexibility of the regression model $g_\s(\xs)$, and the optimization routine. We can either train a separate regression model $g_\s(\xs)$ for each $\s \in \pow^*(\M)$ or we can train a single regression model $g(\tilde{\x}_\s)$ which approximates the contribution function $v(\s)$ for all $\s \in \pow^*(\M)$ simultaneously. Here $\tilde{\x}_\s$ is an augmented version of $\xs$ with fixed-length $M$, where the augmented values are mask values to be explained later. 
We elaborate on the notation and details of these two regression methodologies in \Cref{subsec:ConditionalShapleyValues:SeparateModels,subsec:ConditionalShapleyValues:SurrogateModel}, respectively.


\section{Conditional Expectation Estimation}
\label{sec:ConditionalShapleyValuesApproaches}
Computing conditional Shapley values is difficult due to the complexity of estimating the conditional distributions, which are not directly available from the training data. In this section, we give a methodological introduction to different methods for estimating the conditional expectation in \eqref{eq:ContributionFunc} via either Monte Carlo integration or regression, while we provide implementation details in \Cref{Appendix:Implementation}. We organize the methods into six method classes in accordance with those described in \textcite[Section 5.1.3]{chen2022algorithms} and \textcite[Section 8.2]{covert2021explaining}. The method classes we consider are called; \independence, \empirical, \parametric, \generative, \texttt{separate regression}, and \texttt{surrogate regression}, and they are described in \Cref{subsec:ConditionalShapleyValues:independence,subsec:ConditionalShapleyValues:empirical,subsec:ConditionalShapleyValues:ParametricAssumption,subsec:ConditionalShapleyValues:GenerativeModel,subsec:ConditionalShapleyValues:SeparateModels,subsec:ConditionalShapleyValues:SurrogateModel}, respectively. The first four classes estimate the conditional expectation in \eqref{eq:ContributionFunc} using Monte Carlo integration, while the last two classes use regression. 





\subsection{The Independence Method}
\label{subsec:ConditionalShapleyValues:independence}
\textcite{lundberg2017unified} avoided estimating the complex conditional distributions by implicitly assuming feature independence. In the \independence\ approach, the conditional distribution $p(\xsb|\xs)$ simplifies to $p(\xsb)$, and the corresponding Shapley values are the interventional Shapley values discussed in \Cref{Introduction}. The Monte Carlo samples $\boldsymbol{x}_{\thickbar{\mathcal{S}}}^{(k)} \sim p(\boldsymbol{x}_{\thickbar{\mathcal{S}}})$ are generated by randomly sampling observations from the training data, thus, no modeling is needed and $\boldsymbol{x}_{\thickbar{\mathcal{S}}}^{(k)}$ follows the assumed true data distribution. However, for dependent features, which is common in observational studies, the \independence\ approach produces biased estimates of the contribution function \eqref{eq:KerSHAPConditionalFunction} and the conditional Shapley values. Thus, the \independence\ approach can lead to incorrect Shapley value explanations for real-world data \parencite{aas2019explaining,merrick_explanation_2020,frye_shapley-based_2020,Olsen2022}.

\subsection{The Empirical Method}
\label{subsec:ConditionalShapleyValues:empirical}
Instead of sampling randomly from the training data, the \empirical\ method samples only from similar observations in the training data. The optimal procedure is to use only samples which perfectly match the feature values $\xss$, as this approach exactly estimates the conditional expectation when the number of matching observations tends to infinity \parencite{chen2022algorithms}. However, this is not applicable in practice, as data sets can have few observations, contain a high number of features to match, or continuous features where an exact match is very unlikely.
A natural extension is to relax the perfect match criterion and allow for similar observations \parencite{mase2019explaining,sundararajan2020many,aas2019explaining}. 
However, this procedure will also be influenced by the curse of dimensionality as conditioning on many features can yield few similar observations, and thereby inaccurate estimates of the conditional expectation \eqref{eq:ContributionFunc}. We can relax the similarity criterion and include less similar observations, but then we break the feature dependencies. The \empirical\ approach coincide with the \independence\ approach when the similarity measure defines all observations in the training data as similar. 

We use the \empirical\ approach described in \textcite{aas2019explaining}. 
The approach uses a scaled version of the Mahalanobis distance to calculate a distance $D_\s(\x^*, \x^{[i]})$ between the observation being explained $\x^*$ and every training instance $\x^{[i]}$. Then they use a Gaussian distribution kernel to convert the distance into a weight $w_\s(\x^*, \x^{[i]})$ for a given bandwidth parameter $\sigma$. All the weights are sorted in increasing order with $\x^{\set{k}}$ having the $k$th largest value. Finally, they approximate \eqref{eq:ContributionFunc} by a weighted version of \eqref{eq:KerSHAPConditionalFunction}, namely, $\hat{v}(\mathcal{S}) 
    =
    \sum_{k=1}^{K^*}
    [
    w_\s(\x^*, \x^{\set{k}})
    f(\boldsymbol{x}_{\thickbar{\mathcal{S}}}^{\set{k}}, \boldsymbol{x}_{\mathcal{S}}^*)
    ]
    \big/
    \sum_{k=1}^{K^*}
    w_\s(\x^*, \x^{\set{k}})$.
The number of samples used is $K^* = \min_{L \in \N} \big\{\sum_{k=1}^L w_\s(\x^*, \x^{\set{k}}) \big/ \sum_{i=1}^{N_\text{train}} w_\s(\x^*, \x^{[i]}) > \eta \big\}$, that is, the ratio between the sum of the $K^*$ largest weights and the sum of all weights must be at least $\eta$, for instance $0.95$.  

Note that as \textcite{aas2019explaining} use the Mahalanobis distance, their approach is limited to continuous features. One could potentially extend their method by using a distance measure which supports mixed data, for example, the Gower's distance \parencite{gower1971general,podani1999extending}. Another solution is to use, for example, encodings like one-hot-encoding or entity embeddings to represent the categorical variables as numerical \parencite{guo2016entity}, although that would increase the computational demand due to increased dimension.

\subsection{The Parametric Method Class}
\label{subsec:ConditionalShapleyValues:ParametricAssumption}
In the \parametric\ method class, we make a parametric assumption about the distribution of the data. This simplifies the process of generating the conditional Monte Carlo samples $\boldsymbol{x}_{\thickbar{\mathcal{S}}}^{(k)} \sim p(\boldsymbol{x}_{\thickbar{\mathcal{S}}} | \boldsymbol{x}_{\mathcal{S}} = \boldsymbol{x}_{\mathcal{S}}^*)$. The idea is to assume a distribution whose conditional distributions have closed-form solutions or are otherwise easily obtainable after estimating the parameters of the full joint distribution. The \parametric\ approaches can yield very accurate representations if the data truly follows the assumed distribution, but they may impose large bias for incorrect parametric assumptions. In this section, we discuss two previously proposed \parametric\ approaches and introduce two new methods. The current \parametric\ approaches do not support categorical features, which is a major drawback, but one can potentially use the same type of encodings or entity embeddings of the categorical variables as for the \empirical\ method.

\subsubsection{Gaussian}
\label{subsec:ConditionalShapleyValues:ParametricAssumption:Gaussian}
Both \textcite{chen2020true, aas2019explaining} assume that the observations are multivariate Gaussian distributed with mean $\bmu$ and covariance matrix $\Sigma$. That is, $p(\x) = p(\xs, \xsb) = \mathcal{N}_M(\bmu, \bSigma)$, where $\bmu = [\bmu_{\s}, \bmu_{\sbb}]^T$ and $\boldsymbol{\Sigma} = \Big[\begin{smallmatrix} \bSigma_{\s\s} & \bSigma_{\s\sbb} \\ \bSigma_{\sbb\s} & \bSigma_{\sbb\sbb} \end{smallmatrix}\Big]$. The conditional distribution is also multivariate Gaussian, that is, $p(\xsb| \xs = \xss) = \mathcal{N}_{|\sbb|} (\bmu_{\sbb|\s}, \bSigma_{\sbb|\s})$, where $\bmu_{\sbb|\s} = \bmu_{\sbb} + \bSigma_{\sbb\s} \bSigma_{\s\s}^{-1}(\xss - \bmu_{\s})$ and $\bSigma_{\sbb|\s} = \bSigma_{\sbb\sbb} - \bSigma_{\sbb\s}\bSigma_{\s\s}^{-1}\bSigma_{\s\sbb}$. The parameters $\bmu$ and $\bSigma$ are easily estimated using the sample mean and covariance matrix of the training data, respectively. In the \Gaussian\ approach, we sample the conditional samples $\boldsymbol{x}_{\thickbar{\mathcal{S}}}^{(k)}$ from $p(\boldsymbol{x}_{\thickbar{\mathcal{S}}} | \boldsymbol{x}_{\mathcal{S}} = \boldsymbol{x}_{\mathcal{S}}^*)$, for $k=1,2,\dots,K$ and $\s \in \pow^*(\M)$, and use them in \eqref{eq:KerSHAPConditionalFunction} to estimate the Shapley values in \eqref{eq:ShapleyValuesDef}.  

\subsubsection{Gaussian Copula}
\label{subsec:ConditionalShapleyValues:ParametricAssumption:GaussianCopula}
\textcite{aas2019explaining} also proposed an alternative approach if the features are far from multivariate Gaussian distributed, namely the (Gaussian) \copula\ approach. The idea is to represent the marginals of the features by their empirical distributions and then model the dependence structure by a Gaussian copula. See \Cref{Appendix:AdditionalInformationParametric:Copula} for additional information about copulas. 

Assuming a Gaussian copula, \textcite{aas2019explaining} use the following procedure to generate the $K$ conditional Monte Carlo samples $\boldsymbol{x}_{\thickbar{\mathcal{S}}}^{(k)} \sim p(\boldsymbol{x}_{\thickbar{\mathcal{S}}} | \boldsymbol{x}_{\mathcal{S}} = \boldsymbol{x}_{\mathcal{S}}^*)$:
\begin{enumerate}
    \item Convert each marginal $x_j$ of the feature distribution $\boldsymbol{x}$ to a Gaussian feature $v_j$ by $v_j = \Phi^{-1}(\hat{F}(x_j))$, where $\hat{F}(x_j)$ is the empirical distribution function of marginal $j$.
    \item Assume that $\boldsymbol{v}$ is distributed according to a multivariate Gaussian (the quality of this assumption will depend on how close the Gaussian copula is to the true copula), and sample from the conditional distribution $p(\boldsymbol{v}_{\sbb} | \boldsymbol{v}_{\s} = \boldsymbol{v}_{\s}^*)$ using the method described in \Cref{subsec:ConditionalShapleyValues:ParametricAssumption:Gaussian}.
    \item Convert the margins $v_j$ in the conditional distribution to the original distribution using $\hat{x}_j = \hat{F}_j^{-1}(\Phi(v_j))$.
\end{enumerate}

\subsubsection{Burr and Generalized Hyperbolic}
\label{subsec:ConditionalShapleyValues:ParametricAssumption:Burr}
The multivariate Gaussian distribution is probably the most well-known multivariate distribution with closed-form expressions for the conditional distributions. However, any other distribution with easily obtainable conditional distributions is also applicable, for example, the multivariate Burr distribution \parencite{takahasi_note_1965,Yari2006InformationAC} and the multivariate generalized hyperbolic (GH) distribution \parencite{barndorff1977exponentially,mcneil2015quantitative,browne2015mixture,wei2019mixtures}. We call these two approaches for \tm{Burr} and \tm{GH}, respectively. In contrast to the Gaussian distribution, whose parameters can easily be estimated by the sample means and covariance, the parameters of the Burr and GH distributions are more cumbersome to estimate. We describe the distributions in more details in \Cref{Appendix:AdditionalInformationParametric}. The GH distribution is unbounded and can model any continuous data set, while the Burr distribution is strictly positive and is therefore limited to positive data sets. 
The GH distribution is related to the Gaussian distribution through the t-distribution, where the latter is a special case of the GH distribution and coincide with the Gaussian distribution when the degrees of freedom tends to infinity.

\subsection{The Generative Method Class}
\label{subsec:ConditionalShapleyValues:GenerativeModel}
The \generative\ and \parametric\ methods are similar in that they both generate Monte Carlo samples from the estimated conditional distributions. However, the \generative\ methods do not make a parametric assumption about the data. We consider two \generative\ approaches; the \ctree\ approach of \textcite{redelmeier2020} and the \vaeac\ approach of \textcite{Olsen2022}. The latter is an extension of the approach suggested by \textcite{frye_shapley-based_2020}. Both methods support mixed, i.e., continuous and categorical, data. 


\subsubsection{Ctree}
\label{subsec:ConditionalShapleyValues:GenerativeModel:Ctree}
\textcite{redelmeier2020} compute conditional Shapley values by modeling the dependence structure between the features with conditional inference trees (\ctree). A \ctree\ is a type of recursive partitioning algorithm that builds trees recursively by making binary splits on features until a stopping criterion is satisfied \parencite{HothornCtree}. The process is sequential, where the splitting feature is chosen first using statistical significance tests, and then the splitting point is chosen using any type of splitting criterion. The \ctree\ algorithm is independent of the dimension of the response, which in our case is $\xsb$, while the input features are $\xs$, which varies in dimension based on the coalition $\s$. That is, for each coalition $\s \in \pow^*(\M)$, a \ctree\ with $\xs$ as the features and $\xsb$ as the response is fitted to the training data. For a given $\xss$, the \ctree\ approach finds the corresponding leaf node and samples $K$ observations with replacement from the $\xsb$ part of the training observations in the same node to generate the conditional Monte Carlo samples $\boldsymbol{x}_{\thickbar{\mathcal{S}}}^{(k)} \sim p(\boldsymbol{x}_{\thickbar{\mathcal{S}}} | \boldsymbol{x}_{\mathcal{S}} = \boldsymbol{x}_{\mathcal{S}}^*)$. We get duplicated Monte Carlo samples when $K$ is larger than the number of samples in the leaf node. Thus, the \ctree\ method weight the Monte Carlo samples based on their sampling frequencies to bypass redundant calls to $f$. Therefore, the contribution function $v(\s)$ is not estimated by \eqref{eq:KerSHAPConditionalFunction} but rather by the weighted average $\hat{v}(\s) = \sum_{k=1}^{K^*} w_kf(\boldsymbol{x}_{\thickbar{\mathcal{S}}}^{(k)}, \boldsymbol{x}_{\mathcal{S}}^*) \big/ \sum_{k=1}^{K^*}w_k$, where $K^*$ is the number of unique Monte Carlo samples. For more details, see \textcite[Section 3]{redelmeier2020}.



\subsubsection{VAEAC}
\label{subsec:ConditionalShapleyValues:GenerativeModel:VAEAC}
\textcite{Olsen2022} use a type of variational autoencoder called \vaeac\ \parencite{ivanov_variational_2018} to generate the conditional Monte Carlo samples. Briefly stated, the original variational autoencoder \parencite{kingma2014autoencoding,Kingma2019AnIT,pmlr-v32-rezende14} gives a probabilistic representation of the true unknown distribution $p(\boldsymbol{x})$. The \vaeac\ model extends this methodology to all conditional distributions $p(\xsb | \xs = \xss)$ simultaneously. That is, a single \vaeac\ model can generate Monte Carlo samples $\x_{\thickbar{\mathcal{S}}}^{(k)} \sim p(\boldsymbol{x}_{\thickbar{\mathcal{S}}} | \boldsymbol{x}_{\mathcal{S}} = \boldsymbol{x}_{\mathcal{S}}^*)$ for all coalitions $\mathcal{S} \in \pow^*(\mathcal{M})$. It is advantageous to only have to fit a single model for all coalitions, as in higher dimensions the number of coalitions is $2^M-2$. That is, the number of coalitions increases exponentially with the number of features. In contrast, \ctree\ trains $2^M-2$ different models, which eventually becomes computationally intractable for large $M$. The \vaeac\ model is trained by maximizing a variational lower bound, which conceptually corresponds to artificially masking features and then trying to reproduce them using a probabilistic representation. In deployment, the \vaeac\ method considers the unconditional features $\xsb$ as masked features to be imputed.

\subsection{The Separate Regression Method Class}
\label{subsec:ConditionalShapleyValues:SeparateModels}
The next two method classes use regression instead of Monte Carlo integration to estimate the conditional expectation in \eqref{eq:ContributionFunc}. 
In the \separatereg\ methods, we train a new regression model $g_\s(\xs)$ to estimate the conditional expectation for each coalition of features. 
Related ideas have been explored by \textcite{lipovetsky2001analysis,strumbelj2009explaining,williamson2020efficient}. However, to the best of our knowledge, we are the first to compare different regression models for estimating the conditional expectation as the contribution function $v(\mathcal{S})$ in the local Shapley value explanation framework. 

The idea is to estimate $v(\mathcal{S}) = \E\left[ f(\x) | \xs = \xss \right] = \E\left[ f(\xsb, \xs) | \xs = \xss \right]$ separately for each coalition $\s$ using regression. As in \Cref{subsec:ShapleyValuesExplainability}, let $\mathcal{X} = \{ \x^{[i]}, y^{[i]} \}_{i=1}^{N_{\text{train}}}$ denote the training data, where $\x^{[i]}$ is the $i$th $M$-dimensional input and $y^{[i]}$ is the associated response. For each $\s \in \pow^*(\M)$, the corresponding training data set is
\begin{align*}
            \mathcal{X}_\mathcal{S} 
            =
            \{\x_\s^{[i]}, f(\underbrace{\x_\sbb^{[i]}, \x_\s^{[i]}}_{\x^{[i]}})\}_{i=1}^{N_{\text{train}}}
            =
            \{\x_\s^{[i]}, \underbrace{f(\x^{[i]})}_{z^{[i]}}\}_{i=1}^{N_{\text{train}}}
            =
            \{\x_\s^{[i]}, z^{[i]}\}_{i=1}^{N_{\text{train}}}.
\end{align*}

For each data set $\mathcal{X}_\s$, we train a regression model $g_\s(\xs)$ with respect to the mean squared error loss function. The optimal model, with respect to the loss function, is $g^*_\s(\xs) = \E[z|\xs] = \E[f(\xsb, \xs)|\xs]$, which corresponds to the contribution function $v(\s)$. The regression model $g_\s$ aims for the optimal, hence, it resembles/estimates the contribution function, i.e., $g_\s(\xs) = \hat{v}(\mathcal{S}) \approx v(\mathcal{S}) = \E[f(\xsb, \xs) | \xs = \xss]$. 

A wide variety of regression models minimize the MSE, and we describe a selection of them in \Cref{ConditionalShapleyValues:SeparateModels:linear,ConditionalShapleyValues:SeparateModels:gam,ConditionalShapleyValues:SeparateModels:ppr,ConditionalShapleyValues:SeparateModels:rf,ConditionalShapleyValues:SeparateModels:boosting}. The selection discussed below consists of classical regression models and those that generally perform well for many experiments. 


\subsubsection{Linear Regression Model}
\label{ConditionalShapleyValues:SeparateModels:linear}
The simplest regression model we consider is the linear regression model. It takes the form $g_\s(\xs) = \beta_{\s,0} + \sum_{j \in S} \beta_{\s,j}x_j = \x_\s^T\boldsymbol{\beta}_\s$, where the coefficients $\boldsymbol{\beta}_\s$ are estimated by the least squares solution, that is, $\hat{\boldsymbol{\beta}}_\s = \argmin_{\boldsymbol{\beta}} \|X_\s\boldsymbol{\beta} - \z\|^2 = (X_\s^TX_\s)^{-1}X_\s^T\z$, for all $\s \in \mathcal{P}^*(\mathcal{M})$. 
Here $X_\s$ is the design matrix with the first column consisting of $1$s to also estimate the intercept $\beta_{\s,0}$. We call this approach \texttt{LM separate}.

\subsubsection{Generalized Additive Model}
\label{ConditionalShapleyValues:SeparateModels:gam}
The generalized additive model (GAM) extends the linear regression model and allows for nonlinear effects between the features and the response \parencite{wood2006generalized,hastie2009elements}. The fitted GAM takes the form $g_\s(\xs) = \beta_{\s,0} + \sum_{j \in \s} g_{\s,j}(x_{\s,j})$, where the effect functions $g_{\s,j}$ are penalized regression splines. We call this approach \texttt{GAM separate}.

\subsubsection{Projection Pursuit Regression}
\label{ConditionalShapleyValues:SeparateModels:ppr}
The projection pursuit regression (PPR) model extends the GAM model \parencite{friedman1981projection, hastie2009elements}. The PPR model takes the form $g_\s(\xs) = \beta_{\s,0} + \sum_{l=1}^L g_{\s,l}(\boldsymbol\beta_{\s,l}^T\xs)$, where the parameter vector $\boldsymbol\beta_{\s,l}$ is an $|\s|$-dimensional unit vector. The PPR is an additive model, but in the transformed features $\boldsymbol\beta_{\s,l}^T\xs$ rather than in the original features $\xs$. The ridge functions $g_{\s,l}$ are unspecified, and are estimated along with the parameters $\boldsymbol\beta_{\s,l}$ using some flexible smoothing method. The PPR model combines nonlinear functions of linear combinations, producing a large class of potential models. Moreover, it is an universal approximator for continuous functions for arbitrary large $L$ and appropriate choice of $g_{\s,l}$ \parencite[Section 11.2]{hastie2009elements}. We call this approach \texttt{PPR separate}.

\subsubsection{Random Forest}
\label{ConditionalShapleyValues:SeparateModels:rf}
A random forest (RF) is an ensemble model consisting of a multitude of decision trees, where the average prediction of the individual trees is returned. The first algorithm was developed by \textcite{ho1995random}, but \textcite{breiman2001random} later extended the algorithm to include bootstrap aggregating to improve the stability and accuracy. We call this approach \texttt{RF separate}.

\subsubsection{Boosting}
\label{ConditionalShapleyValues:SeparateModels:boosting}
A (tree-based) boosted model is an ensemble learner consisting of weighted weak base-learners which has been iteratively fitted to the error of the previous base-learners and together they form a strong learner \parencite{hastie2009elements}. The seminal boosting algorithm was developed by \textcite{freund1997decision}, but multitudes of boosting algorithms has later been developed \parencite{mayr2014evolution}, for example, \texttt{CatBoost} \parencite{catboost}. We call this approach \texttt{CatBoost separate}.

\subsection{The Surrogate Regression Method Class}
\label{subsec:ConditionalShapleyValues:SurrogateModel}
Since the \separatereg\ methods train a new regression model $g_\s(\xs)$ for each coalition $\mathcal{S} \in \pow^*(\M)$, a total of $2^M-2$ models has to be trained, which can be time-consuming for slowly fitted models. The \surrogatereg\ method class builds on the ideas from the \separatereg\ class, but instead of fitting a new regression model for each coalition, we train a single regression model $g(\tilde{\x}_\s)$ for all coalitions $\s \in \pow^*(\M)$, where $\tilde{\x}_\s$ is defined in \Cref{ConditionalShapleyValues:SurrogateModel:NewMethods}. The \surrogatereg\ idea is used by \textcite{frye_shapley-based_2020, covert2021explaining}, but their setup is limited to neural networks. In \Cref{ConditionalShapleyValues:SurrogateModel:NewMethods}, we propose a general and novel framework which allows us to use any regression model. Then, we relate our framework to the previously proposed neural network setup in \Cref{ConditionalShapleyValues:SurrogateModel:NN}.



\subsubsection{General Surrogate Regression Framework}
\label{ConditionalShapleyValues:SurrogateModel:NewMethods}
To construct a \surrogatereg\ method, we must consider that most regression models $g$ rely on a fixed-length input, while the size of $\xs$ varies with coalition $\s$. Thus, we are either limited to regression models which support variable-length input, or we can create a fixed-length representation $\tilde{\x}_\s$ of $\xs$ for all coalitions $\s$. The $\tilde{\x}_\s$ representation must also include fixed-length information about the coalition $\s$ to enable the regression model $g$ to distinguish between coalitions. Finally, we need to augment the training data to reflect that $g$ is to predict the conditional expectation for all coalitions $\s$.


In our framework, we augment the training data by systematically applying all possible coalitions to all training observations. We can then train a single regression model $g$ on the augmented training data set, and the corresponding regression model can then (in theory) estimate the contribution function $v(\s)$ for all coalitions $\s \in \mathcal{P}^*(\mathcal{M})$ simultaneously. 

To illustrate the augmentation idea, we consider a small example with $M = 3$ features and $N_\text{train} = 2$ training observations. Let 
$\mathcal{X} = 
\big[
\begin{smallmatrix}
    x_{11} & x_{12} & x_{13} \\
    x_{21} & x_{22} & x_{23} 
\end{smallmatrix}\big]$ and $\z = 
\big[
\begin{smallmatrix}
    f(\x_{1})  \\
    f(\x_{2})  
\end{smallmatrix}\big]$ denote the training data and responses, respectively. In this setting, $\M = \set{1,2,3}$ and $\pow^*(\M) = \set{\set{1}, \set{2}, \set{3}, \set{1,2}, \set{1,3}, \set{2,3}}$ consists of six different coalitions $\s$, or equivalently masks $\sbb = \M \backslash \s$. Assuming that $g$ relies on fixed-length input, we must represent both the observed values $\xs$ and coalition $\s$ in a fixed-length notation for the \surrogate\ methods to work.  

To solve this, we first introduce $I(\s) = \left\{\mathbf{1}(j \in \s): j = 1, \dots, M\right\} \in \{0, 1\}^M$, where $\mathbf{1}(j \in \s)$ is the indicator function which is one if $j \in \s$ and zero otherwise. Then, $I(\s)$ is an $M$-dimensional binary vector where the $j$th element $I(\s)_j$ is one if the $j$th feature is in $\s$ (i.e., observed/conditioned on) and zero if it is in $\sbb$ (i.e., unobserved/unconditioned). The $I$ function ensures fixed-length representations of the coalitions/masks, and note that $I(\sbb) = \mathbf{1}^M - I(\s)$, where $\mathbf{1}^M$ is the size $M$ vector of $1$s. Second, to obtain a fixed-length representation $\hat{\x}_{\s}$ of the observed/conditioned feature vector $\x_{\s}$, we apply the fixed-length mask $I(\s)$ to $\x$ as an element-wise product, that is, $\hat{\x}_{\s} = \x \circ I(\s) = \x \circ (\mathbf{1}^M-I(\sbb))$, where $\circ$ is the element-wise product. Finally, we concatenate the fixed-length representations together to form the augmented version of $\xs$, namely, $\tilde{\x}_{\s} = \{\hat{\x}_{\s}, I(\sbb)\}$, which has $2M$ entries. We include $I(\sbb)$ in $\tilde{\x}_{\s}$ such that the model $g$ can distinguish between actual zeros in $\xs$ and those induced by the masking procedure when creating $\hat{\x}_{\s}$. We treat $I(\sbb)$ as binary categorical features.


After carrying out this procedure for all coalitions and training observations, we obtain the following augmented training data and responses:
\begin{align}
\label{eq:surrogate_augmented_data}
    \mathcal{X}_\text{aug}  
    =
    \begin{bmatrix}
    \tilde{\x}_{1,\set{1}} \\
    \tilde{\x}_{1,\set{2}} \\
    \tilde{\x}_{1,\set{3}} \\
    \tilde{\x}_{1,\set{1,2}} \\
    \tilde{\x}_{1,\set{1,3}} \\
    \tilde{\x}_{1,\set{2,3}} \\
    \tilde{\x}_{2,\set{1}} \\
    \tilde{\x}_{2,\set{2}} \\
    \tilde{\x}_{2,\set{3}} \\
    \tilde{\x}_{2,\set{1,2}} \\
    \tilde{\x}_{2,\set{1,3}} \\
    \tilde{\x}_{2,\set{2,3}}
    \end{bmatrix}
    =
    \begin{bmatrix}
        \phantom{\,\,}x_{11} & 0      & 0 \phantom{\,\,}     & 0 & 1 & 1 \\
        \phantom{\,\,}0      & x_{12} & 0 \phantom{\,\,}     & 1 & 0 & 1 \\
        \phantom{\,\,}0      & 0      & x_{13}\phantom{\,\,} & 1 & 1 & 0 \\
        \phantom{\,\,}x_{11} & x_{12} & 0  \phantom{\,\,}    & 0 & 0 & 1 \\
        \phantom{\,\,}x_{11} & 0      & x_{13}\phantom{\,\,} & 0 & 1 & 0 \\
        \phantom{\,\,}0      & x_{12} & x_{13}\phantom{\,\,} & 1 & 0 & 0 \\
        \phantom{\,\,}x_{21} & 0      & 0 \phantom{\,\,}     & 0 & 1 & 1 \\
        \phantom{\,\,}0      & x_{22} & 0 \phantom{\,\,}     & 1 & 0 & 1 \\
        \phantom{\,\,}0      & 0      & x_{23}\phantom{\,\,} & 1 & 1 & 0 \\
        \phantom{\,\,}x_{21} & x_{22} & 0 \phantom{\,\,}     & 0 & 0 & 1 \\
        \phantom{\,\,}x_{21} & 0      & x_{23}\phantom{\,\,} & 0 & 1 & 0 \\
        \undermat{\text{Observed values}}{\phantom{\,\,}0    & x_{22} & x_{23}\phantom{\,\,}} & \undermat{\text{Mask}}{1 & 0 & 0}
    \end{bmatrix} \text{ and }  
    \z_\text{aug}  
    =
    \begin{bmatrix}
    f(\x_{1})  \\
    f(\x_{1})  \\
    f(\x_{1})  \\
    f(\x_{1})  \\
    f(\x_{1})  \\
    f(\x_{1})  \\
    f(\x_{2})  \\
    f(\x_{2})  \\
    f(\x_{2})  \\
    f(\x_{2})  \\
    f(\x_{2})  \\
    f(\x_{2})
    \end{bmatrix}
    \!.
\end{align}
\vspace{2ex}

\noindent The number of rows in $\mathcal{X}_\text{aug}$ and $\z_\text{aug}$ is $N_\text{train}(2^M-2)$. For example, with $N_\text{train} = 1000$ and $M=8$ the augmented data $\mathcal{X}_\text{aug}$ consists of $254\,000$ rows, while the number of rows is $65\,534\,000$ when $M=16$. This exponential growth can make it computationally intractable to fit some types of regression models to the augmented training data $\set{\mathcal{X}_\text{aug}, \z_\text{aug}}$ in high-dimensions. The data instance $\x^*$, which we want to explain, is augmented by the same procedure, and $g(\tilde{\x}^*_{\s})$ then approximates the corresponding contribution function $v(\s, \x^*)$. 

\paragraph{Methods:}
For the \surrogatereg\ method class, we consider the same regression models as in \Cref{subsec:ConditionalShapleyValues:SeparateModels}. We call the methods for \texttt{LM surrogate}, \texttt{GAM surrogate}, \texttt{PPR surrogate}, \texttt{RF surrogate}, and \texttt{CatBoost surrogate}, and they take the following forms:
\begin{enumerate}[align=left, leftmargin=3em, itemsep=2pt, topsep=6pt] 
    \item [\texttt{LM surrogate}:] $g(\tilde{\x}_{\s}) = \beta_{0} + \sum_{j = 1}^{2M} \beta_{j}\tilde{x}_{\s, j} = \tilde{\x}_\s^{T}\boldsymbol{\beta}$. 
    \item [\texttt{GAM surrogate}:]  $g(\tilde{\x}_{\s}) = \beta_{0} + \sum_{j = 1}^{M} g_{j}(\tilde{x}_{\s, j}) + \sum_{j = M+1}^{2M} \beta_{j}\tilde{x}_{\s, j}$. That is, we add nonlinear effect functions to the augmented features $\hat{\x}_{\s} = \x \circ I(\s)$ in $\tilde{\x}_{\s}$ while letting the binary mask indicators $I(\sbb)$ in $\tilde{\x}_{\s}$ be linear.
    
    \item [\texttt{PPR surrogate}:] $g(\tilde{\x}_{\s}) = \beta_{0} + \sum_{l=1}^L g_{l}(\boldsymbol\beta_{l}^T\tilde{\x}_{\s})$, where $g_{l}$ and $\boldsymbol\beta_{l}$ are the $l$th ridge function and parameter vector, respectively.
    \item [\texttt{RF surrogate}:] $g(\tilde{\x}_{\s})$ is a \texttt{RF} model fitted to the augmented data on the same form as in \eqref{eq:surrogate_augmented_data}.
    \item [\texttt{CatBoost surrogate}:] $g(\tilde{\x}_{\s})$ is a \texttt{CatBoost} model fitted to the augmented data on the same form as in \eqref{eq:surrogate_augmented_data}.
\end{enumerate}

\subsubsection{Surrogate Regression: Neural Networks}
\label{ConditionalShapleyValues:SurrogateModel:NN}
The \surrogate\ neural network (\texttt{NN-Frye surrogate}) approach in \textcite{frye_shapley-based_2020} differs from our general setup above in that they do not train the model on the complete augmented data. Instead, for each observation in every batch in the training process, they randomly sample a coalition $\s$ with probability $\frac{|\mathcal{S}|!(M-|\mathcal{S}|-1)!}{M!}$. Then they set the masked entries of the observation, i.e., the features not in $\s$, to an off-distribution value not present in the data. Furthermore, they do \emph{not} concatenate the masks to the data, as we do in \eqref{eq:surrogate_augmented_data}. 

We propose an additional neural network (\texttt{NN-Olsen surrogate}) approach to illustrate that one can improve on the \texttt{NN-Frye surrogate} method. The main conceptual differences between the methods are the following. First, for each batch, we generate a missing completely at random (MCAR) mask with paired sampling. MCAR means that the binary entries in the mask $\sbb$ is Bernoulli distributed with probability $0.5$, which ensures that all coalitions are equally likely to be considered. Further, paired sampling means that we duplicate the observations in the batch and apply the complement mask, $\s$, on these duplicates. This ensures more stable training as the network can associate both $\x_{\s}$ and $\x_{\sbb}$ with the response $f(\x)$. Second, we set the masked entries to zero and include the binary mask entries as additional features, as done in \eqref{eq:surrogate_augmented_data} and \textcite{Olsen2022}. This enables the network to learn to distinguish actual zeros in the data set and zeros induced by the masking, removing the need to set an off-distribution masking value. Additional differences due to implementation, for example, network architecture and optimization routine, are elaborated in \Cref{Appendix:Implementation}.

\subsection{Additional Methods in the Appendix}
\label{subsec:ConditionalShapleyValues:AdditionalMethods}
In addition to the methods described in \Cref{subsec:ConditionalShapleyValues:independence,subsec:ConditionalShapleyValues:empirical,subsec:ConditionalShapleyValues:ParametricAssumption,subsec:ConditionalShapleyValues:GenerativeModel,subsec:ConditionalShapleyValues:SeparateModels,subsec:ConditionalShapleyValues:SurrogateModel}, we include dozens more \generative, \separate, and \surrogate\ methods in the \nameref{Appendix}. These methods are not included in the main text as they generally perform worse than the introduced methods. For the \generative\ method class, we consider three additional \vaeac\ approaches with methodological differences 
and point to eleven other potential generative methods. For the \separate\ method class, we consider twenty other regression models, and most of these are also applicable to the \surrogate\ method class. Among the regression methods are: linear regression with interactions, polynomial regression with and without interactions, elastic nets, generalized additive models, principal component regression, partial least squares, K-nearest neighbors, support vector machines, decision trees, boosting, and neural networks. In the \nameref{Appendix}, we apply the additional methods to the numerical simulation studies and real-world data experiments conducted in \Cref{sec:Numerical_Simulation_Studies,sec:real_world_data}, respectively.

\section{Numerical Simulation Studies}
\label{sec:Numerical_Simulation_Studies}

A major problem of evaluating explanation frameworks is that there is no ground truth for authentic real-world data. In this section, we simulate data for which we can compute the true Shapley values $\bphi_\text{true}$ and compare how close the estimated Shapley values $\hat{\bphi}_\texttt{q}$ are when using approach $\texttt{q}$. We gradually increase the complexity of the setups in the simulation studies to uncover in which settings the different methods described in \Cref{sec:ConditionalShapleyValuesApproaches} perform the best and should be used. 
Additionally, as we focus on conditional Shapley values, we vary the dependencies between the features within each simulation setup to also investigate how the methods cope with different dependence levels.

In all experiments, we generate univariate prediction problems with $M = 8$-dimensional features simulated from a multivariate Gaussian distribution $p(\x) = \mathcal{N}_{8}(\0, \Sigma)$, where $\Sigma_{ij} = \rho^{\abs{i-j}}$ for $\rho \in \set{0, 0.3, 0.5, 0.9}$ and $1$ on the diagonal. Larger values of $\rho$ correspond to higher dependencies between the features. Higher feature dimensions are possible, but we chose $M=8$ to keep the computation time of the simulation studies tractable. The real-word data sets in \Cref{sec:real_world_data} contain more features. See also \textcite[Section 5.2]{chen2022algorithms} for estimation strategies used in the literature to compute Shapley values in higher dimensions. 

We let the number of training observations be $N_\text{train} = 1000$, while we explain $N_\text{test} = 250$ test observations. Thus, the training data set is $\set{\boldsymbol{x}^{[i]}, y^{[i]}}_{i=1}^{N_{\text{train}}}$, where $\x^{[i]} \sim\mathcal{N}_{8}(\0, \Sigma)$ and the response $y^{[i]} = f_\text{true}(\boldsymbol{x}^{[i]}) + \varepsilon^{[i]}$. The function $f_\text{true}$ is different in different experiments and $\varepsilon^{[i]} \sim \mathcal{N}(0, 1)$. The test data sets are created by the same procedure.  We provide additional experiments with other settings and some illustrative plots of the data in \Cref{Appendix:PairPlots,Appendix:ExtendedSimulations}. 


We evaluate the performance of the different approaches by computing the mean absolute error (MAE) between the true and estimated Shapley values, averaged over all test observations and features. This criterion has been used in \textcite{redelmeier2020, aas2019explaining, aas2021explaining, Olsen2022}.
The MAE is given by
\begin{align}
    \label{eq:MAE}
    \operatorname{MAE} 
    = 
    \operatorname{MAE}_{\phi}(\text{method } \texttt{q}) 
    =
    \frac{1}{N_\text{test}} \sum_{i=1}^{N_\text{test}} \frac{1}{M} \sum_{j=1}^M |\phi_{j, \texttt{true}}(\boldsymbol{x}^{[i]}) - \hat{\phi}_{j, \texttt{q}}(\boldsymbol{x}^{[i]})|.  
\end{align}

The true Shapley values are in general unknown, but we can compute them with arbitrary precision in our setup with multivariate Gaussian distributed data, as the conditional distributions $p_\texttt{true}(\xsb|\xs)$ are analytically known and samplable for all $\s$. Thus, by sampling $\boldsymbol{x}_{\thickbar{\mathcal{S}}, \texttt{true}}^{(k)} \sim p_\texttt{true}(\xsb|\xs)$, we can compute the true contribution function $v_\texttt{true}(\s)$ in \eqref{eq:ContributionFunc} by using \eqref{eq:KerSHAPConditionalFunction}.
The true Shapley values are then obtained by inserting the $v_\texttt{true}(\s)$ quantities into the Shapley value formula in \eqref{eq:ShapleyValuesDef}. The $v_\texttt{true}(\s)$ quantities can be arbitrarily precise by choosing a sufficiently large number of Monte Carlo samples, e.g., $K = 10\, 000$. 



\subsection{Linear Regression Models}
\label{subsection:Simulation:lm}
The first simulation setup should be considered a sanity check, as we generate the response $y^{[i]}$ according to the following linear regression models:
\begin{enumerate}[align=left, leftmargin=3.5em, itemsep=1pt, topsep=4pt] 
    \item [\texttt{lm\textunderscore no\textunderscore interactions}:] $f_{\lm, \no}(\x) =  \beta_0 + \sum_{j=1}^{M} \beta_jx_j$, 
    
    \item [\texttt{lm\textunderscore more\textunderscore interactions}:] $f_{\lm, \more}(\x) =  f_{lm, \no}(\x) + \gamma_1x_1x_2 + \gamma_2x_3x_4$,
    
    \item [\texttt{lm\textunderscore numerous\textunderscore interactions}:] $f_{\lm, \numerous}(\x) =  f_{\lm, \more}(\x) + \gamma_3x_5x_6 + \gamma_4x_7x_8$, 
\end{enumerate}
where $\boldsymbol{\beta} = \{1.0,  0.2, -0.8, 1.0, 0.5, -0.8, 0.6, -0.7, -0.6\}$ and $\boldsymbol{\gamma} = \{0.8, -1.0 -2.0, 1.5\}$. For each setup, we fit a predictive linear model $f$ with the same form as the true model. E.g., for the \texttt{lm\textunderscore more\textunderscore interactions} setup, the predictive linear model $f$ has eight linear terms and two interaction terms reflecting the form of $f_{\lm, \more}$. We fit the predictive models using the \texttt{lm} function in base \textsc{R}.

\begin{figure}[!t]
   \centering
   \centerline{\includegraphics[width=1\textwidth]{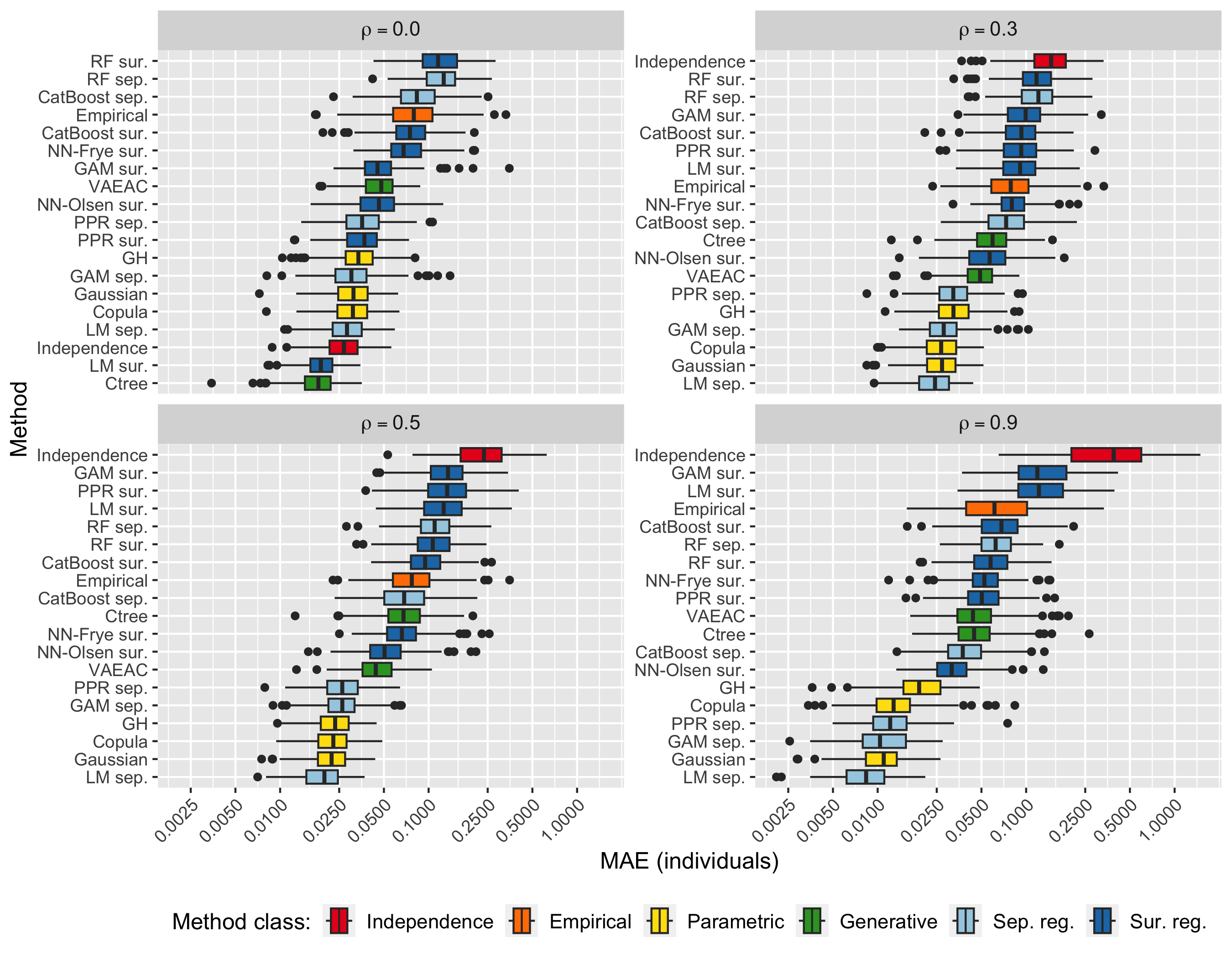}}
   \caption{{\small Results of the \texttt{lm\textunderscore no\textunderscore interactions} experiment: boxplots of the mean absolute error between the true and estimated Shapley values for the test individuals using different methods and for different dependence levels $\rho$. The methods are sorted based on the mean over all test observations.}}
   \label{fig:lm_no}
\end{figure}

\begin{figure}[!t]
    \centering
    \centerline{\includegraphics[width=1.0\textwidth]{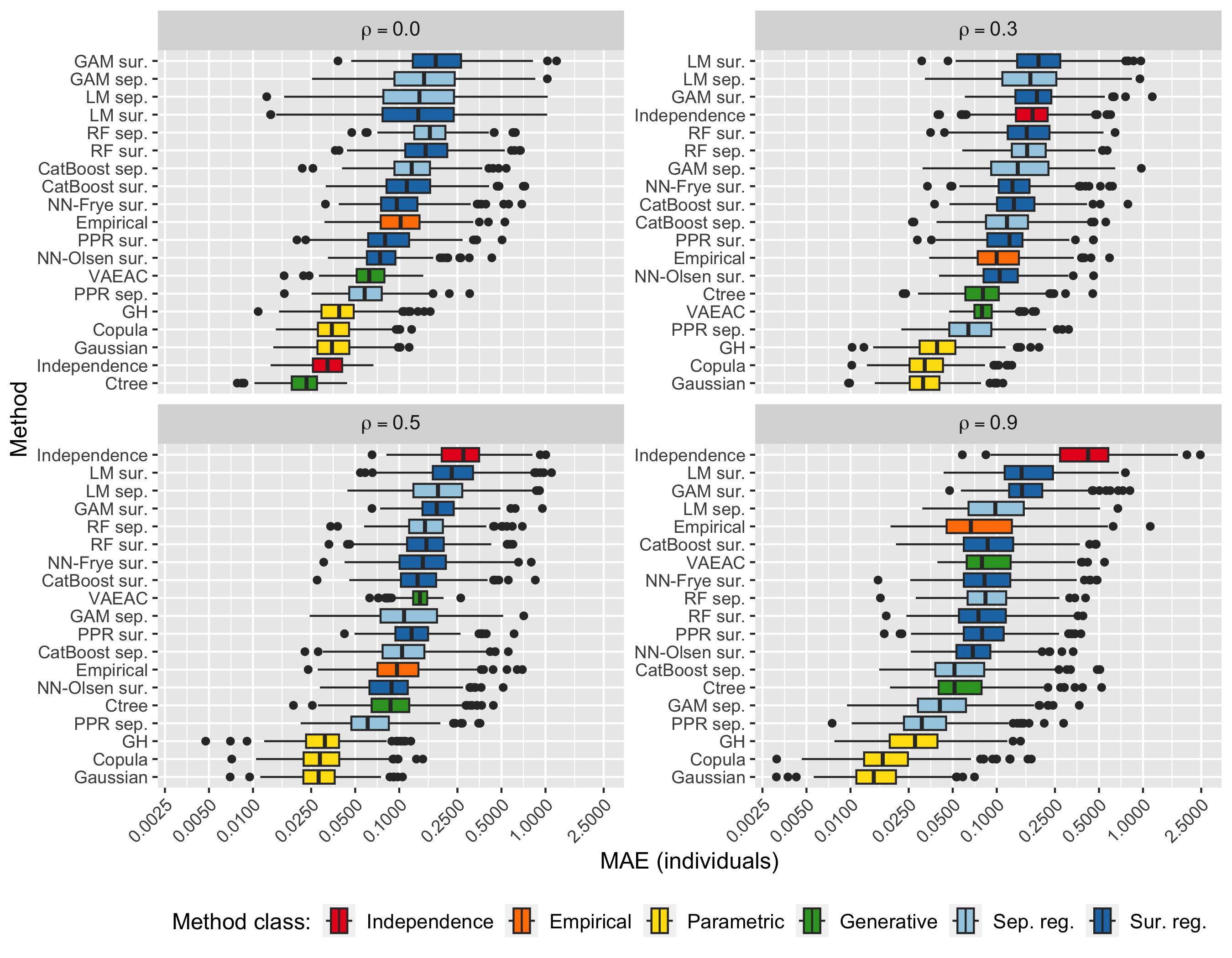}}
    \caption{{\small Results of the \texttt{lm\textunderscore more\textunderscore interactions} experiment; see \Cref{fig:lm_no} for description.}}
    \label{fig:lm_more}
\end{figure}

\begin{figure}[!t]
    \centering
    \centerline{\includegraphics[width=1.0\textwidth]{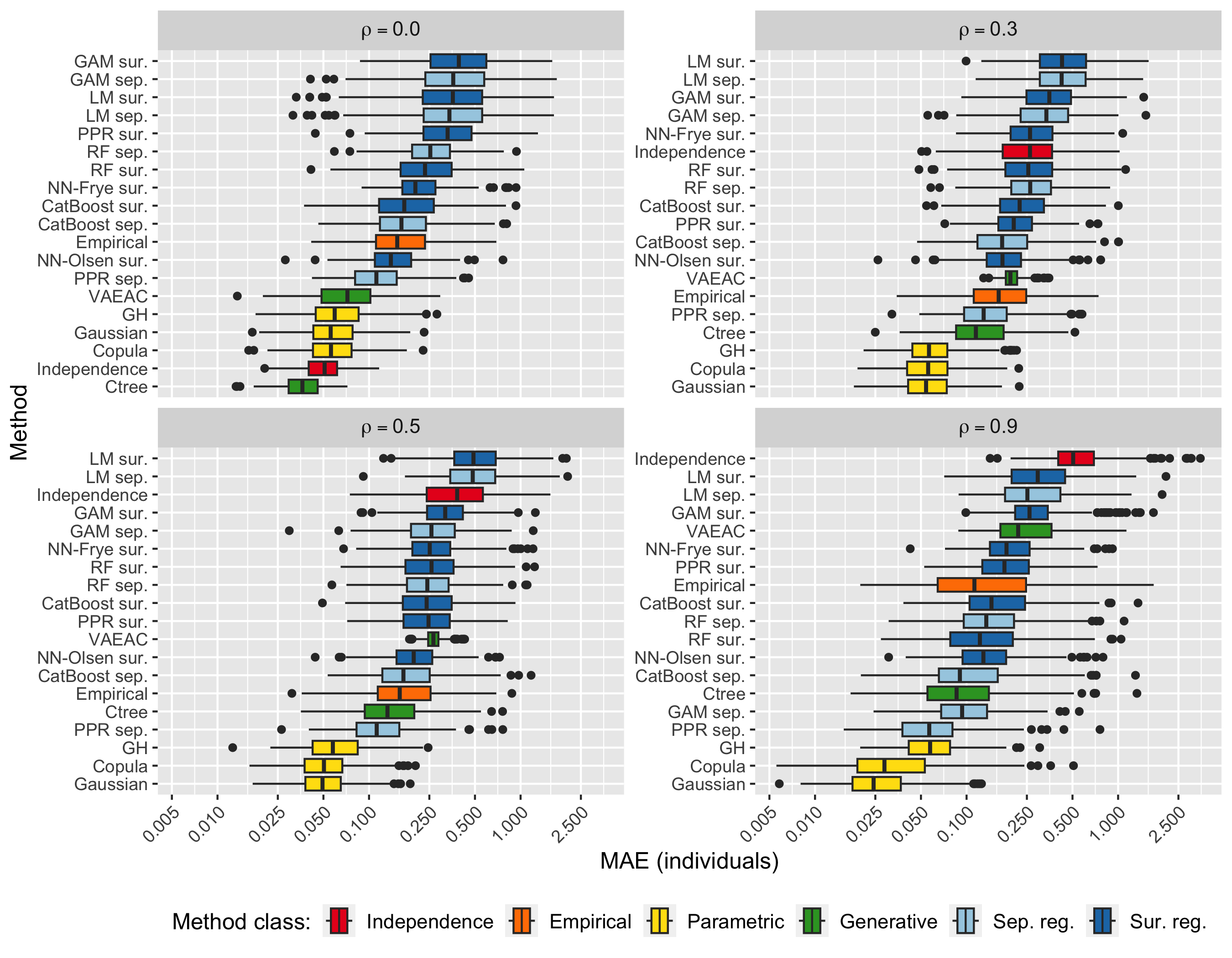}}
    \caption{{\small Results of the \texttt{lm\textunderscore numerous\textunderscore interactions} experiment; see \Cref{fig:lm_no} for description.}}
    \label{fig:lm_numerous}
\end{figure}

In \Cref{fig:lm_no,fig:lm_more,fig:lm_numerous}, we show the MAE for each test observations (i.e., the absolute error averaged only over the features) and the methods are sorted based on the overall MAE (i.e., when also averaged over the test observations). In what follows, we provide a short summary of the results for the different simulation setups.

\begin{enumerate}[align=left, leftmargin=3.5em, itemsep=1pt, topsep=4pt] 
    \item [\texttt{lm\textunderscore no\textunderscore interactions} (\Cref{fig:lm_no}):] For $\rho = 0$, we see that \texttt{ctree} and \tm{LM surrogate} perform the best. The \independence\ approach, which makes the correct feature independence assumption, is close behind. For $\rho > 0$, the \parametric\ and \separate\ (\texttt{LM}, \texttt{GAM}, and \texttt{PPR}) methods generally perform the best. In particular, the \tm{LM separate} method, which makes the correct model assumption, is the best performing approach. The \generative\ and \empirical\ approaches form the mid-field, while the \surrogate\ and \independence\ methods seem to be the least precise.
    
    \item [\texttt{lm\textunderscore more\textunderscore interactions} (\Cref{fig:lm_more}):] In this case, the \tm{LM separate} method performs poorly, which is reasonable due to the incorrect model assumption. For $\rho = 0$, the \ctree\ approach is the most accurate, but the \independence\ and \parametric\ methods are close behind. For $\rho > 0$, the \parametric\ methods are clearly the best approaches as they make the correct parametric assumption. The \texttt{PPR separate} method performs very well, and the \generative\ approaches are almost on par for moderate correlation. The \texttt{NN-Olsen surrogate} method is the most accurate \tm{surrogate} \tm{regression} approach. In general, the \separate\ approaches perform better as $\rho$ increases, in particular the \texttt{GAM separate} approach. 
    
    \item [\texttt{lm\textunderscore numerous\textunderscore interactions} (\Cref{fig:lm_numerous}):] The overall tendencies are very similar to those for \texttt{lm\textunderscore more\textunderscore interactions}. The \parametric\ methods are by far the most accurate. Further, \ctree\ is the best \generative\ approach, the \tm{NN-Olsen surrogate} is the best \surrogate\ method, and the \texttt{PPR separate} method is the best \separate\ approach. 
\end{enumerate}

\subsection{Generalized Additive Models}
\label{subsection:Simulation:GAM}
In this section, we first investigate situations with a nonlinear relationship between the features and the response, while we later also include pairwise nonlinear interaction terms. More specifically, we first gradually progress from the \texttt{lm\textunderscore no\textunderscore interactions} model to a full generalized additive model by applying the nonlinear function $\operatorname{cos}(x_j)$ to a subset of the features in $\x$. Then we extend the full generalized additive model by also including pairwise nonlinear interaction terms of the form $g(x_j,x_k) = x_jx_k + x_jx_k^2 + x_kx_j^2$. We generate the features $\x^{[i]}$ as before, but the response value $y^{[i]}$ is now generated according to:
\begin{enumerate}[align=left, leftmargin=3.5em, itemsep=1pt, topsep=4pt] 
    \item [\texttt{gam\textunderscore three}:] $f_{\gam, \three}(\x) = \beta_0 + \sum_{j=1}^{3}\beta_j\cos(x_j) + \sum_{j=4}^{M} \beta_jx_j$,

    \item [\texttt{gam\textunderscore all}:] $f_{\gam, \all}(\x) = \beta_0 + \sum_{j=1}^{M}\beta_j\cos(x_j)$,

    \item [\texttt{gam\textunderscore more\textunderscore interactions}:] $f_{\gam, \more}(\x) =  f_{\gam, \all}(\x) + \gamma_1g(x_1, x_2) + \gamma_2g(x_3, x_4)$, 

    \item [\texttt{gam\textunderscore numerous\textunderscore interactions}:] $f_{\gam, \numerous}(\x) =  f_{\gam, \more}(\x) + \gamma_3g(x_5, x_6) + \gamma_4g(x_7, x_8)$, 
\end{enumerate}
where $\boldsymbol{\beta} = \set{1.0,  0.2, -0.8, 1.0, 0.5, -0.8, 0.6, -0.7, -0.6}$ and $\boldsymbol{\gamma} = \set{0.8, -1.0, -2.0, 1.5}$, i.e., the same coefficients as in \Cref{subsection:Simulation:lm}. 

As the true models contain smooth nonlinear effects and smooth pairwise nonlinear interaction terms, we let the corresponding predictive models be GAMs with splines for the nonlinear terms and tensor product smooths for the nonlinear interaction terms. E.g., for the \texttt{gam\textunderscore three} experiment, the fitted predictive model $f$ uses splines on the three first features while the others are linear. For the \texttt{gam\textunderscore more\textunderscore interactions} experiment, $f$ uses splines on all eight features and tensor product smooths on the two nonlinear interaction terms. We fit the predictive models using the \texttt{mgcv} package with default parameters \parencite{wood2006low, mgcv}. In what follows, we provide a short summary of the results for the different simulation setups.



\begin{figure}[!t]
    \centering
    \centerline{\includegraphics[width=1.0\textwidth]{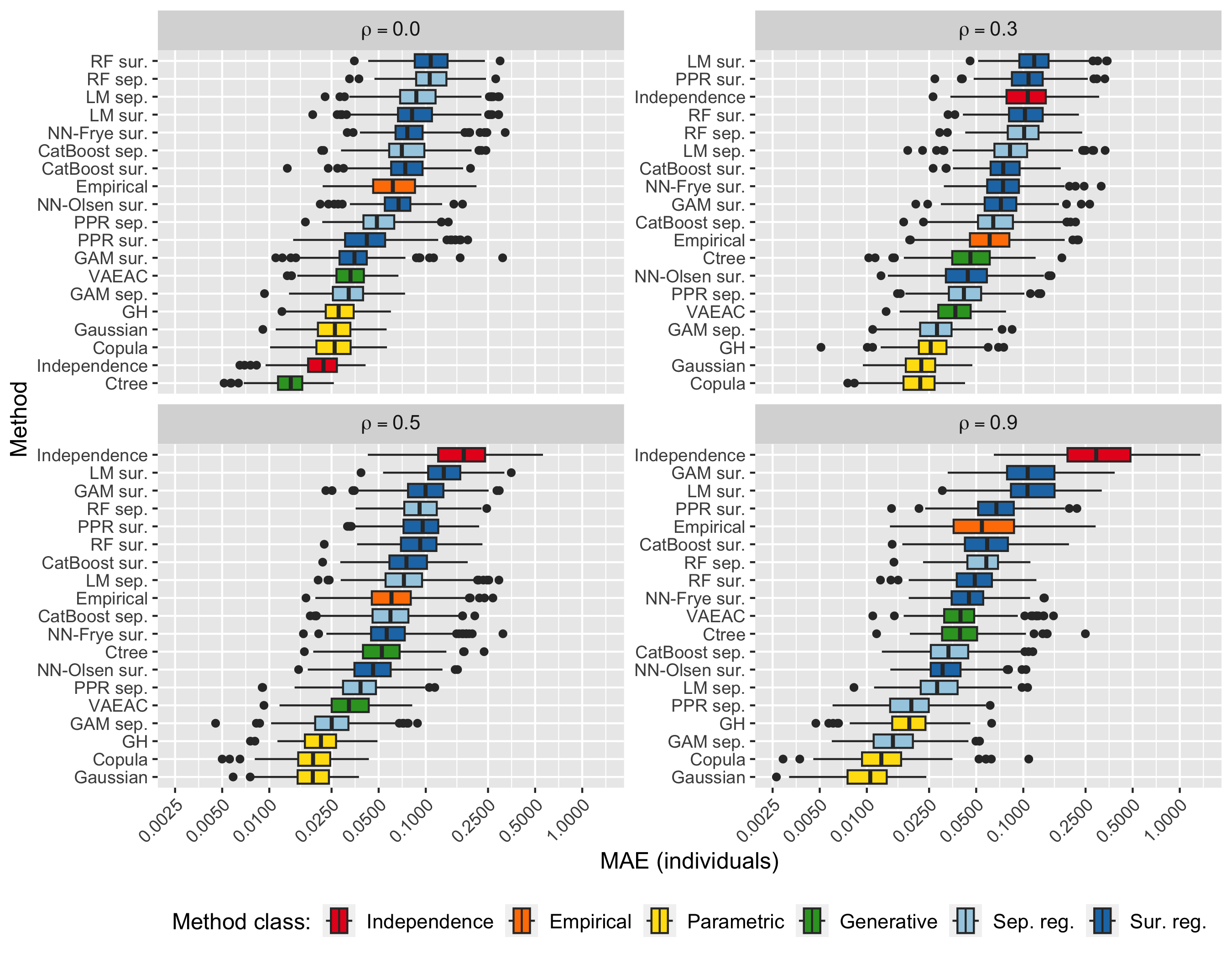}}
    \caption{{\small Results of the \texttt{gam\textunderscore three} experiment; see \Cref{fig:lm_no} for description.}}
    \label{fig:lm_to_gam_three}
\end{figure}

\begin{figure}[!t]
    \centering
    \centerline{\includegraphics[width=1.0\textwidth]{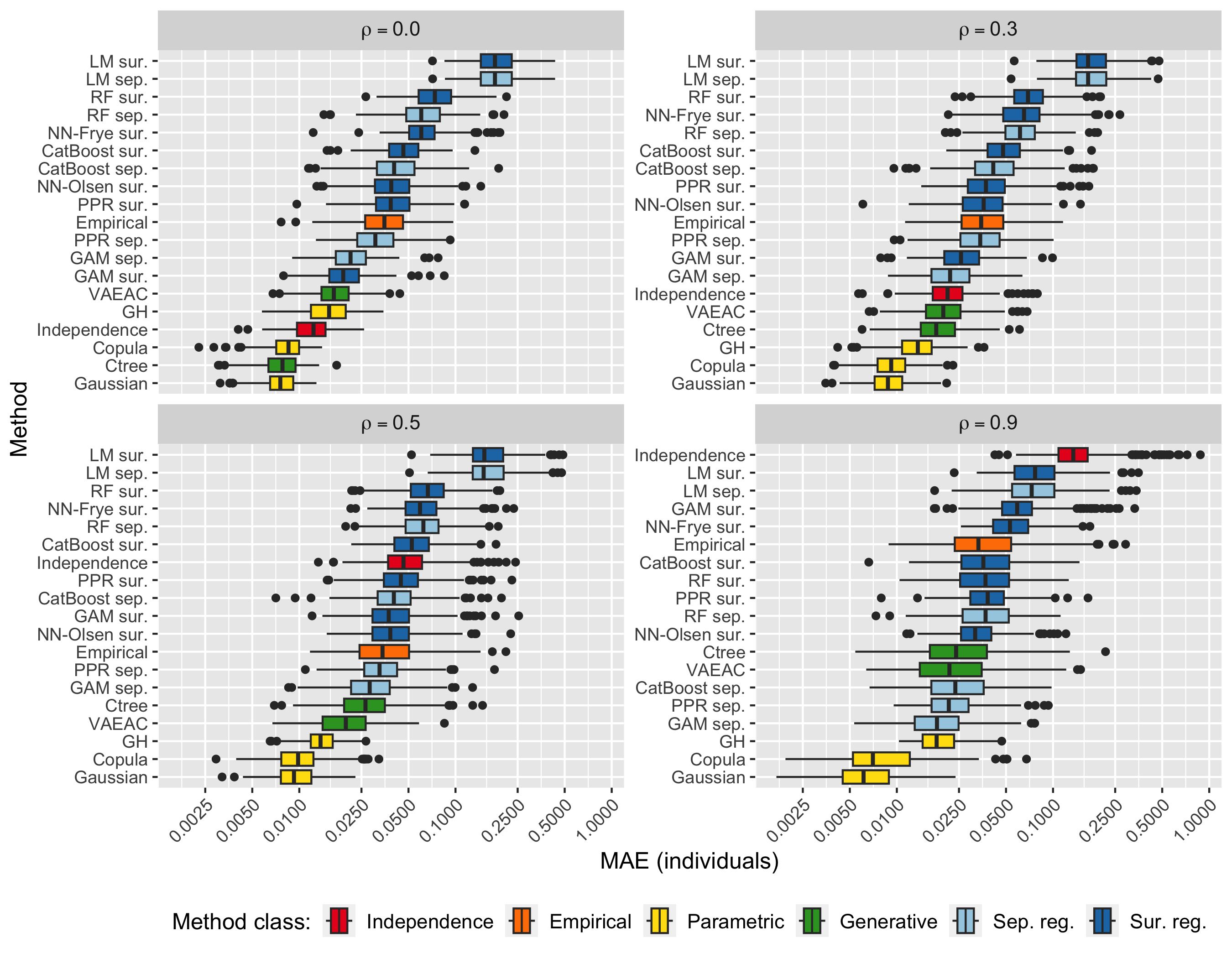}}
    \caption{{\small Results of the \texttt{gam\textunderscore all} experiment; see \Cref{fig:lm_no} for description.}}
    \label{fig:lm_to_gam_all}
\end{figure}

\begin{figure}[!t]
    \centering
    \centerline{\includegraphics[width=1.0\textwidth]{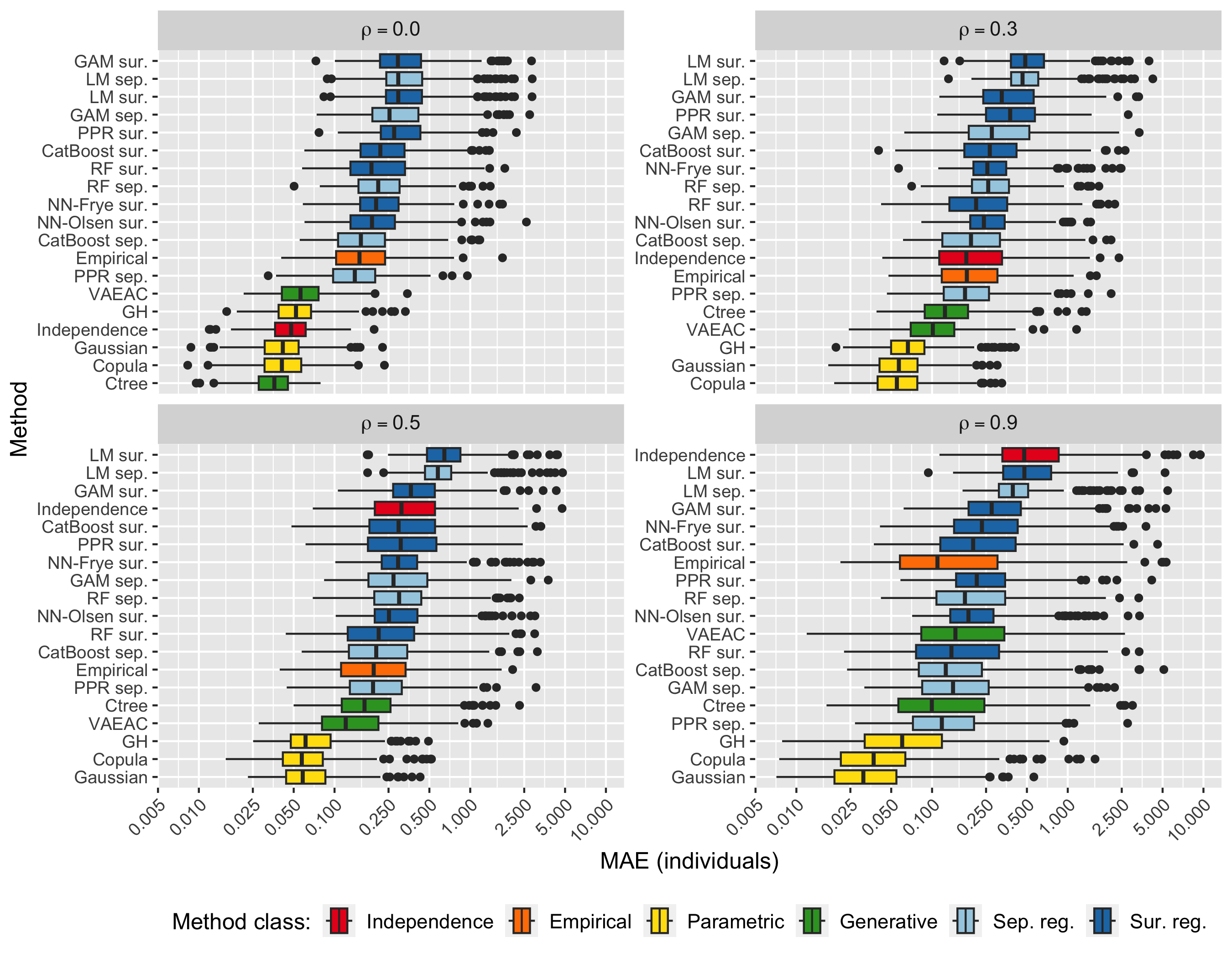}}
    \caption{{\small 
    Results of the \texttt{gam\textunderscore more\textunderscore interactions} experiment; see \Cref{fig:lm_no} for description.}}
    \label{fig:gam_more}
\end{figure}

\begin{figure}[!t]
    \centering
    \centerline{\includegraphics[width=1.0\textwidth]{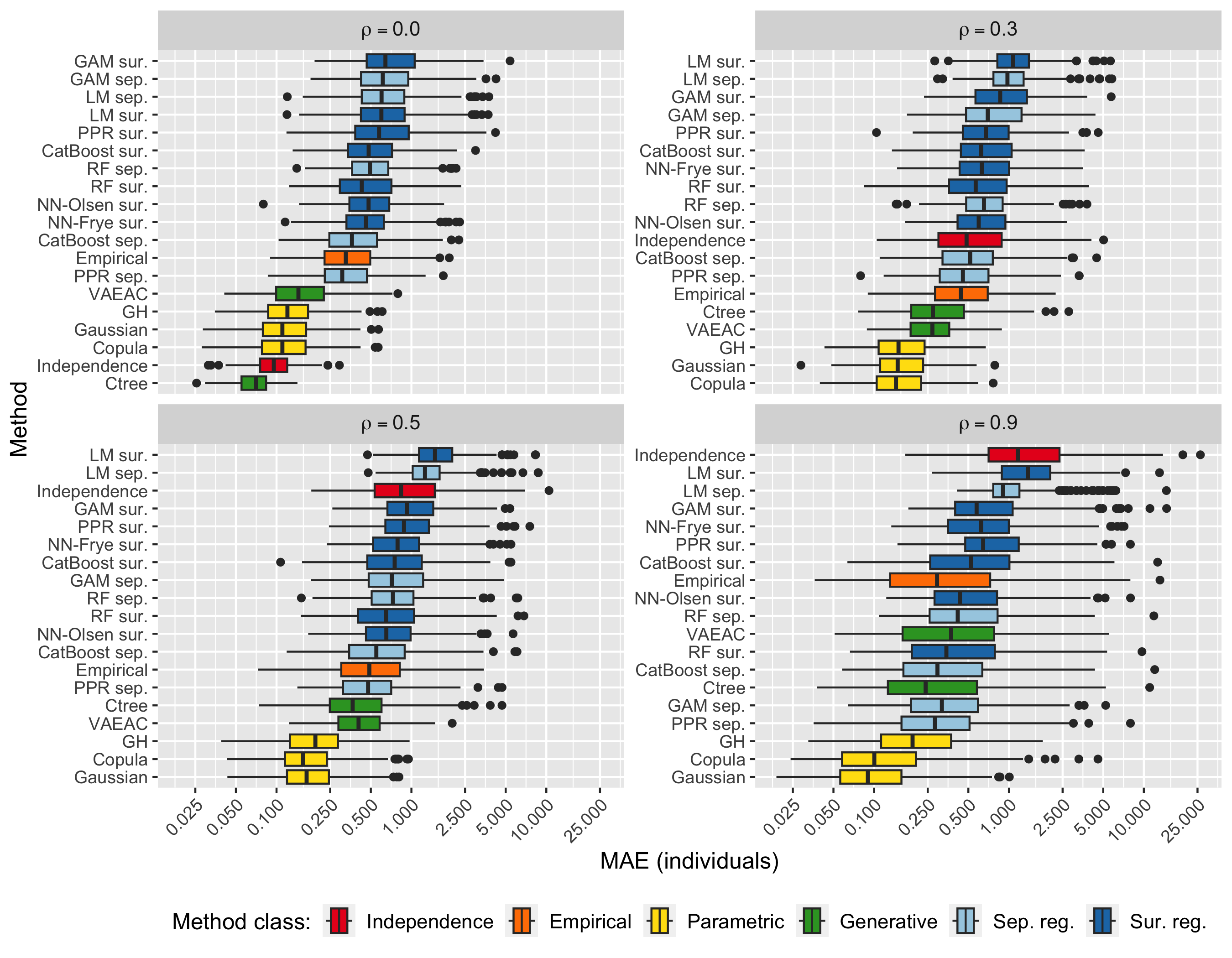}}
    \caption{{\small Results of the \texttt{gam\textunderscore numerous\textunderscore interactions} experiment; see \Cref{fig:lm_no} for description.}}
    \label{fig:gam_numerous}
\end{figure}

\begin{enumerate}[align=left, leftmargin=3.5em, itemsep=1pt, topsep=4pt]
    \item [\texttt{gam\textunderscore three} (\Cref{fig:lm_to_gam_three}):]
    On the contrary to the \texttt{lm\tu no\tu interactions} experiment, we see that the \texttt{LM separate} approach performs much worse than the \texttt{GAM separate} approach, which makes sense as we have moved from a linear to a nonlinear setting. For $\rho = 0$, we see that \ctree\ and \independence\ are the best approaches. For $\rho > 0$, the \parametric\ approaches are superior, but the \texttt{GAM separate} approach is not far behind, while the \texttt{NN-Olsen surrogate} method is the best \surrogate\ approach.
    
    \item [\texttt{gam\textunderscore all} (\Cref{fig:lm_to_gam_all}):] The performance of the \texttt{LM} approaches continue to degenerate. The \separate\ methods get gradually better for higher values of $\rho$, but the \parametric\ methods are still superior. The \generative\ methods constitute the second-best class for $\rho \in \{0.3, 0.5\}$, but the \texttt{GAM separate} and \texttt{PPR separate} approaches are relatively close. The latter approaches outperform the \generative\ methods when $\rho = 0.9$.

    \item [\texttt{gam\textunderscore more\textunderscore interactions} (\Cref{fig:gam_more}):] We see similar results as for the \texttt{gam\textunderscore all} experiment. The \parametric\ approaches are superior in all settings. The \generative\ methods perform quite well for $\rho < 0.5$, but are beaten by the \texttt{PPR separate} method for $\rho = 0.9$. Note that the \texttt{GAM separate} approach now falls behind the \texttt{PPR separate} approach, as it is not complex enough to model the nonlinear interaction terms. This indicates that complex \separatereg\ approaches are needed to model complex predictive models. Furthermore, the \texttt{RF surrogate} method is on par or outperforms the \texttt{NN} based \surrogate\ approaches.
     
    \item [\texttt{gam\textunderscore numerous\textunderscore interactions} (\Cref{fig:gam_numerous}):] We get nearly identical results as in the previous experiment. Hence, we do not provide further comments to the results.
\end{enumerate}

\vspace{-1em}
\subsection{Computation Time}
\label{Numerical_Simulation_Studies:Time}

\begin{table}[!t]
\centering
\begin{adjustbox}{max width=1\textwidth}
\begin{tabular}{lrrrrr}
  \toprule
Method & Training & Generating $\x_\s^{(k)}$ & Predicting $v(\s)$ & Total CPU Time \\ 
  \midrule
  \rowcolor{col_ind!10} \tm{Independence} & 4.5 & 20.1 & 35:16.8 & 35:41.4 \\ 
  \rowcolor{col_emp!10} \tm{Empirical} & 4.4 & 19.0 & 8:47.6 & 9:11.0 \\ 
  \rowcolor{col_par!10} \tm{Gaussian} & 0.0 & 1:22.7 & 35:36.3 & 36:59.0 \\ 
  \rowcolor{col_par!10} \tm{Copula} & 0.0 & 5:53.5 & 35:18.0 & 41:11.5 \\ 
  \rowcolor{col_par!10} \tm{GH} & 2:05.3 & 1:34.7 & 36:15.8 & 39:55.8 \\ 
  \rowcolor{col_gen!10} \tm{Ctree} & 3.2 & 3:47.1 & 10:31.1 & 14:21.4 \\ 
  \rowcolor{col_gen!10} \tm{VAEAC} & 56.3 & 3:07.7 & 34:41.1 & 38:45.1 \\ 
  \rowcolor{col_sep!10} \tm{LM sep.} & 0.3 & --- & 0.2 & 0.5 \\ 
  \rowcolor{col_sep!10} \tm{GAM sep.} & 34.6 & --- & 4.8 & 39.4 \\ 
  \rowcolor{col_sep!10} \tm{PPR sep.} & 1:39.2 & --- & 0.3 & 1:39.5 \\ 
  \rowcolor{col_sep!10} \tm{RF sep.} & 58:46.0 & --- & 5.3 & 58:51.3 \\ 
  \rowcolor{col_sep!10} \tm{CatBoost sep.} & 5:44.8 & --- & 0.1 & 5:44.9 \\ 
  \rowcolor{col_sur!10} \tm{LM sur.} & 2.3 & --- & 0.4 & 2.7 \\ 
  \rowcolor{col_sur!10} \tm{GAM sur.} & 12.5 & --- & 8.6 & 21.1 \\ 
  \rowcolor{col_sur!10} \tm{PPR sur.} & 3:49.5 & --- & 0.5 & 3:50.0 \\ 
  \rowcolor{col_sur!10} \tm{RF sur.} & 1:05:55.5 & --- & 7.9 & 1:06:03.4 \\ 
  \rowcolor{col_sur!10} \tm{CatBoost sur.} & 38.8 & --- & 0.4 & 39.2 \\ 
  \rowcolor{col_sur!10} \tm{NN-Frye sur.} & 13:56:43.9 & --- & 1.8 & 13:56:45.7 \\ 
  \rowcolor{col_sur!10} \tm{NN-Olsen sur.} & 7:31:43.8 & --- & 1.9 & 7:31:45.7 \\ 
   \bottomrule
\end{tabular}
\end{adjustbox}
\caption{{\small The CPU times used by the methods to compute Shapley values for the $N_\text{test} = 250$ test observations in the \texttt{gam\tu more\tu interactions} experiment with $\rho = 0.5$ and $N_\text{train} = 1000$. The format of the CPU times is hours:minutes:seconds, where we omit the larger units of time if they are zero, and the colors indicate the different method classes.}}
\label{tab:SimStudyTimeCol}
\end{table}

In this section, we discuss the computation time used by the different methods to estimate the Shapley values, as a proper evaluation of the methods should not only be limited to their accuracy. We report the CPU times to get a fair comparison between the approaches, as some methods are parallelized and would therefore benefit from multiple cores when it comes to elapsed time. The CPU times for the different methods will vary significantly depending on operating system, hardware, and implementation. 
The times we report here are based on an Intel(R) Core(TM) i5-1038NG7 CPU@2.00GHz with 16GB 3733MHz LPDDR4X RAM running R version $4.2.0$ on the macOS Ventura (13.0.1) operating system. Throughout this article, we mean CPU time when we discuss time. 

In \Cref{tab:SimStudyTimeCol}, we report the time it took to estimate the Shapley values using the different methods for the \texttt{gam\tu more\tu interactions} experiment with $\rho = 0.5$, $N_\text{train} = 1000$, and $N_\text{test} = 250$ in \Cref{subsection:Simulation:GAM}. We split the total time into three time categories: time used training the approaches, time used generating the Monte Carlo samples, and time used predicting the $v(\s)$ using Monte Carlo integration (including the calls to $f$) or regression. We denote these three categories by: \textit{training}, \textit{generating}, and \textit{predicting}, respectively. The matrix multiplication needed to estimate the Shapley values from the estimated contribution functions is almost instantaneous and is part of the predicting time. Furthermore, creating the augmented training data for the \surrogate\ methods in \Cref{ConditionalShapleyValues:SurrogateModel:NewMethods} takes around one second and is part of the training time. We see a wide spread in the times, but, in general, the Monte Carlo approaches take on average around half an hour, while the regression methods are either much faster or slower, depending on the approach. 

The Monte Carlo methods make a total of $N_\text{test}K(2^M-2)$ calls to the predictive model $f$ to explain the $N_\text{test}$ test observations with $M$ features and $K$ Monte Carlo samples. In our setting with $N_\text{test} = 250$, $M=8$, and $K = 250$, the predictive model is called $N_f = 15\,875\,000$ times, thus, the speed of calculating $f$ greatly effects the explanation time. For example the GAM model in the \texttt{gam\tu more\tu interactions} experiment is slow, as we can see in \Cref{tab:SimStudyTimeCol}, since the predicting time constitutes the majority of the total time. To compare, $N_f$ calls to the linear model in the \texttt{lm\tu more\tu interactions} experiment takes approximately $3$ CPU seconds, while the GAMs in the \texttt{gam\tu three} and \texttt{gam\tu more\tu interactions} experiments take roughly $13$ and $35$ CPU minutes, respectively. For the latter experiment, the PPR and RF models in \Cref{Numerical_Simulation_Studies:OtherPredictiveModels} take around $0.5$ and $40$ CPU minutes, respectively. 

The training and generating times of the \texttt{independence} approach are higher than expected, but this is because the \texttt{independence} method is implemented as a special case of the \texttt{empirical} approach in version $0.2.0$ of the \texttt{shapr}-package. Furthermore, the \empirical\ and \ctree\ approaches have lower predicting time than the other Monte Carlo-based methods due to fewer calls to $f$ since they use weighted Monte Carlo samples; see \Cref{subsec:ConditionalShapleyValues:empirical,subsec:ConditionalShapleyValues:GenerativeModel:Ctree}. The three influential time factors for the Monte Carlo methods are: the training time of the approach (estimating the parameters), the sampling time of the Monte Carlo samples, and the computational cost of calling $f$.

In contrast, both the \separate\ and \surrogate\ methods use roughly the same time to estimate the Shapley values for different predictive models $f$, as $f$ is only called $N_\text{train}$ times when creating the training data sets. After that, we train the \separate\ and \surrogate\ approaches and use them to directly estimate the contribution functions. The influential factors for the regression methods are: the training time of the $2^M-2$ separate models (or the one surrogate model) and the prediction time of calling them a total of $N_\text{test}(2^M-2)$ times. The former is the primary factor, and it is influenced by, e.g., hyperparameter tuning and the training data size. The latter can be a problem for the augmented training data for the \surrogate\ methods, as we will see in \Cref{real_world_data:Adult}. The \texttt{NN} approaches are the slowest methods and the training time is the cause. We can reduce the training time, at the cost of precision, by using default values instead of conducting cross-validation. This would approximately reduce the time by a factor of six and nine for the \tm{NN-Frye surrogate} and \tm{NN-Olsen surrogate} methods, respectively.

When excluding the time of the training step, which is only done once and can be considered as an upfront time cost, it is evident that the regression-based methods produce the Shapley value explanations considerably faster then the Monte Carlo-based methods. For example, consider the most accurate Monte Carlo and regression-based methods in the \texttt{gam\tu more\tu interactions} experiment with $\rho = 0.5$, i.e., the \Gaussian\ and \tm{PPR separate} methods, respectively. The \Gaussian\ approach uses approximately $37$ CPU minutes to explain $250$ predictions, an average of $8.88$ seconds per explanation. In contrast, the \tm{PPR separate} method explains all the $N_\text{test} = 250$ predictions in half a second. Thus, the \tm{PPR separate} method is approximately $4440$ times faster than the \Gaussian\ approach per explanation, which is essential for large values of $N_\text{test}$. However, note that this factor is lower for predictive models that are less computationally expensive to call.

\subsection{Number of Training Observations}
\label{Numerical_Simulation_Studies:TrainingSizes}
We repeated the experiments in \Cref{subsection:Simulation:lm,subsection:Simulation:GAM} with $N_\text{train} \in \set{100, 5000}$, and some of them with $N_\text{train} = 20\,000$, to investigate if the MAE based ordering of the methods depends on the training data size. We obtained nearly identical results, except for three distinctions. First, the \independence\ approach became relatively more accurate compared to the other methods when $N_\text{train} = 100$, and worse when $N_\text{train} \in \set{5000, 20\,000}$. This is intuitive, as modeling the data distribution/response is easier when the methods have access to more data. Second, for the simple experiments in \Cref{subsection:Simulation:lm,subsection:Simulation:GAM} and $N_\text{train} \in \set{5000, 20\,000}$, the \texttt{GAM separate} and \texttt{PPR separate} approaches became even better, but were still beaten by the \Gaussian\ and \copula\ approaches in most experiments. Third, we observed that the MAE had a tendency to decrease when $N_\text{train}$ increased. However, we cannot directly compare the MAE scores as they depend on the fitted predictive model $f$ which changes when $N_\text{train}$ is adjusted. 

\subsection{Other Choices for the Predictive Model}
\label{Numerical_Simulation_Studies:OtherPredictiveModels}
In practice, it might be difficult to identify the pairwise interactions in \Cref{subsection:Simulation:GAM}. Hence, one would potentially fit a model without them. We included them above as we knew the data generating processes and wanted a precise model, but we now pretend otherwise and fit other predictive models. We consider two different types of complex black-box predictive models: projection pursuit regression (PPR) and random forest (RF), and we conduct the same type of hyperparameter tuning as for the other experiments. However, we conduct no feature transformations and directly use the original features when fitting the models. 
These models are less precise than the GAMs in \Cref{subsection:Simulation:GAM}, which have an unfair advantage as they use the true formulas. E.g., for the \texttt{gam\textunderscore more\textunderscore interactions} experiment, the test prediction MSE was $1.32, 3.67, 7.36$ for GAM, PPR, RF, respectively, where $1$ is the theoretical optimum as $\Var(\varepsilon) = 1$. 

{
\begin{figure}[!t]
    \centering
    \centerline{\includegraphics[width=1\textwidth]{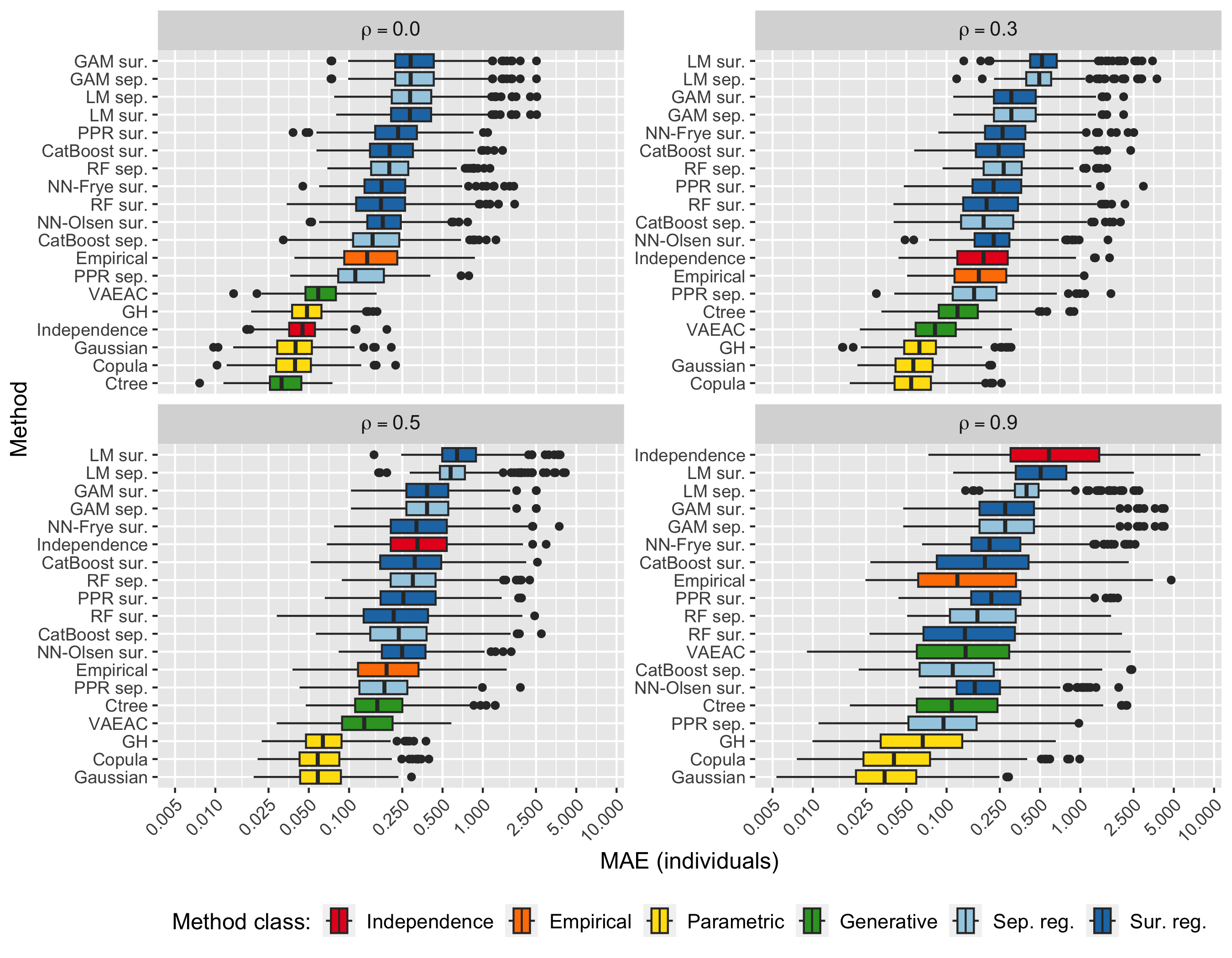}}
    \caption{{\small Experiment \texttt{gam\textunderscore more\textunderscore interactions} with a PPR model; see \Cref{fig:lm_no} for description.}}
    \label{fig:gam_more_ppr}
\end{figure}

\begin{figure}[!t]
    \centering
    \centerline{\includegraphics[width=1\textwidth]{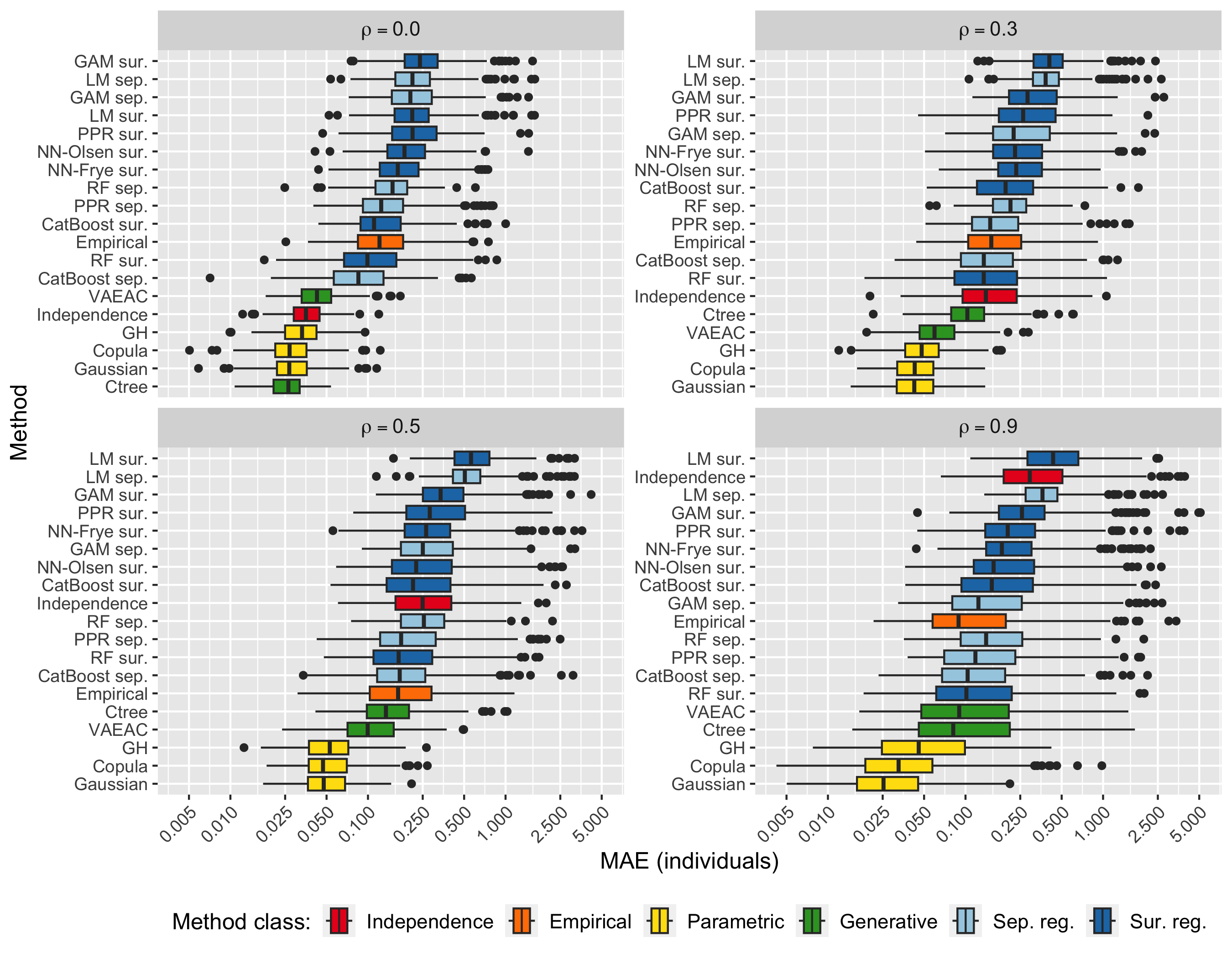}}
    \caption{{\small Experiment \texttt{gam\textunderscore more\textunderscore interactions} with a RF model; see \Cref{fig:lm_no} for description.}}
    \label{fig:gam_more_rf}
\end{figure}
}



We only include the figures for the \texttt{gam\textunderscore more\textunderscore interactions} experiment, as the corresponding figures for the other experiments are almost identical. The results are displayed in \Cref{fig:gam_more_ppr,fig:gam_more_rf} for the PPR and RF models, respectively, and the results are quite similar to those obtained for the GAM model in \Cref{fig:gam_more}. In general, the \parametric\ methods are superior, followed by the \generative\ methods, while the \empirical, \separate, and \surrogate\ approaches are worse. Some \separate\ approaches perform however much better for high dependence. The \independence\ method performs well when $\rho = 0$, but gradually degenerates as the dependence level increases, as expected. We see that the \texttt{PPR separate} approach performs well for the PPR predictive model, but it is outperformed by the \texttt{CatBoost separate} method for the RF models. These results indicate that for our experiments, it is beneficial to choose a regression method similar to the predictive model; that is, for a non-smooth model, one should consider using a non-smooth regression method. However, note that the difference in the MAE is minuscule.

\subsection{Different Data Distribution}
\label{Numerical_Simulation_Studies:DataDistribution}
In \Cref{Appendix:ExtendedSimulations:Intermediate_burr}, we repeat all the experiments described in \Cref{subsection:Simulation:lm,subsection:Simulation:GAM} but with multivariate Burr-distributed features instead of Gaussian ones. 
The Burr distribution allows for heavy-tailed, skewed marginals, and nonlinear dependence. In this case, the \parametric\ \texttt{Burr} approach, which assumes Burr distributed data, not surprisingly, is the most accurate. The \Gaussian\ method, which now incorrectly assumes Gaussian distributed data, performs worse. The \vaeac\ approach performs very well on the Burr distributed data, which was also observed by \textcite{Olsen2022}. In general, \vaeac\ is the second-best approach after \texttt{Burr}. The \texttt{PPR separate} method also performs well, but compared to the \texttt{Burr} and \vaeac\ approaches, it is less precise in the experiments with nonlinear interaction terms.

\subsection{Summary of the Experiments}
\label{Numerical_Simulation_Studies:Summary}
Making the correct (or nearly correct) parametric assumption about the data is advantageous, as the corresponding \parametric\ methods significantly outperform the other approaches in most settings. In general, if the distribution is unknown, the second-best option for low to moderate levels of dependence is the \generative\ methods. The \separatereg\ approaches improve relative to the other methods when the feature dependence increases, and for highly dependent features, the \texttt{PPR separate} approach is a prime choice. Furthermore, the \separatereg\ methods which match the form of $f$ often give more accurate Shapley value estimates. The PPR model in the \texttt{PPR separate} approach is simple to fit but is still very flexible and can, therefore, accurately model complex predictive models. The \independence\ approach is accurate for no (or very low) feature dependence, but it is often the worst approach for high feature dependence. The \tm{NN-Olsen surrogate} method outperforms the \tm{NN-Frye surrogate} approach in most settings and is generally the best \surrogate\ approach. 


We found it (often) necessary to conduct some form of cross-validation to tune (most of) the \separate\ and \surrogate\ methods to make them more competitive. Using default hyperparameter values usually resulted in less accurate Shapley value explanations; see \Cref{Appendix:ExtendedSimulations}. The hyperparameter tuning can be time-consuming, but it was feasible in our setting with $M=8$ features and $N_\text{train} = 1000$ training observations. The regression-based methods use most of their computation time on training, while the predicting step is almost instantaneous for several methods. The opposite holds for the Monte Carlo-based approaches, which are overall slower than most regression-based methods. Hence, we have a trade-off between computation time and Shapley value accuracy in the numerical simulation studies. We did not conduct hyperparameter tuning for the \empirical, \parametric, and \generative\ methods. Thus, the methods where we conduct hyperparameter tuning have an unfair advantage regarding the precision of the estimated Shapley values.

\section{Real-World Data Experiments}
\label{sec:real_world_data}

In this section, we fit several predictive models to different real-world data sets from the UCI Machine Learning Repository (\url{https://archive.ics.uci.edu/ml/datasets.php}) and then use Shapley values to explain the models' predictions. The models range from complex statistical models to black-box machine learning methods. We consider four data sets: Abalone, Diabetes, Wine, and Adult. Some illustrative data plots are provided in \Cref{Appendix:PairPlots}.

For real-world data sets, the true Shapley values are unknown. Hence, we cannot use the MAE evaluation criterion from \Cref{sec:Numerical_Simulation_Studies} to evaluate and rank the approaches. Instead, we use the $\operatorname{MSE}_{v}$ criterion proposed by \textcite{frye_shapley-based_2020} and later used by \textcite{Olsen2022}. The $\operatorname{MSE}_{v}$ is given by
\begin{align}
    \label{eq:MSE_v}
    \operatorname{MSE}_{v} = \operatorname{MSE}_{v}(\text{method } \texttt{q}) 
    =
     \frac{1}{N_\mathcal{S}} \sum_{\s \in \pow^*(\mathcal{M})} \frac{1}{N_\text{test}} \sum_{i=1}^{N_\text{test}} \left( f(\boldsymbol{x}^{[i]}) - \hat{v}_{\texttt{q}}(\s, \boldsymbol{x}^{[i]})\right)^2\!,
\end{align}
where $N_\mathcal{S} = |\pow^*(\mathcal{M})| = 2^M-2$ and $\hat{v}_{\texttt{q}}$ is the estimated contribution function using method $\texttt{q}$. The motivation behind the $\operatorname{MSE}_{v}$ criterion is that $\E_\s\E_{\x} (v_{\texttt{true}}(\s, \boldsymbol{x}) - \hat{v}_{\texttt{q}}(\s, \boldsymbol{x}))^2$ can be decomposed as
\begin{align}
    \label{eq:expectation_decomposition}
    \E_\s\E_{\x} (v_{\texttt{true}}(\s, \boldsymbol{x})- \hat{v}_{\texttt{q}}(\s, \boldsymbol{x}))^2 
    =
    \E_\s\E_{\x} (f(\x) - \hat{v}_{\texttt{q}}(\s, \boldsymbol{x}))^2 - \E_\s\E_{\x} (f(\x)-v_{\texttt{true}}(\s, \boldsymbol{x}))^2,
\end{align}
see \textcite[Appendix A]{covert2020understanding}. The first term on the right-hand side of \eqref{eq:expectation_decomposition} can be estimated by \eqref{eq:MSE_v}, while the second term is a fixed (unknown) constant not influenced by the approach \texttt{q}. Thus, a low value of \eqref{eq:MSE_v} indicates that the estimated contribution function $\hat{v}_{\texttt{q}}$ is closer to the true counterpart $v_{\texttt{true}}$ than a high value.

\begin{figure}[!t]
    \centering
    \centerline{\includegraphics[width=1\textwidth]{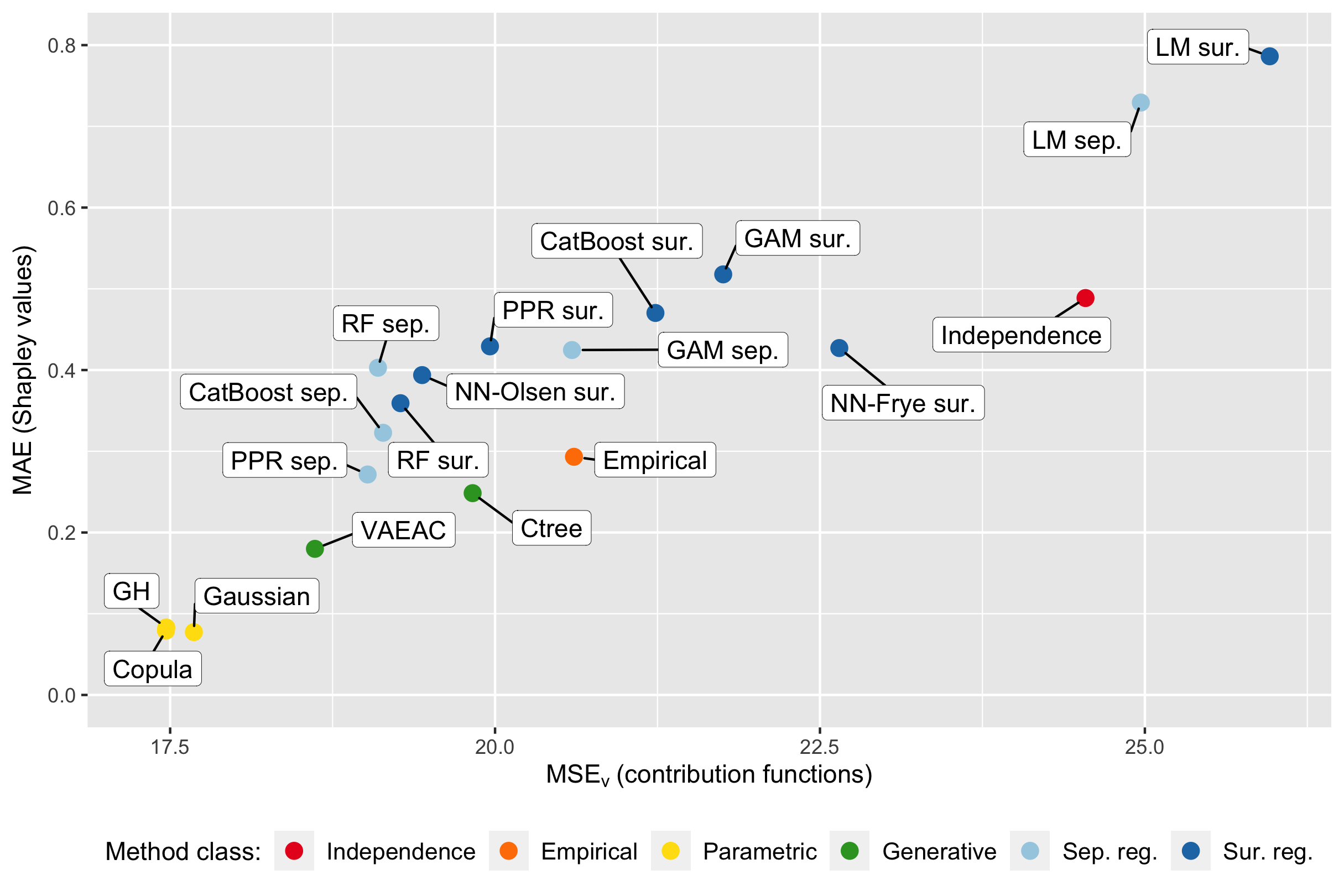}}
    \caption{{\small Figure illustrating a fairly strong linear relationship between the $\operatorname{MSE}_{v}$ and $\operatorname{MAE}$ criteria for the \texttt{gam\textunderscore more\tu interactions} experiment with Gaussian data and $\rho = 0.5$.}}
    \label{fig:MAE_vs_MSE:gam_more_interactions}
\end{figure}

An advantage of the $\operatorname{MSE}_{v}$ criterion is that $v_\texttt{true}$ is not involved. Thus, we can apply it to real-world data sets. However, the criterion has two drawbacks. First, we can only use it to rank the methods and not assess their closeness to the optimum since the minimum value of the $\operatorname{MSE}_{v}$ criterion is unknown. Second, the criterion evaluates the contribution functions and not the Shapley values. It might be the case that the estimates for $v(\s)$ overshoot for some coalitions and undershoot for others, and such errors may cancel each other out in the Shapley value formula in \eqref{eq:ShapleyValuesDef}. Nevertheless, for the numerical simulation studies in \Cref{sec:Numerical_Simulation_Studies}, we computed both criteria to compare the ordering of two criteria empirically. We generally observe a relatively linear relationship between the $\operatorname{MAE}$ and $\operatorname{MSE}_{v}$ criteria. That is, a method that achieves a low $\operatorname{MSE}_{v}$ score also tends to obtain a low $\operatorname{MAE}$ score, and vice versa. To illustrates this tendency, we include \Cref{fig:MAE_vs_MSE:gam_more_interactions}, where we plot the $\operatorname{MSE}_{v}$ criterion against the $\operatorname{MAE}$ criterion for the \texttt{gam\textunderscore more\tu interactions} experiment with Gaussian distributed data with $\rho = 0.5$. Note that the orderings of the two criteria are not one-to-one, but they give very similar rankings of the methods. 

In \Cref{tab:real-world-data}, we report the $\operatorname{MSE}_{v}$ scores and CPU times of the different methods for the four data sets. 
The Abalone, Diabetes, and Wine data sets were run on the same system as specified in \Cref{Numerical_Simulation_Studies:Time}, while the Adult data set was run on a shared computer server running Red Hat Enterprise Linux 8.5 with two Intel(R) Xeon(R) Gold 6226R CPU@2.90GHz (16 cores, 32 threads each) and 768GB DDR4 RAM, due to memory constraints on the former system. Thus, one should not compare the CPU times across these systems but only the CPU times of the different methods within the same experiment. 
More detailed decomposition of the CPU times and additional methods are provided in \Cref{Appendix:RealWorld}.

\begin{table}[!t]
\centering
\small
\sisetup{detect-weight,mode=text}
\addtolength{\tabcolsep}{-2pt}
\centerline{
\begin{adjustbox}{max width=1.175\textwidth}
\begin{tabular}{lrrcrrcrrcrrcrr}
\toprule
& \multicolumn{2}{c}{Abalone$_\text{cont}$ ($M$=7)} && \multicolumn{2}{c}{Abalone$_\text{all}$ ($M$=8)} && \multicolumn{2}{c}{Diabetes ($M$=10)} && \multicolumn{2}{c}{Wine ($M$=11)} && \multicolumn{2}{c}{Adult ($M$=14)}  \\
\cmidrule{2-3} \cmidrule{5-6} \cmidrule{8-9} \cmidrule{11-12} \cmidrule{14-15}
\rowcolor{white} Method & $\operatorname{MSE}_{v}$ & Time & & $\operatorname{MSE}_{v}$ & Time & & $\operatorname{MSE}_{v}$ & Time & & $\operatorname{MSE}_{v}$ & Time & & $\operatorname{MSE}_{v}$ & Time \\
\midrule 
\rowcolor{col_ind!10} \tm{Independence}              & 8.672 &       3:58.1 &&   --- &          --- && 0.203 &      52.3 && 0.145 &    3:56:55.9 &&   --- &           --- \\
\rowcolor{col_ind!10} \tm{Independence}$^{*}$        & 8.679 &       1:24.2 && 9.144 &       3:51.5 && 0.196 &      38.4 && 0.145 &    4:00:53.5 && 0.041 &     1:10:36.3 \\
\rowcolor{col_emp!10} \tm{Empirical}                 & 1.540 &       3:43.2 &&   --- &          --- && 0.143 &      15.1 && 0.088 &    2:33:18.5 &&   --- &           --- \\
\rowcolor{col_par!10} \tm{Gaussian}                  & 1.349 &       3:44.0 &&   --- &          --- && 0.127 &    2:35.1 && 0.118 &    4:08:28.6 &&   --- &           --- \\
\rowcolor{col_par!10} \tm{Copula}                    & 1.223 &      15:05.5 &&   --- &          --- && 0.127 &   10:54.5 && 0.107 &    4:53:36.6 &&   --- &           --- \\
\rowcolor{col_par!10} \tm{GH}                        & 1.292 &       8:39.7 &&   --- &          --- && 0.133 &    7:31.3 && 0.109 &    4:23:39.9 &&   --- &           --- \\
\rowcolor{col_par!10} \tm{Burr}                      & 5.640 &       5:22.3 &&   --- &          --- &&   --- &     ---   && 0.202 &    3:52:32.7 &&   --- &           --- \\
\rowcolor{col_gen!10} \tm{Ctree}                     & 1.393 &       7:40.8 && 1.424 &      19:14.1 && 0.158 &    6:46.9 && 0.102 &    1:02:41.5 &&   --- &           --- \\
\rowcolor{col_gen!10} \tm{VAEAC}                     & 1.182 &    2:34:03.6 &&\B1.180&   11:48:22.1 && 0.128 &   21:57.8 && 0.093 &    6:09:37.9 && 0.027 &  5:12:31:03.7 \\
\rowcolor{col_sep!10} \tm{LM sep.}                   & 1.684 &          0.3 && 1.581 &          0.9 &&\B0.126&       1.9 && 0.146 &          4.6 && 0.043 &     9:00:13.5 \\
\rowcolor{col_sep!10} \tm{GAM sep.}                  & 1.298 &         33.7 && 1.299 &       1:09.7 &&\B0.126&    1:03.6 && 0.124 &       3:24.6 && 0.033 &     2:38:52.6 \\
\rowcolor{col_sep!10} \tm{PPR sep.}                  &\B1.169&       2:15.1 && 1.185 &       3:34.6 &&\B0.126&    5:22.4 && 0.129 &      25:19.4 && 0.032 & 14:12:19:47.6 \\
\rowcolor{col_sep!10} \tm{RF sep.}                   & 1.239 &    1:09:15.8 && 1.259 &    2:31:09.7 && 0.143 & 1:00:23.6 &&\B0.071&    9:21:12.9 && 0.027 & 98:13:33:27.4 \\
\rowcolor{col_sep!10} \tm{CatBoost sep.}             & 1.190 &       6:17.1 && 1.213 &      18:25.0 && 0.135 &   18:40.6 && 0.082 &    1:41:32.7 &&\B0.026& 35:09:59:59.1 \\
\rowcolor{col_sur!10} \tm{LM sur.}                   & 2.912 &          3.7 && 2.770 &          7.2 && 0.165 &       4.4 && 0.162 &         25.9 &&   --- &           --- \\
\rowcolor{col_sur!10} \tm{GAM sur.}                  & 2.611 &         41.1 && 2.557 &       1:24.6 && 0.168 &      21.3 && 0.145 &       4:01.4 &&   --- &           --- \\
\rowcolor{col_sur!10} \tm{PPR sur.}                  & 1.548 &      14:57.5 && 1.538 &      55:31.6 && 0.136 &    4:13.0 && 0.149 &    1:20:15.5 &&   --- &           --- \\
\rowcolor{col_sur!10} \tm{RF sur.}                   & 1.281 &    1:14:30.8 && 1.311 &    3:46:34.1 && 0.143 & 1:33:51.3 && 0.085 & 1:15:42:46.3 &&   --- &           --- \\
\rowcolor{col_sur!10} \tm{CatBoost sur.}             & 1.298 &       9:10.8 && 1.348 &      29:28.6 && 0.140 &      53.4 && 0.108 &      29:06.4 &&   --- &           --- \\
\rowcolor{col_sur!10} \tm{NN-Frye sur.}              & 1.244 & 3:01:48:38.7 && 1.320 & 3:16:01:07.5 && 0.154 & 3:11:39.4 && 0.170 & 1:10:53:27.4 && 0.085 &  3:16:13:50.3 \\
\rowcolor{col_sur!10} \tm{NN-Olsen sur.}             &\B1.169& 3:22:23:46.0 && 1.192 & 2:01:07:30.3 && 0.135 & 1:28:54.8 && 0.130 &   23:56:47.0 && 0.045 &  3:11:33:57.9 \\
\bottomrule
\end{tabular}
\end{adjustbox}
}
\caption{{\small The $\operatorname{MSE}_v$ scores and CPU times for the methods applied to the real-world data sets in \Cref{sec:real_world_data}. The format of the CPU times is days:hours:minutes:seconds, where we omit the larger units of time if they are zero, and the colors indicate the different method classes.}}
\label{tab:real-world-data}
\end{table}

\subsection{Abalone}
\label{real_world_data:Abalone}
We first consider the classical Abalone data set with mixed features. The data set originates from a study by the Tasmanian Aquaculture and Fisheries Institute \parencite{nash1994population} and has been used in several XAI papers \parencite{vilone2020comparative, aas2021explaining, frye_shapley-based_2020, Olsen2022}. The data set contains clear nonlinearity and heteroscedasticity among the pairs of features, and there is a significant pairwise correlation between the features, as all continuous features have a pairwise correlation above $0.775$. The mean correlation is $0.89$, and the maximum is $0.99$. Furthermore, all marginals are skewed. 

We split the $4177$ observations into training ($75\%$) and testing ($25\%$) data sets. The goal is to predict the age of the abalone based on $M=8$ easily obtainable features: \tm{Length}, \tm{Diameter}, \tm{Height}, \tm{WholeWeight}, \tm{ShuckedWeight}, \tm{VisceraWeight}, \tm{ShellWeight}, and \tm{Sex}. All features are continuous except for \texttt{Sex} which is a three-level categorical feature (infant, female, male). Thus, the \empirical\ and \parametric\ methods are not applicable. However, to remedy this, we train two PPR models to act as our predictive models; one based on all features ($\text{PPR}_\text{all}$) and another based solely on the continuous features ($\text{PPR}_\text{cont}$). We chose the PPR model as it outperformed the other prediction models we fitted (GAM, RF, CatBoost). The test MSE increases from $2.04$ to $2.07$ when excluding $\texttt{Sex}$.  
Cross-validation determined that number of terms in $\text{PPR}_\text{all}$ and $\text{PPR}_\text{cont}$ should be $4$ and $7$, respectively. 


\Cref{tab:real-world-data} shows that the best approaches for explaining the PPR predictive models are the \tm{PPR separate}, \texttt{NN-Olsen surrogate}, and \vaeac\ methods. For the $\text{Abalone}_\text{cont}$ data set, the \tm{PPR separate} and \texttt{NN-Olsen surrogate} methods perform equally well and share first place, but both methods are marginally outperformed by the \vaeac\ approach for the $\text{Abalone}_\text{all}$ data set. However, both the \vaeac\ and \texttt{NN-Olsen surrogate} methods are very slow compared to the \tm{PPR separate} approach. The second-best Monte Carlo-based method for the $\text{Abalone}_\text{cont}$ data set is the Gaussian \copula\ approach, even though the Abalone data set is far from Gaussian distributed. This is probably because the \copula\ method does not make a parametric assumption about the marginal distributions of the data, but rather the copula/dependence structure, which makes it a more robust method than the \Gaussian\ approach. 


\subsection{Diabetes}
\label{real_world_data:Diabetes}
The diabetes data set stems from \textcite{efron2004least} and contains $M=10$ baseline features; \tm{Age}, \tm{Sex}, \tm{BMI}, \tm{BP} (blood pressure), and six blood serum measurements (\tm{S1}, \tm{S2}, \tm{S3}, \tm{S4}, \tm{S5}, \tm{S6}) obtained from $442$ diabetes patients. The response of interest is a quantitative measure of disease progression one year after the baseline. Like \textcite{efron2004least}, we treat \tm{Sex} as numerical and standardize all features; hence, we can apply all methods. Many features are strongly correlated, with a mean absolute correlation of $0.35$, while the maximum is $0.90$. The \tm{Age} feature is the least correlated with the other features. Most scatter plots and marginal density functions display structures and marginals somewhat similar to the Gaussian distribution, except those related to the \tm{S4} feature, which has a multi-modal marginal. 
We split the data into a training and test data set at a $75-25$ ratio, and we let the predictive model be a principle component regression (PCR) model with six principal components. This model outperformed the linear model and cross-validated random forest, XGBoost, CatBoost, PPR, and NN models in prediction error on the test data. The PCR model is not easily interpretable as it does not directly depend on the features but on their principal components. 


\Cref{tab:real-world-data} shows that the \texttt{LM separate}, \texttt{GAM separate}, and \texttt{PPR separate} methods obtain the lowest $\operatorname{MSE}_v$ scores, with the \vaeac, \texttt{Gaussian}, and \copula\ approaches having nearly as low $\operatorname{MSE}_v$ scores. We are not surprised that the latter two methods are competitive due to the Gaussian-like structures in the Diabetes data set. The \texttt{LM separate} method is the fastest approach, with a CPU time of $1.9$ seconds. 


\subsection{Red Wine}
\label{real_world_data:Wine}
The Red Wine data set contains information about variants of the Portuguese Vinho Verde wine \parencite{cortez2009using}. The response is a \tm{quality} score between $0$ and $10$, while the $M = 11$ continuous features are based on physicochemical tests: \tm{fixed acidity}, \tm{volatile acidity}, \tm{citric acid}, \tm{residual sugar}, \tm{chlorides}, \tm{free sulfur dioxide}, total \tm{sulfur dioxide}, \tm{density}, \tm{pH}, \tm{sulphates}, and \tm{alcohol}. For the Red Wine data set, most scatter plots and marginal density functions display structures and marginals far from the Gaussian distribution, as most of the marginals are right-skewed. 
Many of the features have no to moderate correlation, with a mean absolute correlation of $0.20$, while the largest correlation in absolute value is $0.683$ between \texttt{pH} and \tm{fix\_acid}. The data set contains $1599$ wines, and we split it into a training ($1349$) and a test ($250$) data set. A cross-validated XGBoost model and a random forest with $500$ trees perform equally well on the test data, and we use the latter as the predictive model $f$.

\Cref{tab:real-world-data} shows that the \tm{RF separate} approach is the best method by far. Next, we have the \tm{CatBoost separate}, \tm{RF surrogate}, \empirical, and \vaeac\ methods. The \tm{RF surrogate} and \tm{CatBoost surrogate} perform well compared to the other \surrogate\ methods. The good performance of the non-smooth \texttt{RF separate} and \texttt{CatBoost separate} methods on the non-smooth predictive model $f$ supports our findings from the simulation studies, where we observed that using a \separatereg\ method with the same form as $f$ was beneficial. The \generative\ methods perform better than the \GH\ and \copula\ methods, while the \Gaussian\ method falls behind. This is intuitive as the data distribution of the Red Wine data set is far from Gaussian distributed.


\subsection{Adult}
\label{real_world_data:Adult}
The Adult data set is based on the 1994 Census database, and the goal is to predict whether a person makes over \$$50\,000$ a year based on $M=14$ mixed features: \tm{age} (cont.), \tm{workclass} (7 cat.), \tm{fnlwgt} (cont.), \tm{education} (16 cat.), \tm{education-num} (cont.), \tm{marital-status} (7 cat.), \tm{occupation} (14 cat.), \tm{relationship} (6 cat.), \tm{race} (5 cat.), \tm{sex} (2 cat.), \tm{capital-gain} (cont.), \tm{capital-loss} (cont.), \tm{hours-per-week} (cont.), and \tm{native-country} (41 cat.). The pairwise Pearson correlation coefficients for the continuous features are all close to zero, with a mean absolute correlation of $0.06$. The data set contains $30\,162$ individuals, and we split it into a training ($30\,000$) and a test ($162$) data set. We train a CatBoost model on the training data to predict an individual's probability of making over \$$50\,000$ a year and use the test data to compute the evaluation criterion. We used a relatively small test set due to memory constraints, and we chose the CatBoost as it outperformed the other prediction models we fitted (LM, GAM, RF, NN).  


\Cref{tab:real-world-data} shows that the best method is the \texttt{CatBoost separate} approach, while second place is shared by the \texttt{RF separate} and \vaeac\ methods. Note that the difference in the $\operatorname{MSE}_v$ score is very small. Like for the previous experiments, we observe that using a \separatereg\ method with the same form as $f$ is beneficial. The \ctree\ approach supports mixed data, but we deemed it infeasible due to a very long computation time. 
Furthermore, the \surrogate\ methods based on \eqref{eq:surrogate_augmented_data} ran out of memory as $\mathcal{X}_\text{aug}$ consists of $30\, 000\times(2^{14}-2) = 491\,460\,000$ training observations. The training time constitutes the majority of the total time for the \separate\ and \surrogate\ methods, while the predicting step only takes a couple of minutes.

\section{Recommendations}
\label{sec:Recommendations}

In this section, we propose a list of advice for when to use the different classes and methods based on the results of the simulations studies and real-world data experiments. The list is not exhaustive. Hence, it must not be interpreted as definite rules but as guidance and points that should be considered when using conditional Shapley values for model explanation. 


\begin{enumerate}
\item For data sets with no or minuscule feature dependencies, the \independence\ approach is the simplest method to use.


\item In general, a \parametric\ approach with the correct (or nearly correct) parametric assumption about the data distribution generates the most accurate Shapley values. 
\begin{itemize}
    \item The \copula\ method does not make an assumption about the marginals of the data, but rather the copula/dependence structure, which makes it a more robust method. 

    \item For features that do not fit the assumed distribution in the \parametric\ approach, one can consider transformations, for example, power transformations, to make the data more Gaussian-like distributed. 
    
    \item For categorical features, one can use, e.g., encodings or entity embeddings to represent the categorical features as numerical. This is needed, as no directly applicable multivariate distribution exists for mixed data. However, there exist copulas that support mixed data. 
    
    \item If the \parametric\ methods are not applicable, the next best option is (often) a \generative\ or \separate\ method, where all considered approaches support mixed data sets by default.
\end{itemize}




\item 
For the \tm{separate} and \surrogate\ methods, using a method with the same form as the predictive model $f$ provides more precise Shapley value estimates.
\begin{itemize}
    \item For some predictive models, e.g., the linear regression model in \Cref{fig:lm_no}, we know that the true conditional model is also a linear model. Thus, using a regression method which can model a linear model (e.g., \texttt{lm}, \texttt{GAM}, \texttt{PPR}) produces more accurate Shapley values. However, the form of the true conditional model is usually unknown for most predictive models.  
    
    \item It is important that the regression method used is flexible enough to properly estimate/model the predictive model $f$.
    
    \item In the numerical simulation studies, the \separate\ methods performed relatively better compared to the other method classes for higher feature dependence. In the real-world experiments, the \separatereg\ methods were also (among) the best approaches on data sets with moderate dependence. 
    
    \item In general, conducting hyperparameter tuning of the regression methods improve the precision of the produced explanations, but this increases the computation time.
    
    \item In the simulation studies, a $\texttt{PPR separate}$ approach with fixed $L = |\s|$ (often) provides fast and accurate Shapley value explanations; see \Cref{Appendix:ExtendedSimulations}. 
\end{itemize}

\item The modeling of the conditional distributions $p(\xsb|\xs)$ in the Monte Carlo-based methods is independent of the predictive model $f$.
\begin{itemize}
    \item For popular data sets, one can fine-tune an \empirical, \parametric, or \generative\ method and let other researchers reuse the method to estimate Shapley values for their own predictive models.
    \item If a researcher is to explain several predictive models fitted to the same data, then reusing the generated Monte Carlo samples will save computation time.
\end{itemize}

\item There is time-accuracy trade-off between the different method classes and approaches.
\begin{itemize}
    \item The simplest \tm{separate} and \surrogate\ methods are rapidly trained, while the complex methods are time-consuming. This is however a one-time upfront time cost. In return, all regression-based methods produce the Shapley value explanations almost instantly. Thus, developers can develop the predictive model $f$ simultaneously with a suitable regression-based method and deploy them together. The user of $f$ will then get predictions and explanations almost instantaneously.
    


    \item In contrast, several of the Monte Carlo-based methods are trained considerably faster than many of the regression-based methods but are, in return, substantially slower at producing the Shapley value explanations. Generating Monte Carlo samples and using them to estimate the Shapley values for new predictions are computationally expensive and cannot be done in the development phase. Thus, the Monte Carlo-based methods cannot produce explanations in real-time.
    
    \item If the predictive model $f$ is computationally expensive to call, then the Monte Carlo-based methods will be extra time-consuming due to $\mathcal{O}(KN2^M)$ calls to $f$. Here $K$, $N$, and $M$ are the number of Monte Carlo samples, predictions to explain, and features, respectively. In contrast, the \tm{separate} and \surrogate\ methods make only $\mathcal{O}(N2^M)$ calls to their fitted regression model(s). 
    
    \item The regression-based methods can be computationally tractable when the Monte Carlo-based methods are not, for example, when $N$ is large. We can reduce the time by decreasing the number of Monte Carlo samples $K$, but this results in less stable and accurate Shapley value explanations.


    \item If accurate Shapley values are essential then a suitable \parametric, \generative, or \separate\ approach with the same form as $f$ yields desirable estimates, depending on the dependence level. The \tm{NN-Olsen surrogate} method also provided accurate Shapley values for some real-world data sets.

    \item If coarsely estimated Shapley values are acceptable, then some of the simple \separate\ methods can be trained and produce estimates almost immediately, such as \texttt{LM separate}. The \texttt{PPR separate} approach with fixed $L = |\s|$ is often a fair trade-off between time and accuracy, especially for smooth predictive functions.
\end{itemize}


\item 
The number of training observations $N_\text{train}$ did not significantly affect the method classes' overall ordering in our simulation studies. However, individual approaches, such as the \texttt{PPR separate}, performed even better when trained on more training observations. 


\item All method classes benefit from having access to multiple CPUs when properly implemented. For example, the Monte Carlo-based approaches can generate the samples for different coalitions and test observations on different cores, and the same when predicting the responses. A \separate\ method can train the individual models in parallel, while a \surrogate\ method can cross-validate the model's hyperparameters on different cores.

\item For high-dimensional settings, the number of models to fit in the \separate\ class is infeasible. Then, the \surrogate\ methods and the \vaeac\ approach with arbitrary conditioning can be useful. However, their accuracy will likely also decrease with higher dimensions. In high-dimensional settings, one can, e.g., group the features into relevant groups \parencite{jullum2021groupshapley} or use tractable estimation strategies \parencite[Section 5.2]{chen2022algorithms} to simplify the Shapley value computations.

\end{enumerate}

\section{Conclusion}
\label{sec:Conclusion}
In this article, we have discussed a large sample of Monte Carlo integration and regression-based methods used to estimate conditional Shapley values for model explanation. In agreement with the literature \parencite{covert2021explaining, chen2022algorithms}, we have divided the studied methods into six different method classes. For each class, we have given an overview of the idea, reviewed earlier proposed methods within the class, and finally proposed and developed several new approaches for most classes. The existing and novel approaches have been systematically evaluated through a series of simulation studies with increasing complexity, as such evaluation has until now been lacking in the field of conditional Shapley values \parencite{chen2022algorithms}. We also conducted several experiments on real-world data sets from the UCI Machine Learning Repository. The ranking of the method classes and approaches differed slightly in the numerical simulation studies and real-world experiments. 

The most accurate Shapley value explanations in the simulation studies were generally produced by a \parametric\ method with a correctly (or nearly correctly) assumed data distribution. This is intuitive, as making a correct parametric assumption is advantageous throughout statistics. However, the true data distribution is seldom known, e.g., for real-world data sets. In the simulation studies with moderate feature dependence levels, the second-best method class was generally the \generative\ class with the \ctree\ and \vaeac\ methods, which outperformed the \independence, \empirical, \separate, and \surrogate\ methods. For high feature dependence, the \separate\ methods improved relative to the other classes, particularly the \texttt{PPR separate} method. Using a \separate\ method with the same form as the predictive model proved beneficial.


In the real-world experiments, the \parametric\ methods fell behind the best approaches, except for the simplest data set with Gaussian-like structures. In general, the best approaches in the real-world experiments belong to the \separate\ method class and have the same form as the predictive model. However, the \tm{NN-Olsen surrogate} method tied the best \separate\ method in one experiment, and the \vaeac\ approach was marginally more precise in another experiment. The second-best method class varied for the different data sets, with all method classes, except the \independence\ and \empirical, taking at least one second place each. 



In addition to the accuracy of the methods, we also investigate the computation time. The regression-based methods are often slowly trained, but they produce the Shapley value explanations almost instantaneously. In contrast, the Monte Carlo-based method are often faster to train but drastically slower at generating the Shapley value explanations.  Finally, we gave some recommendations and considerations for when to use the different method classes and approaches. 


In further work, one direction is to extend the investigation into higher dimensions to verify that the tendencies and order of the methods we discovered remain. However, one would then probably need to sample a subset of the coalitions to cope with the exponential complexity of Shapley values. In agreement with \textcite{chen2022algorithms}, one can also try to determine robust architectures, training procedures, and hyperparameter optimization for the \generative\ and \surrogate\ methods, investigate how non-optimal approaches change the estimated conditional Shapley values, and finally evaluate bias in estimated conditional Shapley values for data with known conditional distributions.

\subsection*{Acknowledgments}
The Norwegian Research Council supported this research through the BigInsight Center for research-driven innovation, project number $237718$.

\clearpage
\newpage

\appendix
\section*{Appendix}
\label{Appendix}
In \Cref{Appendix:Implementation}, we describe implementation details for the methods introduced in \Cref{sec:ConditionalShapleyValuesApproaches}. We provide more details about the \parametric\ methods in \Cref{Appendix:AdditionalInformationParametric}. In \Cref{Appendix:Methods}, we elaborate on approaches for estimating conditional Shapley value explanations used in the main text and describe other methods. We provide additional simulation studies in \Cref{Appendix:ExtendedSimulations}. In \Cref{Appendix:PairPlots}, we include plots of some simulated and real-world data sets. We apply additional methods to the real-world experiments and decompose the computation times of the methods in \Cref{Appendix:RealWorld}. Finally, in \Cref{Appendix:SchematicOverviewMethods}, we provide a schematic overview of the conditional Shapley value explanation framework and the estimation methods within the explainable artificial intelligence field.
In addition, we have explored $884$ simulation configurations and include all result figures in the Supplement.

\section{Implementation Details}
\label{Appendix:Implementation}
In this section, we describe implementation details for the methods introduced in \Cref{sec:ConditionalShapleyValuesApproaches}. We use the \textsc{R}-package \texttt{shapr} \parencite{shapr}, version $0.2.0$, to compute the Shapley values. The package computes the Shapley values as the solution of a weighted least squares problem \parencite{charnes1988extremal,lundberg2017unified,aas2019explaining}. More precisely, the solution is given by $\bphi = (\boldsymbol{Z}^T\boldsymbol{W}\boldsymbol{Z})^{-1}\boldsymbol{Z}^T\boldsymbol{W}\boldsymbol{v} = \boldsymbol{R}\boldsymbol{v}$. The $\boldsymbol{Z}$ matrix is a $2^M \times (M+1)$ binary matrix where the first column consists of $1$s and the remaining columns are the $I(\s)$ representations from \Cref{ConditionalShapleyValues:SurrogateModel:NewMethods} for all $\s \in \pow(\M)$. The $\boldsymbol{W}$ matrix is a $2^M \times 2^M$ diagonal matrix containing the Shapley kernel weights $k(M, |\s|) = (M-1)/(\binom{M}{|\s|}|\s|(M-|\s|))$, while $\boldsymbol{v}$ is a $2^M$-dimensional vector containing the estimated contribution functions $\hat{v}(\s)$. The $\s$ in the latter two cases resembles the coalition of the corresponding row in $\boldsymbol{Z}$. The $\boldsymbol{Z}$ and $\boldsymbol{W}$ matrices are independent of the instance to be explained and are computed by the \texttt{shapr} package, which sets the infinite Shapley kernel weights $k(M,0) = k(M, M) = \infty$ to a large constant $C = 10^6$. When explaining $N_\text{test}$ predictions, we replace $\boldsymbol{v}$ with a $2^M \times N_\text{test}$ matrix $\boldsymbol{V}$, where column $i$ contains the estimated contribution functions for instance $i$. 

The \independence, \empirical, \Gaussian, \copula, and \ctree\ methods are implemented in the \texttt{shapr} package, and we use default hyperparameter values. For the other methods, we estimate $\boldsymbol{V}$ and multiply it with $\boldsymbol{R}$ to get the estimated Shapley values. \textcite{Olsen2022} implement the \vaeac\ approach as an add-on to the \texttt{shapr} package, and we use the default architecture and hyperparameters. For the numerical simulation studies in \Cref{sec:Numerical_Simulation_Studies}, we train the \vaeac\ approach for $200$ epochs and use the estimated model parameters at the epoch with the lowest validation error, where $25\%$ of the data constitutes the validation data. For the more complex real-world data distributions in \Cref{sec:real_world_data}, the \vaeac\ approach needs more training epochs to learn to model the data distributions properly. For the $\text{Abalone}_\text{cont}$, $\text{Abalone}_\text{all}$, Diabetes, Wine, and Adult data sets, we let the number of epochs be: $10\,000$, $40\,000$, $5000$, $10\,000$, and $200$, respectively. Other configurations than the default architecture and hyperparameters might reduce the number of needed learning epochs. In \Cref{Appendix:ExtendedSimulations,Appendix:RealWorld}, we provide \vaeac\ approaches with other numbers of epochs for the numerical simulation studies and the real-world data experiments, respectively.

Throughout the article, if not otherwise specified, we use $K=250$ Monte Carlo samples in \eqref{eq:KerSHAPConditionalFunction} for the Monte Carlo-based methods, which \textcite{Olsen2022} found to be a fair trade-off between accuracy and computation time. However, recall that the \empirical\ and \ctree\ methods (often) use less samples and rather weight them, as described in \Cref{subsec:ConditionalShapleyValues:empirical,subsec:ConditionalShapleyValues:GenerativeModel:Ctree}, respectively. The \independence\ approach is implemented as a special case of the \empirical\ approach in version $0.2.0$ of the \texttt{shapr} package. Hence, it does not support mixed data. For data sets with categorical features, we have implemented our own \tm{independence}$^*$ approach which directly samples from the training data.

For the \Burr\ and \GH\ approaches, we estimate the parameters of the distributions by maximizing the likelihood function using the Nelder-Mead optimization routine \parencite{nelder1965simplex}, with default parameters in the \texttt{optim} function in base \textsc{R} \parencite{R_language}. Tuning the hyperparameters of the optimization algorithm and/or using a more advanced fitting procedure might improve the approaches. We run the optimization procedures until convergence. The number of parameters to estimate in the Burr and GH distribution is $2M+1$ and $\tfrac{1}{2}(M+1)(M+4)$, respectively. The optimization of the \GH\ method relies on good starting values, which we get from the \texttt{ghyp} package \parencite{ghyp}. The \texttt{ghyp} package uses a sophisticated multi-cycle, expectation, conditional estimation (MCECM) algorithm to estimate the parameters for another more general parameterization of the GH distribution which lacks closed-form expressions for the conditional distributions. 

For the \separate\ methods, we tune (some of) the hyperparameters of the different methods using cross-validation procedures implemented in the packages, by us, or by using the \texttt{caret} package \parencite{caret}. The \texttt{LM separate} approach was fitted using the \texttt{lm} function in the \texttt{stats} package in base \textsc{R}. We use the \texttt{mgcv} package \parencite{mgcv}, with default parameters, to fit the \texttt{GAM separate} method. Note that in the \texttt{mgcv} package, the smoothing parameters in the penalized regression splines are selected by generalized cross-validation during the fitting procedure. The \texttt{PPR separate} method uses the \texttt{ppr} function in the \texttt{stats} package with default parameters, except the number of terms $L$, which we determine by cross-validation. 
The \texttt{RF separate} approach is based on the \texttt{ranger} package \parencite{ranger}. We use $500$ decision trees and the \texttt{caret} package to do cross-validation on \texttt{mtry}, \texttt{splitrule}, and \texttt{min.node.size}, while we use default values for the remaining hyperparameters. Finally, the \texttt{CatBoost separate} method uses the \texttt{CatBoost} algorithm \parencite{catboost}, which is based on gradient-boosted decision trees, with default parameters (most notably, $1000$ trees with depth $6$). We employ early stopping of the \texttt{CatBoost} method if no improvement of the evaluation metric value was made in $100$ iterations. One could employ cross-validation to tune the hyperparameters, but this would increase the computation time drastically as the \texttt{CatBoost} algorithm has many hyperparameters. An alternative is to tune only some of them. 

For the \surrogate\ methods, we use the same packages as above and tune the same hyperparameters if not otherwise specified. For the \texttt{RF surrogate} method, we reduced the number of trees from $500$ to $200$ due to high computation time. 
For the \texttt{CatBoost surrogate} approach, we increase the maximum number of trees to $10\,000$, but we still employ the same early stopping regime. 

Originally, \textcite{frye_shapley-based_2020} let the masking value be $-1$, as they only consider positive data, but this is not applicable for unbounded features. We let the value be $-5$ in the simulations and real-world experiments in \Cref{sec:Numerical_Simulation_Studies,sec:real_world_data}, respectively, which is a value not present in the data sets. For the \texttt{NN-Frye surrogate} approach, we use the same fully connected neural network and carry out the same cross-validation as in \textcite{frye_shapley-based_2020}. That is, $\texttt{depth} = 2$, $\texttt{width} \in \{128, 256, 512\}$, $\texttt{batch\tu size} = 256$, and $\texttt{learning\tu rate} \in \{0.001, 0.0001\}$ in the Adam optimizer \parencite{kingma2014adam}. We use $75\%$ of the data to train the networks and the remaining observations as validation data. We use the network parameters at the epoch with the lowest validation error as the final model. \textcite{frye_shapley-based_2020} use $2000-10\,000$ training epochs, while we use $\texttt{num\tu epochs} = 3000$ to make the method more time-wise competitive and as the validation error obtains its minimum long before the last epoch in the simulation studies. Another alternative is to let \texttt{num\tu epochs} be arbitrarily large and stop the training if no improvement has been made to the validation error for a fixed number of epochs, that is, employing early stopping.





In the \texttt{NN-Olsen surrogate} approach, we use batch normalization layers, ELU activation functions, and skip connections with summation over each layer in the network. We carry out similar hyperparameter tuning as the \texttt{NN-Frye surrogate} approach. That is, $\texttt{depth} = 3$, $\texttt{width} \in \{32, 64, 128\}$, $\texttt{num\tu epochs} = 500$, $\texttt{batch\tu size} = 128$ (as we duplicate the batch size), and $\texttt{learning\tu rate} \in \{0.01, 0.001, 0.0001\}$ in the Adam optimizer.  Instead of specifying $\texttt{num\tu epochs}$, another option could have been to train the network until a stopping criterion was meet, e.g., no improvement in the validation measure for a specific number of epochs. We observe relative small differences between the nine different hyperparameter choices, and one could thus potentially reduce the training time by a factor of nine by using $\texttt{width} = 64$ and $\texttt{learning\tu rate} = 0.001$ as default values. The same also applies to the \texttt{NN-Frye surrogate} approach. The networks are implemented in \texttt{torch} \parencite{rtorch}.

In the real-world data experiments in \Cref{sec:real_world_data}, we omit the cross-validation of the hyperparameters in the \tm{NN surrogate} approaches to make them more time-wise competitive. We let $\texttt{lr} = 0.001$ and $\texttt{width} = 256$ in the \texttt{NN-Frye surrogate} approach, while we use the same learning rate for the \texttt{NN-Olsen surrogate} method but we let $\texttt{width} = 64$. The convergence rates of the networks' validation errors vary in the different real-world data experiments. Hence, we use different \tm{num\_epochs} for each experiment.
For the $\text{Abalone}_\text{cont}$, $\text{Abalone}_\text{all}$, Diabetes, Wine, and Adult data sets, we let \tm{num\_epochs} in the \tm{NN-Frye surrogate} method be: $40\,000$, $40\,000$, $10\,000$, $40\,000$, and $3000$, respectively. The corresponding values for the \tm{NN-Olsen surrogate} method are $20\,000$, $10\,000$, $2500$, $10\,000$, and $500$. Other configurations than the architecture and hyperparameters set above might reduce the number of needed learning epochs. In \Cref{Appendix:RealWorld}, we provide \tm{NN surrogate} methods with other numbers of epochs, as a higher \tm{num\_epochs} can make the methods more precise but at the cost of increased computation time.




\section{Additional Information About the Parametric Methods}
\label{Appendix:AdditionalInformationParametric}
In this section, we elaborate on the \texttt{copula} approach and give a short introduction to the multivariate Burr and generalized hyperbolic distributions.

\subsection{Copulas}
\label{Appendix:AdditionalInformationParametric:Copula}
The definition of an $M$-dimensional copula is a multivariate distribution, $C$, with uniformly distributed marginals $\mathcal{U}[0,1]$. Sklar's theorem states that every multivariate distribution $F$ with marginals $F_1, F_2, \dots, F_M$ can be written as $F(x_1, \dots, x_M) = C(F_1(x_1), \dots, F_M(x_M))$, for some appropriate $M$-dimensional copula $C$. In fact, the copula from the previous equation has the expression $C(u_1, \dots, u_M) = F(F_1^{-1}(u_1), \dots, F_M^{-1}(u_M))$, where the $F_j^{-1}(u_j)$s are the inverse distribution functions of the marginals. While other copulas may be used, the Gaussian copula has the benefit that we may use the analytical expressions for the conditionals for the Gaussian distribution.

The Gaussian copula model used by \textcite{aas2019explaining} is very flexible with regard to the marginal distributions, but quite restrictive in the dependence structures it can capture. It can only represent radially symmetric dependence relationships and does not allow for tail dependence (i.e., joint occurrence of extreme events has small probability). One can use other copulas in the \texttt{copula} approach instead. For example, \textcite{aas2021explaining} use vine copulas, more specifically, a particular type of R-vines (regular vines) called D-vines \parencite{kurowicka2005distribution} when they estimate conditional Shapley values. Regular vines do not exclude categorical data, but the methods become more complicated when categorical features are included; hence, \textcite{aas2021explaining} exclude them. \textcite{zhao2020missing} propose a semiparametric algorithm to impute missing values for mixed data sets via a Gaussian copula.


\subsection{Burr Distribution}
\label{Appendix:AdditionalInformationParametric:Burr}
The Burr distribution allows for heavy-tailed, skewed marginals, and nonlinear dependencies, which can be found in real-world data sets \parencite{takahasi_note_1965}. The density of the $M$-dimensional Burr distribution is given by
\begin{align*}
    p(\boldsymbol{x}) = \frac{\Gamma(\kappa + M)}{\Gamma(\kappa)} \left(\prod_{m=1}^M b_mr_m\right) \frac{\prod_{m=1}^M x_m^{b_m-1}}{\left(1+\sum_{m=1}^M r_mx_m^{b_m}\right)^{\kappa+M}},
\end{align*}
for $x_m > 0$ . The $M$-dimensional Burr distribution has $2M+1$ parameters, namely, $\kappa$, $b_1, \dots b_M$, and $r_1, \dots, r_M$. Furthermore, the Burr distribution is a compound Weibull distribution with the gamma distribution as compounder \parencite{takahasi_note_1965}, and it can also be seen as a special case of the Pareto IV distribution \parencite{Yari2006InformationAC}.

Any conditional distribution of the Burr distribution is in itself a Burr distribution \parencite{takahasi_note_1965}. Without loss of generality, assume that the first $S < M$ features are the unobserved features, then the conditional density $p(x_1,\dots,x_S | x_{S+1} = x^*_{S+1}, \dots, x_M = x_M^*)$, where $\boldsymbol{x}^*$ indicates the conditional values, is an $S$-dimensional Burr density. The associated parameters are then $\tilde{\kappa}, \tilde{b}_1, \dots, \tilde{b}_S$, and $\tilde{r}_1, \dots, \tilde{r}_S$, where $\tilde{\kappa} = \kappa + M - S$, while $\tilde{b}_j = b_j$ and $\tilde{r}_j = \frac{r_j}{1+\sum_{m=S+1}^M r_m(x_m^*)^{b_m}}$, for all $j=1,2,\dots,S$.

\subsection{Generalized Hyperbolic Distribution}
\label{Appendix:AdditionalInformationParametric:GH}
The generalized hyperbolic distribution $\operatorname{GH}(\lambda, \omega,\bmu, \boldsymbol{\Sigma}, \bbeta)$ is parameterized by an index parameter $\lambda$, concentration parameter $\omega$, location vector $\bmu$, dispersion matrix $\boldsymbol{\Sigma}$, and skewness vector $\bbeta$ \parencite{browne2015mixture}. A random variable $\x$ is GH distributed if it can be represented by $\x = \bmu + W\bbeta + \sqrt{W}\boldsymbol{U}$, where $W \sim \operatorname{GIG}(\lambda, \omega, \omega)$, $\boldsymbol{U} \sim \mathcal{N}(0, \boldsymbol{\Sigma})$ and $W$ is independent of $\boldsymbol{U}$. GIG is the generalized inverse Gaussian distribution introduced by \textcite{good1953population}. The density of the $M$-dimensional GH is given by
\begin{align*}
    p(\boldsymbol{x}) = \left[\frac{\omega + \delta(\x, \bmu, \boldsymbol{\Sigma})}{\omega + \bbeta^T\boldsymbol{\Sigma}^{-1}\bbeta}\right]^{\frac{\lambda-M/2}{2}} \frac{K_{\lambda-M/2} \left( \sqrt{(\omega + \delta(\x, \bmu, \boldsymbol{\Sigma}))(\omega + \bbeta^T\boldsymbol{\Sigma}^{-1}\bbeta)} \right)}{(2\pi)^{M/2}|\boldsymbol{\Sigma}|^{1/2}K_{\lambda}(\omega)\exp\set{-(\x - \bmu)^T\boldsymbol{\Sigma}^{-1}\bbeta}},
\end{align*}
where $\delta(\x, \bmu, \boldsymbol{\Sigma}) = (\x - \bmu)^T \boldsymbol{\Sigma}^{-1} (\x - \bmu)$ is the squared Mahalanobis distance between $\x$ and $\bmu$, $K_\lambda$ is the modified Bessel function of the third kind with index $\lambda$. 

\textcite{wei2019mixtures} showed that when $\x$ is partitioned as $(\xs, \xsb)$, the conditional distribution $\xsb | \xs = \xss \sim \operatorname{GH}^*(\lambda_{\sbb|\s}, \chi_{\sbb|\s}, \phi_{\sbb|\s}, \bmu_{\sbb|\s}, \boldsymbol{\Sigma}_{\sbb|\s}, \bbeta_{\sbb|\s})$, where $\lambda_{\sbb|\s} = \lambda - |\s|/2$, $\chi_{\sbb|\s} = \omega + (\xss - \bmu_\s)^T\bSigma_{\s\s}^{-1}(\xss - \bmu_\s)$, $\psi_{\sbb|\s} = \omega + \bbeta_\s^T\bSigma_{\s\s}^{T}\bbeta_\s$, $\bmu_{\sbb|\s} = \bmu_\sbb + \bSigma_{\s\sbb}^T\bSigma_{\s\s}^{-1}(\xss - \bmu_\s)$, $\boldsymbol{\Sigma}_{\sbb|\s} = \bSigma_{\sbb\sbb} - \bSigma_{\s\sbb}^T\bSigma_{\s\s}^{-1}\bSigma_{\s\sbb}$, and $\bbeta_{\sbb|\s} = \bbeta_\sbb - \bSigma_{\s\sbb}^T\bSigma_{\s\s}^{-1}\bbeta_\s$. Here the $\operatorname{GH}^*$ indicates that another parameterization of the GH distribution, proposed by \textcite{mcneil2015quantitative}, is used for the conditional distribution due to technical reasons: 
\begin{align*}
    p(\boldsymbol{x}) = \left[\frac{\chi + \delta(\x, \bmu, \boldsymbol{\Sigma})}{\psi + \bbeta^T\boldsymbol{\Sigma}^{-1}\bbeta}\right]^{\frac{\lambda-M/2}{2}} \frac{(\psi/\chi)^{\lambda/2}K_{\lambda-M/2} \left( \sqrt{(\chi + \delta(\x, \bmu, \boldsymbol{\Sigma}))(\psi + \bbeta^T\boldsymbol{\Sigma}^{-1}\bbeta)} \right)}{(2\pi)^{M/2}|\boldsymbol{\Sigma}|^{1/2}K_{\lambda}(\sqrt{\chi\psi})\exp\set{-(\x - \bmu)^T\boldsymbol{\Sigma}^{-1}\bbeta}}.
\end{align*}

\newpage
\section{Additional Approaches}
\label{Appendix:Methods}
In this section, we provide more information about the methods used in the main text, describe additional approaches we have used, and point out potential methods that can be incorporated into the Shapley value explanation framework in the future. 

\subsection{The Missingness During Training Procedure}
\label{Appendix:Methods:Missingness}
\textcite[Appendix E.2]{covert2021explaining} and \textcite[Section 5.1.3]{chen2022algorithms} describe a procedure where they directly estimate the conditional expectation by modifying the training process of the predictive model $f$ such that it handles missing features. That is, they train $f$ on a particular objective function such that its predictions of observations with missing features are equivalent to marginalizing out the features using the conditional distribution. However, as we focus on model-agnostic post hoc explanation for arbitrary $f$ and most predictive models do not support missingness, we skipped this procedure in the main part of the article. 

\subsection{The Generative Method Class}
\label{Appendix:Methods:Generative}
Here we describe alternative versions of the \vaeac\ approach that we have considered and then list other potential methods in the \generative\ method class.

\subsubsection{VAEAC with Response Feature}
\label{Appendix:Methods:Generative:VAEAC_with_response}
The \vaeacf\ approach includes the predicted response of the predictive model $\hat{y} = f(\x)$ as an additional feature that is always unobserved in the deployment phase. That is, the extended training data takes the form $\{\x^{[i]}, f(\x^{[i]})\}_{i=1}^{N_\text{train}}$. This idea was proposed by \textcite{ivanov_variational_2018}, the creators of the \vaeac\ methodology, and they argued that this extension could improve the modeling of the data, especially for multi-modal data. The \vaeacf\ approach will generate $(\x_{\thickbar{\mathcal{S}}}^{(k)}, \hat{y}_{\mathcal{S}}^{(k)}) \sim p_{\bpsi,\btheta}(\xsb, y_\s| \xs, \sbb)$, for which two possible procedures are available. First, in the indirect \vaeacfindir\ approach, we only use the $\x_{\thickbar{\mathcal{S}}}^{(k)}$ part, combine them with $\xs$, send them through the predictive model $f$, and finally estimate the contribution function with $\hat{v}(\mathcal{S}, \x) = \frac{1}{K}\sum_{k=1}^K f(\x_{\thickbar{\mathcal{S}}}^{(k)}, \xs)$. For the other approach, which we call the direct \vaeacfdir\ approach, we skip the intermediate step where we evaluate the model at the Monte Carlo samples by rather using the $\hat{y}_{\mathcal{S}}^{(k)}$ samples, that is, $\hat{v}(\mathcal{S}, \x) = \frac{1}{K}\sum_{k=1}^K \hat{y}_{\mathcal{S}}^{(k)}$. This saves time if $f$ is computationally expensive to call. We use the same hyperparameters for these two approaches as for the original \vaeac\ approach and the estimated model parameters at the epoch with the lowest validation error; see \Cref{Appendix:Implementation}. 
We consider several maximum numbers of epochs and indicate this by including the number in the method name. For example, \tm{VAEAC-f-dir-500} means that we trained the \tm{VAEAC-f-dir} method for $500$ epochs.

\subsubsection{VAEAC with Paired Sampling}
\label{Appendix:Methods:Generative:VAEAC_with_paired_sampling}
The \tm{VAEAC-paired} approach is identical to the \vaeac\ method described in \Cref{subsec:ConditionalShapleyValues:GenerativeModel:VAEAC}, except that we used paired sampling when generating the mask. That means that both $\s$ and $\sbb$ are applied to the same observation in the training phase of the \vaeac\ model.

\subsubsection{Potential Generative Methods}
\label{Appendix:Methods:Generative:PotentialMethods}
We can use various applicable generative methods to generate the conditional Monte Carlo samples and Shapley values, such as non-parametric vine copulas \parencite{aas2021explaining}. We now provide a non-exhaustive list of other applicable generative methods, which, to the best of our knowledge, have yet to be used in Shapley value estimation. Computing the Monte Carlo samples coincide with the field of \textit{multiple imputation of missing values}. The methods in this rich field can be categorized into two classes \parencite{zheng2022diffusion}. The first class contains the iterative approaches: the Multivariate Imputations based on Chained Equations (\texttt{MICE}) \parencite{MICE} and \texttt{MissForest} \parencite{MissForest}. The second class contains the deep generative models: Multiple Imputation using Denoising Autoencoders (\texttt{MIDA}) \parencite{MIDA}, Missing Data Importance-weighted Autoencoder (\texttt{MIWAE)} \parencite{MIWAE}, Generative Adversarial Imputation Nets (\texttt{GAIN}) \parencite{yoon_gain_2018}, and Conditional Score-based Diffusion Models for Tabular data (\texttt{CSDI\tu T}) \parencite{zheng2022diffusion}. Other methods are the Arbitrary Conditioning Flow model (\texttt{ACFlow}) \parencite{acflow}, the Neural Conditioner (\texttt{NC}) \parencite{Belghazi}, Neural Autoregressive Distribution Estimation (\texttt{NADE}) \parencite{NADE}, and Universal Marginalizers (\texttt{UM}) \parencite{DouglasUniversalMarginalizer}.

\subsection{The Separate and Surrogate Method Class}
\label{Appendix:Methods:Regression}
In this section, we describe additional regression-based approaches. Note that all the regression methods can, in theory, be used both in the \tm{separate} and \surrogate\ frameworks. Some of them might however be infeasible for the latter in practice due to memory or time constraints, especially for large training data sets and high dimensions. Note that not all the regression methods minimize the mean squared error loss function.

\subsubsection{Polynomial Regression}
Polynomial regression is an extension of the linear regression where we model the relationship as an $p$th degree polynomial for each feature. That is, the model takes the following form:

\begin{equation*}
    \begin{split}
        f(\x) 
        = 
        \beta_0 + \sum_{j=1}^M \p{\beta_{j,1}x_j^1 + \beta_{j,2}x_j^2 + \dots + \beta_{j,p}x_j^p} 
        = 
        \beta_0 + \sum_{j=1}^M\sum_{k=1}^p \beta_{j,k}x_j^k.
    \end{split}
\end{equation*}
We estimate the coefficients of the polynomial model using the \texttt{lm} function in base \Rlang\ with \texttt{formula = "y $\sim$ poly(X1, deg = p) +...+ poly(XM, deg = p)"}, where \texttt{p} is the degree. We call the approach \tm{Poly-p}.

\subsubsection{Linear Regression with Interactions}
We extend the linear regression model by including interactions between the features. For example, we get the following model formula when we include first-order interactions:
\begin{equation*}
    \begin{split}
        f(\x) 
        =
        \beta_0 + \sum_{j=1}^M \beta_jx_j + \sum_{j=1}^{M-1}\sum_{k=j+1}^M\beta_{j,k}x_jx_k.
    \end{split}
\end{equation*}
For second-order interactions, we get the following model:
\begin{equation*}
    \begin{split}
        f(\x) 
        =
        \beta_0 + \sum_{j=1}^M \beta_jx_j + \sum_{j=1}^{M-1}\sum_{k=j+1}^M\beta_{j,k}x_jx_k + \sum_{j=1}^{M-2}\sum_{k=j+1}^{M-1}\sum_{l=k+1}^{M}\beta_{j,k,l}x_jx_kx_l.
    \end{split}
\end{equation*}
We estimate the coefficients in the interaction model using the \texttt{lm} function in base \Rlang\ with \texttt{formula = "y $\sim \texttt{(.)}^{\texttt{o}+1}$"}, where \texttt{o} is the order. We call the approach \tm{LM-inter-o}.

\subsubsection{Polynomial Regression with Interactions}
Here we extend the linear regression model with interactions by also allowing for polynomial terms. For example, the polynomial regression model with polynomial degree $2$ and interactions of one order lower takes the following form:
\begin{equation*}
    \begin{split}
        f(\x) 
        =
        \beta_0 + \sum_{j=1}^M \beta_jx_j + \sum_{j=1}^{M}\sum_{k=j}^M\beta_{j,k}x_jx_k.
    \end{split}
\end{equation*}
We estimate the coefficients in the polynomial interaction model using the \texttt{lm} function in base \Rlang\ with \texttt{formula = "y $\sim$ poly(X1, ..., XM, deg = d)"}, where \texttt{d} is the degree. For the surrogate version, we do not include interactions with the binary mask features, as the number of coefficients to be estimated drastically increases. 
We call the approach \tm{Poly-inter-d}.


\subsubsection{Generalized Additive Models}
We also fit GAMs using the \texttt{gam} package \parencite{gam_package}, which differs slightly from the \texttt{mgcv} package discussed in \Cref{Appendix:Implementation}. In the \texttt{gam} package, we can directly specify the degrees of freedom for the splines. We consider three different versions: one with $\texttt{df} = 5$ (\tm{GAM-5}), another with $\texttt{df} = 10$ (\tm{GAM-10}), and in the last, we conduct cross-validation using the \texttt{caret} package to tune the degrees of freedom (\tm{GAM-CV}). 

\subsubsection{Elastic Net Regression}
Elastic Net models add regularization to the model coefficients, and the popular Lasso and Ridge regression models are special cases. However, they do not minimize the MSE, but we still include them. The objective function for the Gaussian family is: $\min_{(\beta_0, \boldsymbol{\beta}) \in \mathbb{R}^{p+1}}\tfrac{1}{2N} \sum_{i=1}^N (y_i -\beta_0-\boldsymbol{x}_i^T \boldsymbol{\beta})^2+\lambda \left( (1-\alpha)\|\boldsymbol{\beta}\|_2^2/2 + \alpha\|\boldsymbol{\beta}\|_1\right)$, where $\lambda \geq 0$ is a regularization parameter and $0\leq \alpha \leq 1$ is a compromise between Ridge ($\alpha = 0$) and Lasso regression ($\alpha = 1$). We consider $\alpha \in \set{0, 0.5, 1}$ and call the corresponding methods for \texttt{Ridge}, \texttt{Elastic}, and \texttt{Lasso}. We use the \texttt{glmnet} package \parencite{glmnet} to fit the models and use the package's cross-validation procedure to tune $\lambda$. 


\subsubsection{Principal Component Regression}
\label{Appendix:Methods:Regression:pcr}
The difference between regular linear regression and principal component regression (PCR) is that the latter regress the response on the principal components instead of the original features. For more details, see \textcite[pp.\ 79-80]{hastie2009elements}. We use the \texttt{pls} package \parencite{plspcr} to fit the PCR model and use the package's cross-validation procedure to determine the number of principal components to include in the final model. We call the approach \texttt{PCR}.

\subsubsection{Partial Least Squares}
The partial least squares (PLS) regression model is similar to PCR, but the PLS also uses the response when constructing the linear combinations of the features for regression. For more details, see \textcite[pp.\ 80-82]{hastie2009elements}. We use the \texttt{pls} package \parencite{plspcr} to fit the PLS model and use the package's cross-validation procedure to determine the number of components to include in the final regression model. We call the approach \texttt{PLS}.

\subsubsection{Projection Pursuit Regression}
\label{Appendix:Methods:Regression:ppr}
\Cref{ConditionalShapleyValues:SeparateModels:ppr} and \Cref{Appendix:Implementation} describe the PPR model and explain that we use cross-validation to determine the number of terms $L$ in the \texttt{PPR separate} approach. An alternative is the \texttt{PPR-fixed separate} approach where we let $L = |\s|$. This method is much faster than \texttt{PPR separate} but still competitive; see \Cref{Appendix:ExtendedSimulations}. For larger values of $M$, letting $L = \operatorname{min}(|\s|, L_\text{max})$, where $L_\text{max}$ is a threshold, will reduce the computation time even more.

\subsubsection{Support Vector Machines}
Support vector machines (SVM) used for regression are also known as support vector regression, and see \textcite[Ch.\ 12.3]{hastie2009elements} for an introduction. We use $\epsilon$-type regression and a radial kernel, but one could also consider $\nu$-regression and linear, polynomial, and sigmoid kernels. We use the \texttt{WeightSVM} package \parencite{SVM_package} to fit the SVM, as it supports weighting of the observations, but the more well-known \texttt{e1071} package \parencite{e1071} could also have been used.  We call the approach \texttt{SVM}.

\subsubsection{K-Nearest Neighbors}
The K-nearest neighbors (KNN) regression model is a memory-based approach that does not require any model to be fit. For a given individual $\x^*$, the model finds the $K$ closes observations in the training data and return the mean response of these observations. We use the \texttt{kknn} package \parencite{kknn} to train the KNN model and to conduct hyperparameter tuning.  We call the approach \texttt{KNN}.

\subsubsection{Single Decision Tree}
A regression decision tree partitions the feature space into a set of rectangles and predicts a new observation's response as the mean response of the training observations in the particular partition. CART, C4.5, and CTree are popular methods for tree-based regression, which are very simple to understand yet powerful; see, e.g., \textcite[Ch.\ 9.2]{hastie2009elements}. We use the \texttt{rpart} package \parencite{rpart} to fit a decision tree with complexity parameter $0.001$ and then prune the tree afterward. We call the approach \texttt{Tree}.

\subsubsection{Random Forest}
\label{Appendix:Methods:Regression:rf}
In the main text, the \tm{RF} approach was tuned using cross-validation, which leads to a large computation time. Here we propose a default approach called \tm{RF-def} with $500$ trees and default hyperparameter values in the \texttt{ranger} package \parencite{ranger}. A potential improvement is to vary the number of trees based on the coalition size $|\s|$ instead of having a fixed number as, e.g., $500$ trees might be excessive when $\s$ is a singleton.

\subsubsection{Boosting}
Both \texttt{XGBoost} \parencite{chen2015xgboost} and \texttt{CatBoost} \parencite{catboost} are gradient-based boosted decision trees, but they differ in that the latter supports categorical data by default while the former require the user to do, e.g., one-hot encoding. We include two versions of the \texttt{XGBoost} approach: one where we use default hyperparameter values (\texttt{XGBoost-def}) and one where we tune \texttt{nrounds}, \texttt{max\tu depth}, \texttt{eta}, \texttt{gamma}, \texttt{colsample\tu bytree} using the default grid in the \texttt{caret} package. We call the latter approach \texttt{XGBoost}, which is time-consuming due to the extensive hyperparameter tuning.

\subsubsection{Neural Networks and Multilayer Perceptron}
In \Cref{Appendix:ExtendedSimulations,Appendix:RealWorld}, we explore \tm{NN surrogate} methods with hyperparameters defined in \Cref{sec:real_world_data} but with different maximum numbers of learning epochs. E.g., \tm{NN-Olsen-500 surrogate} indicates that $\tm{num\_epochs} = 500$, but we still use the network weights at the epoch with the lowest validation error. We have also implemented a version that employs early stopping, as discussed in \Cref{Appendix:Implementation}. More precisely, we stop the training if no improvement has been made to the validation error in $150$ epochs. Additionally, this version initiates ten networks to reduce the likelihood of poorly initiated network parameters, as discussed in \Cref{Appendix:RealWorld}. We train the ten networks for fifteen epochs and continue only with the network with the lowest validation error. We denote this method by \tm{NN-Olsen-ES} and \tm{NN-Frye-ES}.

The multilayer perceptron (MLP) is a fully connected feedforward artificial neural network. We include some small networks to compare these against the large \tm{NN-Olsen surrogate} and \tm{NN-Frye surrogate} methods described in \Cref{ConditionalShapleyValues:SurrogateModel:NN} and \Cref{Appendix:Implementation}.
We call the approach, e.g., \texttt{NN-[$u$, $v$, $w$]}, which means that the network has three layers where the number of neurons in the layers are $u$, $v$, and $w$, respectively. We use the \texttt{RSNNS} package \parencite{RSNNS}, with default hyperparameters and $200$ epochs.







\subsubsection{Potential Methods}
\textcite{shaff} use a projected random forest to estimate Shapley effects, which is not the same as Shapley values in \eqref{eq:ShapleyValuesDef}. The projected random forest is a \surrogate\ model which provides predictions of the output conditioned on any feature subset. Thus, their procedure can be adapted to estimate conditional Shapley values. \textcite{fastshap} propose another neural network based procedure that skips the modeling of the data/response altogether by training a complex neural network which takes in the full input feature vector $\x$ and directly outputs the Shapley values $\bphi$.

\subsubsection{Surrogate Regression Methods in High-Dimensions}
In high-dimensional settings, to reduce the computational cost, one can consider training the \surrogate\ model on a sampled subset of the augmented representations in \eqref{eq:surrogate_augmented_data}. In that case, one should ensure that all coalitions are present. Uniformly sampling will mostly sample coalitions with approximately half of the features present, as the number of coalitions with $|\s|$ entries is given by $\binom{M}{|\s|}$. Therefore, \textcite{covert2021explaining} propose to first uniformly sample the coalition size, i.e., $|\s| \sim \mathcal{U}[1,M-1]$, and then sample $|\s|$ features with uniform probability. Recall that the number of terms in the Shapley value formula in \eqref{eq:ShapleyValuesDef} grows at an exponential rate; hence, in higher dimensions, it is common practice to estimate the Shapley values based on a sampled collection of coalitions with replacement \parencite{lundberg2017unified, chen2022algorithms, Olsen2022}. We can then create $\mathcal{X}_\text{aug}$ only based on these coalitions. Furthermore, many regression models support weighting of the observations, which we can set as the sampling frequency of the different coalitions. \textcite{Olsen2022} use a similar idea when sampling masks. For regression models which do not support weights, one can duplicate the relevant data, but this na\"ive approach increases the number of rows in the augmented design matrix.




\newpage
\section{Additional Simulation Studies}
\label{Appendix:ExtendedSimulations}
In this section, we extend the numerical simulation studies in \Cref{sec:Numerical_Simulation_Studies} in two directions. First, in \Cref{Appendix:ExtendedSimulations:Intermediate}, we include more setups with Gaussian data. Second, in \Cref{Appendix:ExtendedSimulations:Intermediate_burr}, we use the multivariate Burr distribution instead of the multivariate Gaussian.


\subsection{Gaussian Distributed Experiments}
\label{Appendix:ExtendedSimulations:Intermediate}
Here we include additional setups to the experiments in \Cref{subsection:Simulation:lm,subsection:Simulation:GAM}. We include:
\begin{enumerate}[align=left, leftmargin=3.5em, itemsep=1pt, topsep=3pt] 
    \item [\texttt{lm\textunderscore some\textunderscore interactions}:] $f_{\lm, \some}(\x) =  f_{\lm, \no}(\x) + \gamma_1x_1x_2$, 
    
    \item [\texttt{lm\textunderscore many\textunderscore interactions}:] $f_{\lm, \many}(\x) =  f_{lm, \more}(\x) + \gamma_3x_5x_6$, 
    
    \item [\texttt{gam\textunderscore one}:] $f_{\gam, \one}(\x) = \beta_0 + \sum_{i=1}^{1}\beta_i\cos(x_i) + \sum_{j=2}^{M} \beta_jx_j$,
    
    \item [\texttt{gam\textunderscore two}:] $f_{\gam, \two}(\x) = \beta_0 + \sum_{i=1}^{2}\beta_i\cos(x_i) + \sum_{j=3}^{M} \beta_jx_j$, 

    \item [\texttt{gam\textunderscore five}:] $f_{\gam, \five}(\x) = \beta_0 + \sum_{i=1}^{5}\beta_i\cos(x_i) + \sum_{j=6}^{M} \beta_jx_j$,

    \item [\texttt{gam\textunderscore some\textunderscore interactions}:] $f_{\gam, \some}(\x) =  f_{\gam, \no}(\x) + \gamma_1g(x_1, x_2)$, 

    \item [\texttt{gam\textunderscore many\textunderscore interactions}:] $f_{\gam, \many}(\x) =  f_{\gam, \more}(\x) + \gamma_3g(x_5, x_6)$, 
\end{enumerate}
where $\boldsymbol{\beta} = \set{1.0,  0.2, -0.8, 1.0, 0.5, -0.8, 0.6, -0.7, -0.6}$ and $\boldsymbol{\gamma} = \set{0.8, -1.0, -2.0, 1.5}$, i.e., the same coefficients as in the main text.

The results for the first two linear setups are very similar to those obtained in \Cref{subsection:Simulation:lm}. The correct \parametric\ approaches are the most accurate, while the \generative\ method class is generally the second-best class for moderate $\rho$. However, the \separatereg\ method class is the second-best class for higher values of $\rho$.

The results of the \texttt{gam\textunderscore one} and \texttt{gam\textunderscore two} experiments are almost identical to those obtained in the \texttt{lm\textunderscore no\textunderscore interactions} experiment, which is unsurprising as the setups are very similar. When we include one nonlinear term, the \texttt{LM separate} is the most accurate, but as expected, this changes when we include more nonlinear terms. In this case, the correct \parametric\ methods are the most accurate, together with the \texttt{GAM separate} method. Again, we see a tendency for the \separate\ methods to become more precise for higher dependence levels. The same holds for the \parametric\ methods. The results of the \texttt{gam\textunderscore five} experiment are nearly identical to those described for the \texttt{gam\textunderscore three} experiment in the main text. 

\begin{figure}[!ht]
    \centering
    \vspace{-11ex}
    \centerline{\includegraphics[width=1.09\textwidth]{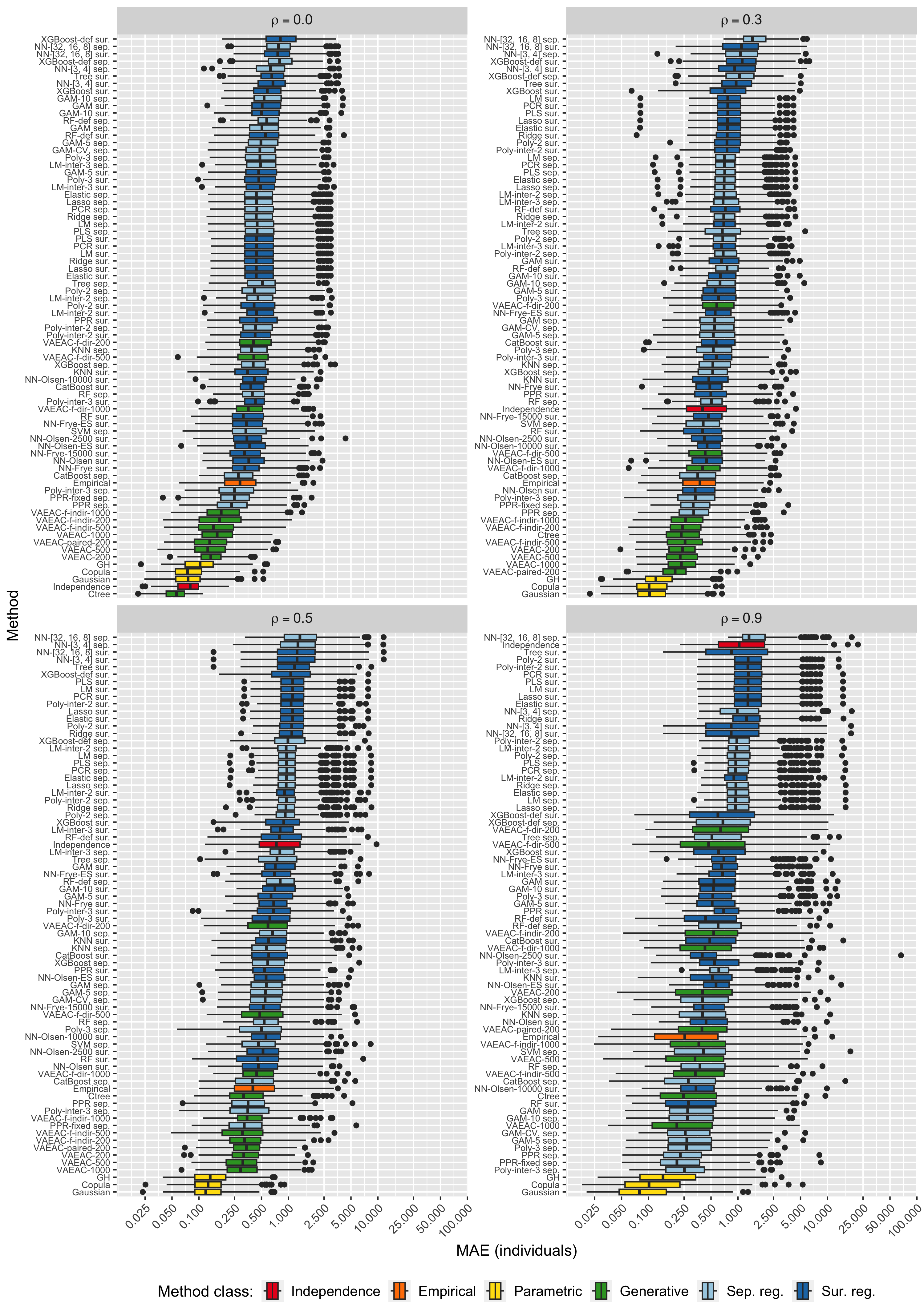}    \vspace{-1.5ex}}
    \caption{{\small Experiment \texttt{gam\textunderscore many\textunderscore interactions} with Gaussian data; see \Cref{fig:lm_no} for description.}}
    \label{fig:gam_many_interactions_gaussian}
\end{figure}

For the \texttt{gam\textunderscore some\textunderscore interactions} and \texttt{gam\textunderscore many\textunderscore interactions} experiments, we obtain indistinguishable results. Furthermore, the results also coincide with the results we observed in \Cref{subsection:Simulation:GAM}; see \Cref{fig:gam_many_interactions_gaussian}. Generally, the \parametric\ methods are superior, with the \generative\ methods as a close runner-up, except for $\rho = 0.9$, where the \separatereg\ methods constitutes the second-best method class, in particular, the \tm{PPR separate} method. Furthermore, using the \texttt{PPR-fixed separate} method introduced in \Cref{Appendix:Methods:Regression:ppr} drastically decreases the computational cost without sacrificing much precision.

\subsection{Burr Distributed Experiments}
\label{Appendix:ExtendedSimulations:Intermediate_burr}
In this section, we repeat the same simulation studies as in \Cref{sec:Numerical_Simulation_Studies} and \Cref{Appendix:ExtendedSimulations:Intermediate}, but we replace the multivariate Gaussian data with multivariate Burr data. The Burr distribution is strictly positive, heavy-tailed, skewed, and has nonlinear dependence and known conditional distributions; see \Cref{Appendix:AdditionalInformationParametric:Burr}.

\begin{figure}[!t]
    \centering
    \vspace{-11ex}
    \centerline{\includegraphics[width=1.09\textwidth]{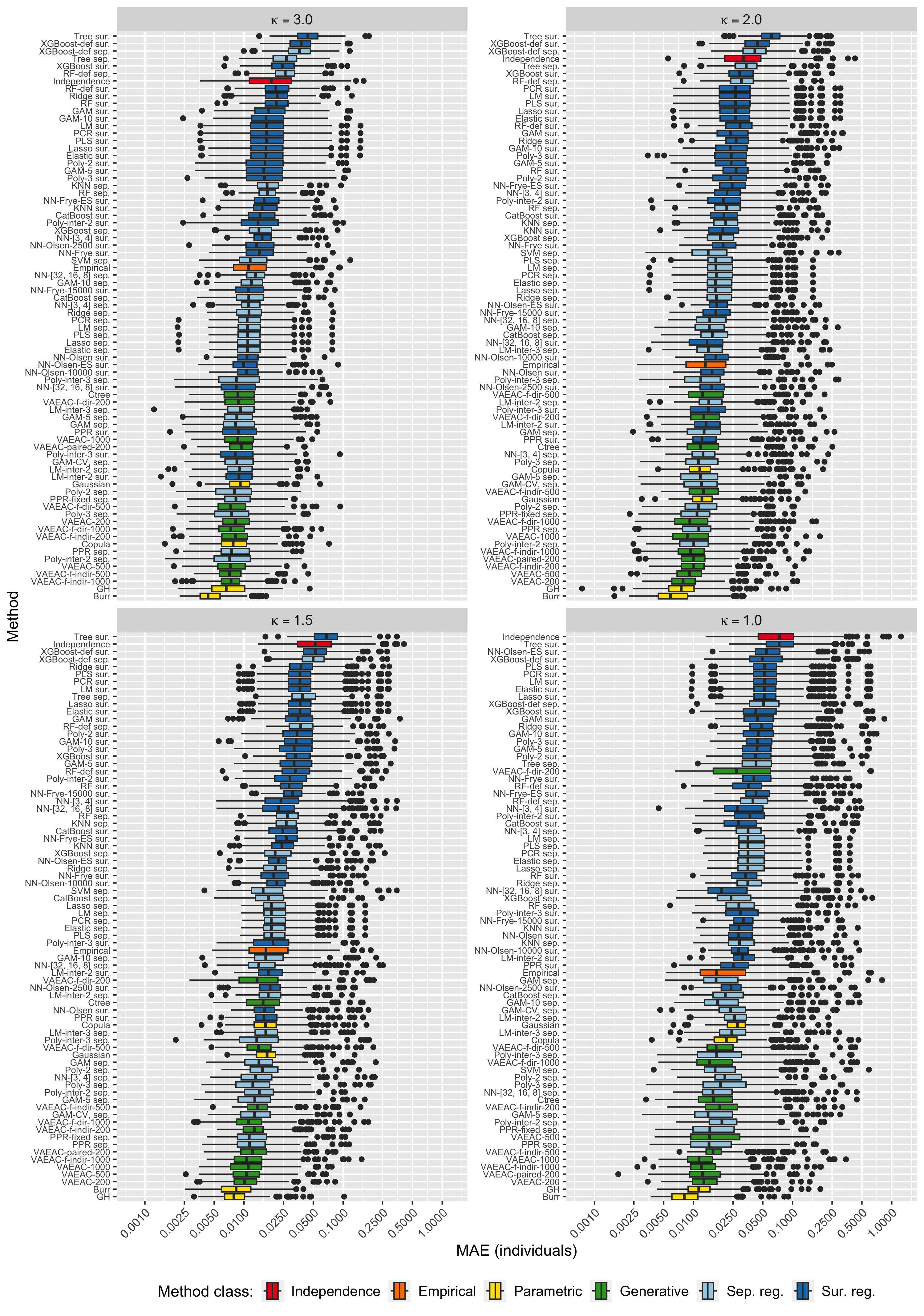}\vspace{-1.5ex}}
    \caption{{\small Experiment \texttt{lm\textunderscore more\tu interactions} with Burr data; see \Cref{fig:lm_no} for description.}}
    \label{fig:lm_more_interactions_burr}
\end{figure}

\begin{figure}[!t]
    \centering
    \vspace{-11ex}
    \centerline{\includegraphics[width=1.09\textwidth]{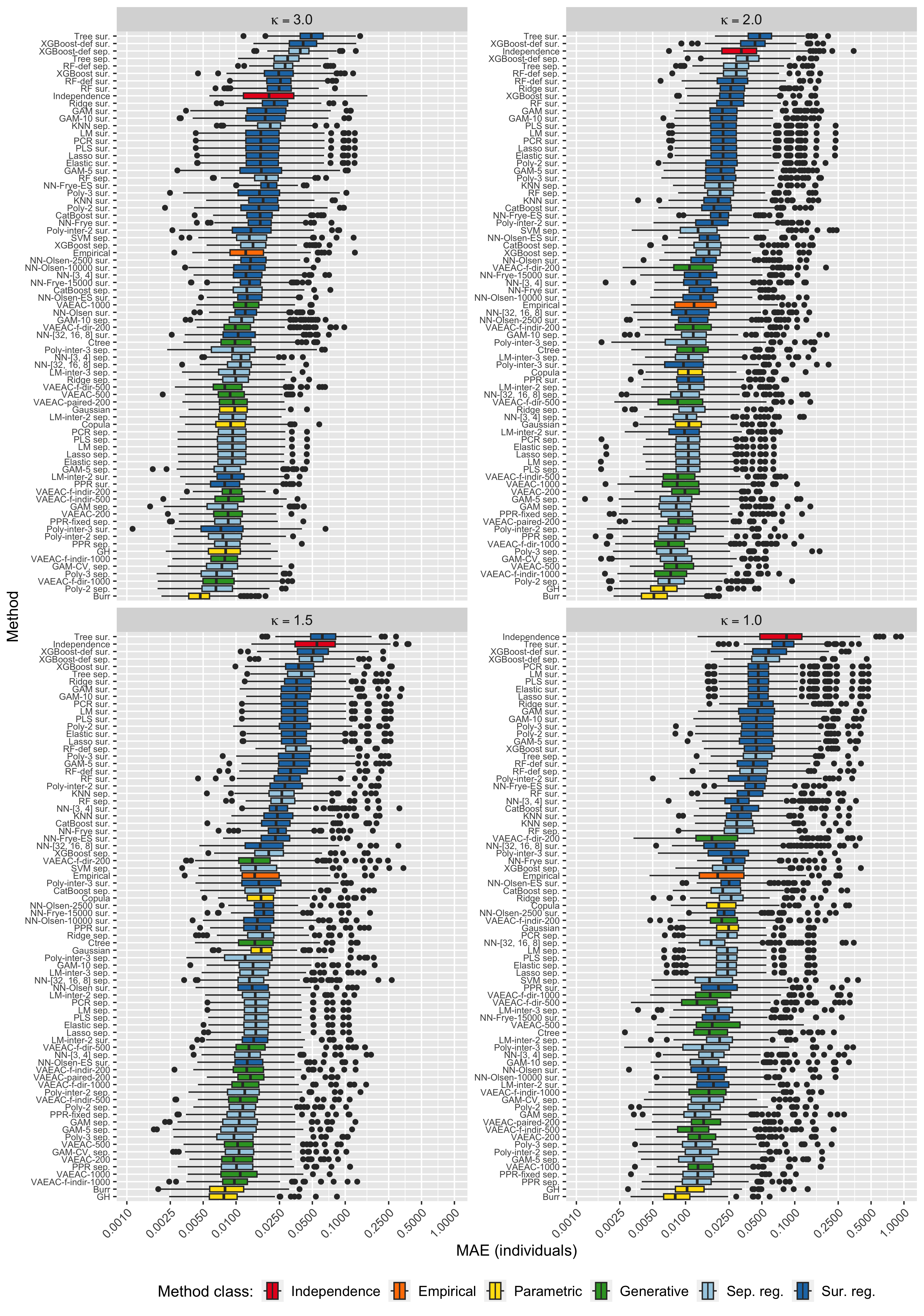}\vspace{-1.5ex}}
    \caption{{\small Experiment \texttt{gam\textunderscore three} with Burr data; see \Cref{fig:lm_no} for description.}}
    \label{fig:gam_three_burr}
\end{figure}

\begin{figure}[!t]
    \centering
    \vspace{-11ex}
    \centerline{\includegraphics[width=1.09\textwidth]{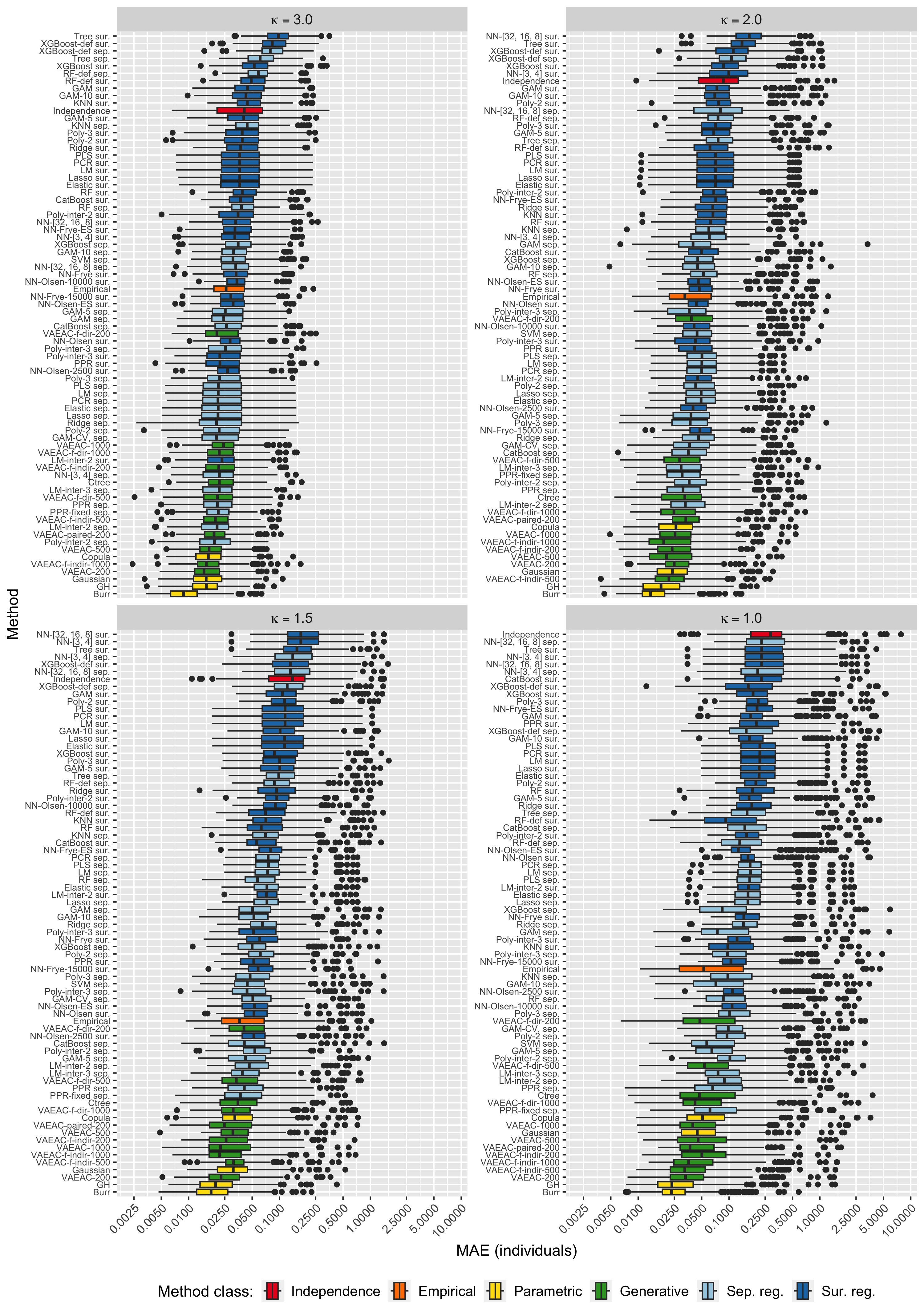}\vspace{-1.5ex}}
    \caption{{\small Experiment \texttt{gam\textunderscore more\tu interactions} with Burr data; see \Cref{fig:lm_no} for description.}}
    \label{fig:gam_more_interactions_burr}
\end{figure}

We sample $N_\text{train} = 1000$ training and $N_\text{test} = 250$ test observations from a $\operatorname{Burr}(\kappa, \boldsymbol{b}, \boldsymbol{r})$ distribution. We let $\boldsymbol{b} = \set{5, 4, 6, 5, 3, 6, 5, 5}$, $\boldsymbol{r} = \set{4, 3, 5, 2, 5, 3, 5, 1}$, while we vary the scale parameter $\kappa \in \set{0.5, 1.0, 1.5, 2.0, 2.5, 3.0}$. Here, a low $\kappa$ indicates high dependency. The average Pearson correlations for the six values of $\kappa$ are $0.80$, $0.63$, $0.46$, $0.36$, $0.29$, and $0.25$, respectively. In \Cref{fig:pairs:Burr1,fig:pairs:Burr3} in \Cref{Appendix:PairPlots}, we display plots of the Burr data when $\kappa = 1$ and $\kappa = 3$, respectively. A larger value of $\kappa$ makes the Burr distribution more Gaussian-like, while lower values of $\kappa$ produce more extreme observations due to the right heavy-tailed property of the Burr distribution. We observe that the methods struggle with these extreme observations, and the individual MAE is (often) higher for these observations in the outer region of the data distribution. That is, the data has few similar observations for the methods to learn the conditional structure. 
In \Cref{fig:lm_more_interactions_burr,fig:gam_three_burr,fig:gam_more_interactions_burr}, we present a selection of the results. The results for the other settings are very similar.

We observe similar results for the Burr data as we did for the Gaussian data. That is, using the correct \parametric\ approach, in this case, the \texttt{Burr} approach, yields the most accurate Shapley value explanations. The \tm{GH} method also performs well, even though it makes an incorrect parametric assumption, while the \texttt{Gaussian} and \copula\ methods perform relatively worse. The best method outside the \parametric\ method class is generally a \vaeac\ approach, where the number in the name indicates the maximum number of epochs. However, for the less complex setups without interaction terms, the \vaeac\ approaches are often outperformed by some of the \separate\ methods, particularly the \texttt{GAM separate} and \tm{PPR separate} approaches. We do not see a systematic benefit of choosing a large value for the maximum number of epochs in the \vaeac\ approaches. Thus, the differences are likely based on better initialized random weights in the networks, similar to what we discuss for the \tm{NN-Olsen surrogate} method in \Cref{Appendix:RealWorld:NN}. The \vaeacfdir\ approaches are consistently outperformed by the \vaeac\ and \vaeacfindir\ methods, and there does not seem to be a systematic winner between the later two methods. Some of the additional regression-based methods proposed in \Cref{Appendix:Methods:Regression} perform relatively well, such as the proposed \tm{LM-inter separate} and \tm{Poly-inter separate} methods.

For the \surrogate\ approaches, we notice that the \tm{LM-inter surrogate} and \tm{Poly-inter surrogate} often outperform the complex \tm{NN-Frye surrogate} and \tm{NN-Olsen surrogate} approaches. However, for the complex \texttt{gam\textunderscore more\tu interactions} experiment, the \tm{NN-Olsen surrogate} approach with a high number of epochs is the overall best \surrogate\ approach. For the complex setups in \Cref{subsection:Simulation:GAM} and \Cref{Appendix:ExtendedSimulations:Intermediate} with interactions, we also let the predictive model $f$ be a random forest and a projection pursuit regression model, as done in \Cref{Numerical_Simulation_Studies:OtherPredictiveModels}. This had a minor effect on the overall results. However, some of the non-smooth regression methods, such as \tm{CatBoost separate}, performed relatively better when $f$ was also non-smooth, just as in the main text. We also looked at $N_\text{train} \in \set{100, 5000}$, and the overall tendencies remained.

\begin{table}[t]
\centering
\begin{adjustbox}{max width=1\textwidth}
\begin{tabular}{lrclrclr}
\toprule
Method          &   Time & & Method                      &   Time & & Method                       &    Time \\
\cmidrule{1-2} \cmidrule{4-5} \cmidrule{7-8}
\cellcolor{col_ind!10}\tm{Independence}       & 36:51.9\cellcolor{col_ind!10} & & \cellcolor{col_sep!10}\tm{Poly-inter-3 sep.}   &     3.1\cellcolor{col_sep!10} & & \cellcolor{col_sur!10}\tm{Poly-inter-2 sur.}   &        7.0\cellcolor{col_sur!10} \\ 
\cellcolor{col_emp!10}\tm{Empirical}          & 13:17.6\cellcolor{col_emp!10} & & \cellcolor{col_sep!10}\tm{GAM sep.}            &    43.7\cellcolor{col_sep!10} & & \cellcolor{col_sur!10}\tm{Poly-inter-3 sur.}   &       17.0\cellcolor{col_sur!10} \\
\cellcolor{col_par!10}\tm{Gaussian}           & 35:59.0\cellcolor{col_par!10} & & \cellcolor{col_sep!10}\tm{GAM-5 sep.}          &    26.6\cellcolor{col_sep!10} & & \cellcolor{col_sur!10}\tm{GAM sur.}            &       22.8\cellcolor{col_sur!10} \\
\cellcolor{col_par!10}\tm{Copula}             & 39:14.1\cellcolor{col_par!10} & & \cellcolor{col_sep!10}\tm{GAM-10 sep.}         &  1:37.7\cellcolor{col_sep!10} & & \cellcolor{col_sur!10}\tm{GAM-5 sur.}          &       12.3\cellcolor{col_sur!10} \\
\cellcolor{col_par!10}\tm{GH}                 & 42:38.5\cellcolor{col_par!10} & & \cellcolor{col_sep!10}\tm{GAM-CV sep.}         & 25:43.0\cellcolor{col_sep!10} & & \cellcolor{col_sur!10}\tm{GAM-10 sur.}         &       55.0\cellcolor{col_sur!10} \\
\cellcolor{col_par!10}\tm{Burr}               & 38:30.5\cellcolor{col_par!10} & & \cellcolor{col_sep!10}\tm{PCR sep.}            &     3.9\cellcolor{col_sep!10} & & \cellcolor{col_sur!10}\tm{PCR sur.}            &       14.9\cellcolor{col_sur!10} \\ 
\cellcolor{col_gen!10}\tm{Ctree}              & 18:45.6\cellcolor{col_gen!10} & & \cellcolor{col_sep!10}\tm{PLS sep.}            &     3.3\cellcolor{col_sep!10} & & \cellcolor{col_sur!10}\tm{PLS sur.}            &       14.1\cellcolor{col_sur!10} \\
\cellcolor{col_gen!10}\tm{VAEAC-200}          & 40:20.1\cellcolor{col_gen!10} & & \cellcolor{col_sep!10}\tm{PPR sep.}            &  1:55.4\cellcolor{col_sep!10} & & \cellcolor{col_sur!10}\tm{PPR sur.}            &     4:29.9\cellcolor{col_sur!10} \\
\cellcolor{col_gen!10}\tm{VAEAC-500}          & 41:58.2\cellcolor{col_gen!10} & & \cellcolor{col_sep!10}\tm{PPR-fixed sep.}      &     4.6\cellcolor{col_sep!10} & & \cellcolor{col_sur!10}\tm{KNN sur.}            &    38:10.4\cellcolor{col_sur!10} \\
\cellcolor{col_gen!10}\tm{VAEAC-1000}         & 44:03.8\cellcolor{col_gen!10} & & \cellcolor{col_sep!10}\tm{SVM sep.}            &    14.0\cellcolor{col_sep!10} & & \cellcolor{col_sur!10}\tm{Tree sur.}           &       13.0\cellcolor{col_sur!10} \\
\cellcolor{col_gen!10}\tm{VAEAC-paired-200}   & 52:54.4\cellcolor{col_gen!10} & & \cellcolor{col_sep!10}\tm{KNN sep.}            &    18.8\cellcolor{col_sep!10} & & \cellcolor{col_sur!10}\tm{RF sur.}             &    55:44.6\cellcolor{col_sur!10} \\
\cellcolor{col_gen!10}\tm{VAEAC-f-indir-200}  & 40:53.8\cellcolor{col_gen!10} & & \cellcolor{col_sep!10}\tm{Tree sep.}           &     3.5\cellcolor{col_sep!10} & & \cellcolor{col_sur!10}\tm{RF-def sur.}         &     2:19.1\cellcolor{col_sur!10} \\
\cellcolor{col_gen!10}\tm{VAEAC-f-indir-500}  & 42:04.1\cellcolor{col_gen!10} & & \cellcolor{col_sep!10}\tm{RF sep.}             & 56:04.1\cellcolor{col_sep!10} & & \cellcolor{col_sur!10}\tm{XGBoost sur.}        &  2:02:39.5\cellcolor{col_sur!10} \\
\cellcolor{col_gen!10}\tm{VAEAC-f-indir-1000} & 44:13.9\cellcolor{col_gen!10} & & \cellcolor{col_sep!10}\tm{RF-def sep.}         &    40.0\cellcolor{col_sep!10} & & \cellcolor{col_sur!10}\tm{XGBoost-def sur.}    &       50.7\cellcolor{col_sur!10} \\
\cellcolor{col_gen!10}\tm{VAEAC-f-dir-200}    &  5:05.3\cellcolor{col_gen!10} & & \cellcolor{col_sep!10}\tm{XGBoost sep.}        & 26:32.7\cellcolor{col_sep!10} & & \cellcolor{col_sur!10}\tm{CatBoost sur.}       &       34.6\cellcolor{col_sur!10} \\
\cellcolor{col_gen!10}\tm{VAEAC-f-dir-500}    &  6:46.0\cellcolor{col_gen!10} & & \cellcolor{col_sep!10}\tm{XGBoost-def sep.}    &    33.5\cellcolor{col_sep!10} & & \cellcolor{col_sur!10}\tm{NN-[3, 4] sur.}      &     3:48.7\cellcolor{col_sur!10} \\
\cellcolor{col_gen!10}\tm{VAEAC-f-dir-1000}   &  8:54.0\cellcolor{col_gen!10} & & \cellcolor{col_sep!10}\tm{CatBoost sep.}       &  4:47.9\cellcolor{col_sep!10} & & \cellcolor{col_sur!10}\tm{NN-[32, 16, 8] sur.} &     7:44.2\cellcolor{col_sur!10} \\
\cellcolor{col_sep!10}\tm{LM sep.}            &     0.6\cellcolor{col_sep!10} & & \cellcolor{col_sep!10}\tm{NN-[3, 4] sep.}      &  1:32.1\cellcolor{col_sep!10} & & \cellcolor{col_sur!10}\tm{NN-Frye sur.}        & 13:44:46.7\cellcolor{col_sur!10} \\
\cellcolor{col_sep!10}\tm{LM-inter-2 sep.}    &     0.7\cellcolor{col_sep!10} & & \cellcolor{col_sep!10}\tm{NN-[32, 16, 8] sep.} &  5:13.1\cellcolor{col_sep!10} & & \cellcolor{col_sur!10}\tm{NN-Frye-15000 sur.}  & 14:53:22.9\cellcolor{col_sur!10} \\
\cellcolor{col_sep!10}\tm{LM-inter-3 sep.}    &     0.8\cellcolor{col_sep!10} & & \cellcolor{col_sur!10}\tm{LM sur.}             &     2.7\cellcolor{col_sur!10} & & \cellcolor{col_sur!10}\tm{NN-Frye-ES sur.}     &    44:08.3\cellcolor{col_sur!10} \\
\cellcolor{col_sep!10}\tm{Lasso sep.}         &    15.5\cellcolor{col_sep!10} & & \cellcolor{col_sur!10}\tm{LM-inter-2 sur.}     &     8.4\cellcolor{col_sur!10} & & \cellcolor{col_sur!10}\tm{NN-Olsen sur.}       &  7:15:32.6\cellcolor{col_sur!10} \\
\cellcolor{col_sep!10}\tm{Ridge sep.}         &    17.3\cellcolor{col_sep!10} & & \cellcolor{col_sur!10}\tm{Lasso sur.}          &     5.2\cellcolor{col_sur!10} & & \cellcolor{col_sur!10}\tm{NN-Olsen-2500 sur.}  &  3:58:45.2\cellcolor{col_sur!10} \\
\cellcolor{col_sep!10}\tm{Elastic sep.}       &    15.9\cellcolor{col_sep!10} & & \cellcolor{col_sur!10}\tm{Ridge sur.}          &     6.5\cellcolor{col_sur!10} & & \cellcolor{col_sur!10}\tm{NN-Olsen-10000 sur.} & 15:22:43.1\cellcolor{col_sur!10} \\
\cellcolor{col_sep!10}\tm{Poly-2 sep.}        &     0.4\cellcolor{col_sep!10} & & \cellcolor{col_sur!10}\tm{Elastic sur.}        &     4.4\cellcolor{col_sur!10} & & \cellcolor{col_sur!10}\tm{NN-Olsen-ES sur.}    &    26:40.1\cellcolor{col_sur!10} \\
\cellcolor{col_sep!10}\tm{Poly-3 sep.}        &     1.7\cellcolor{col_sep!10} & & \cellcolor{col_sur!10}\tm{Poly-2 sur.}         &     4.4\cellcolor{col_sur!10} & & \\
\cellcolor{col_sep!10}\tm{Poly-inter-2 sep.}  &     1.9\cellcolor{col_sep!10} & & \cellcolor{col_sur!10}\tm{Poly-3 sur.}         &     5.1\cellcolor{col_sur!10} & & \\
\bottomrule
\end{tabular}
\end{adjustbox}
\caption{{\small The CPU times used by the methods to compute Shapley values for the $N_\text{test} = 250$ test observations in the \texttt{gam\tu more\tu interactions} experiment with Burr distributed features when $\kappa = 2$ and $N_\text{train} = 1000$ (\Cref{Appendix:ExtendedSimulations:Intermediate_burr}). The format of the CPU times is hours:minutes:seconds, where we omit the larger units of time if they are zero, and the colors indicate the different method classes.}}
\label{tab:SimStudyTimeColBurr}
\end{table}

In \Cref{tab:SimStudyTimeColBurr}, we report the CPU times for the approaches used in the \texttt{gam\tu more\tu interactions} experiment with Burr distributed features, $\kappa = 2$, $N_\text{train} = 1000$, and $N_\text{test} = 250$. We see similar time tendencies as discussed in \Cref{Numerical_Simulation_Studies:Time}, but note the time differences between the \separate\ methods with default and cross-validated versions. For example, the \tm{RF separate} takes approximately $84$ times longer than the \tm{RF-def separate} method due to the extra cost of tuning the hyperparameters. However, the \tm{RF separate} method obtains lower MAE scores; see \Cref{fig:gam_more_interactions_burr}. We observe similar tendencies for the \tm{PPR separate} and \tm{PPR-fixed separate} methods. Here, we only report the total time, but the time decomposition is similar to the one in \Cref{tab:SimStudyTimeCol}. That is, the \separate\ and \surrogate\ methods spend the vast majority of their time on training, while the predicting step only takes a couple of seconds. In contrast, the Monte Carlo-based approaches spend most of the time on the predicting and generating steps in that order.

\clearpage
\newpage

\section{Characteristics of the Data Sets}
\label{Appendix:PairPlots}
\begin{figure}[!t]
    \centering
    \vspace{-8ex}
    \centerline{\includegraphics[width=1.2\textwidth]{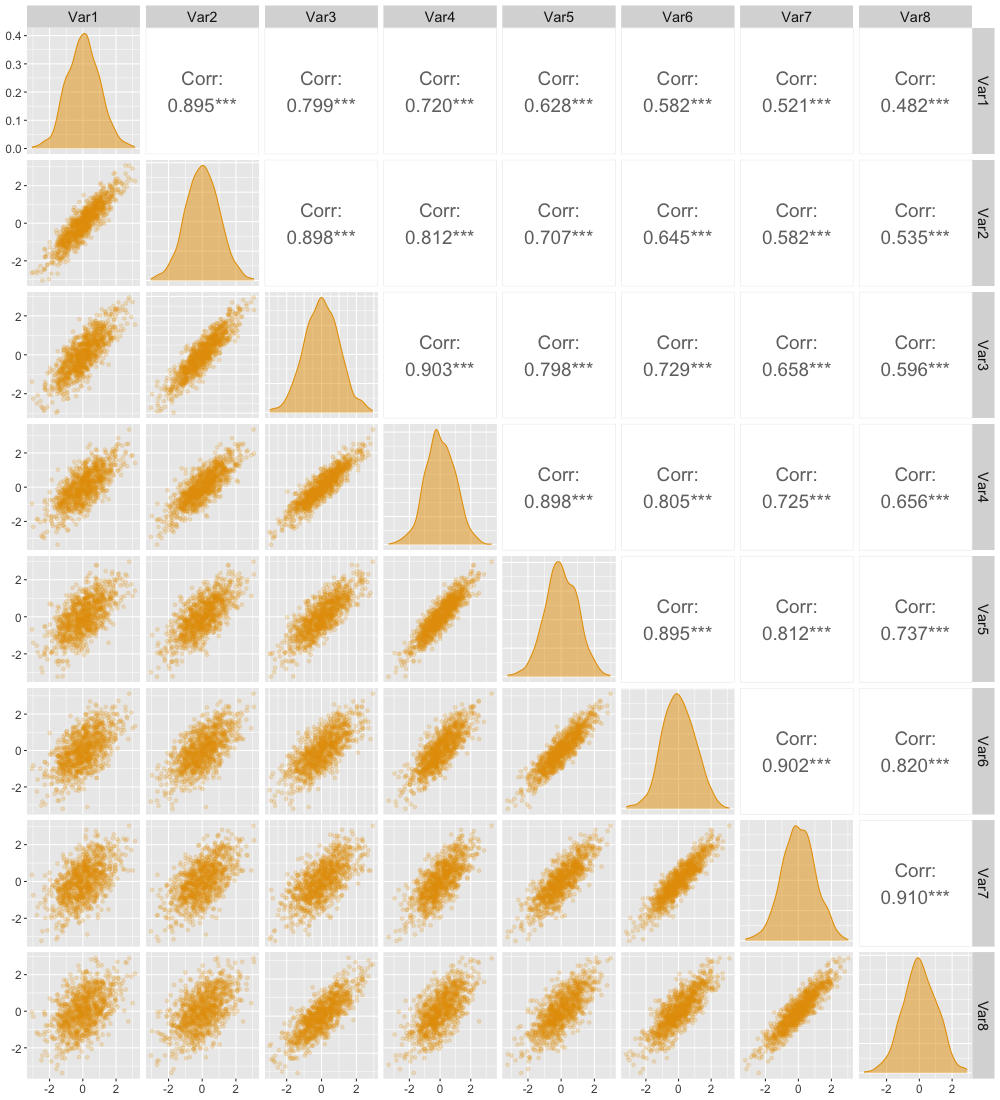}}
    \caption{{\small Pairwise scatter plots, marginal density functions, and pairwise Pearson correlation coefficients for the \textbf{simulated Gaussian distributed features} in \Cref{sec:Numerical_Simulation_Studies} with $\rho = 0.9$. The correlation between feature $i$ and $j$ is given by $\rho^{|i-j|}$. }}
    \label{fig:pairs:Gaussian}
\end{figure}

\begin{figure}[!t]
    \centering
    \vspace{-8ex}
    \centerline{\includegraphics[width=1.2\textwidth]{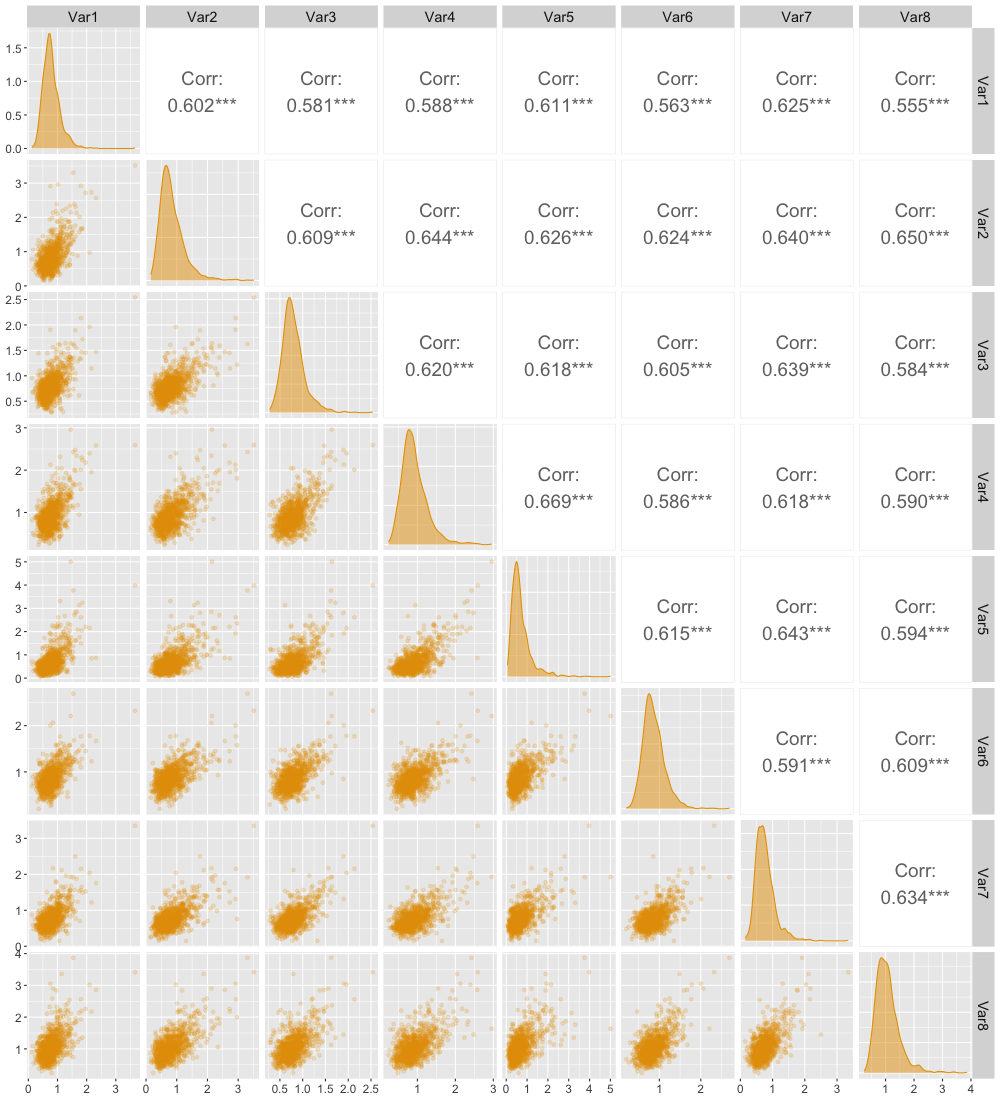}}
    \caption{{\small Pairwise scatter plots, marginal density functions, and pairwise Pearson correlation coefficients for the \textbf{simulated Burr distributed features} in \Cref{Appendix:ExtendedSimulations} with $\kappa = 1$.}}
    \label{fig:pairs:Burr1}
\end{figure}

\begin{figure}[!t]
    \centering
    \vspace{-8ex}
    \centerline{\includegraphics[width=1.2\textwidth]{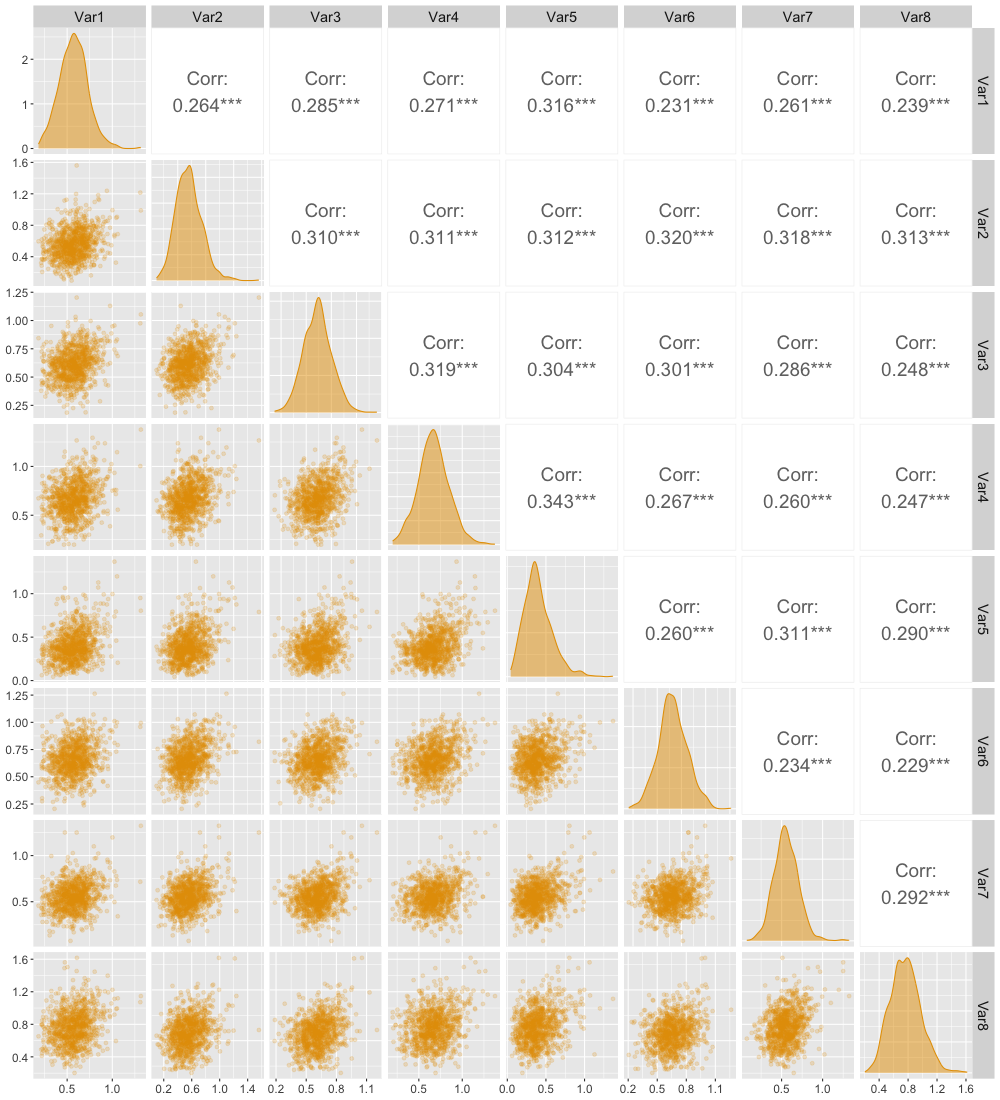}}
    \caption{{\small Pairwise scatter plots, marginal density functions, and pairwise Pearson correlation coefficients for the \textbf{simulated Burr distributed features} in \Cref{Appendix:ExtendedSimulations} with $\kappa = 3$.}}
    \label{fig:pairs:Burr3}
\end{figure}

\begin{figure}[!t]
    \centering
    \vspace{-8ex}
    \centerline{\includegraphics[width=1.2\textwidth]{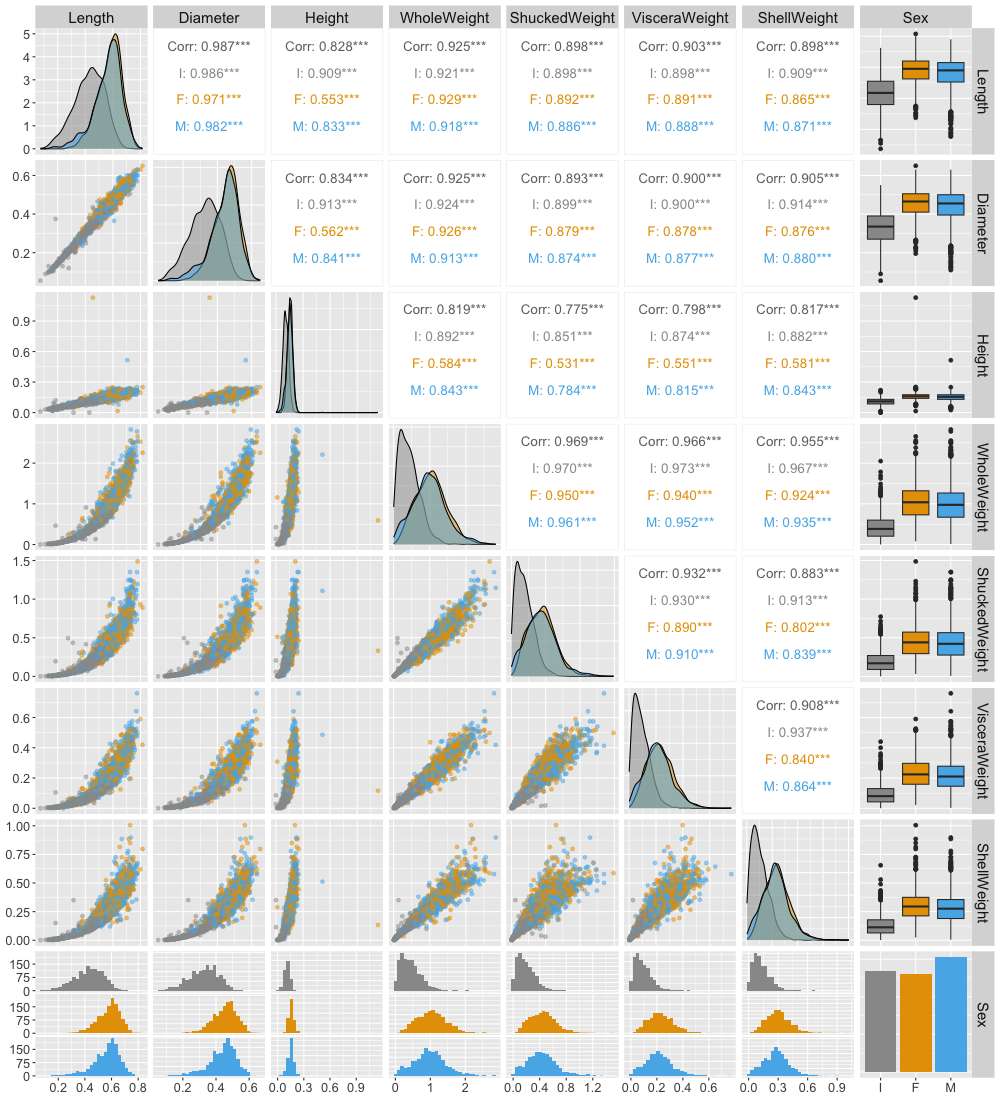}}
    \caption{{\small Pairwise scatter plots, marginal density functions, and pairwise Pearson correlation coefficients for the features in the \textbf{Abalone data set} used in \Cref{sec:real_world_data}. The figure is grouped by \texttt{Sex}, where the infants are gray, females are yellow, and males are blue. The correlations reported in black correspond to all observations, while the colored correlations are grouped based on \texttt{Sex}.}}
    \label{fig:pairs:Abalone}
\end{figure}

\begin{figure}[!t]
    \centering
    \vspace{-8ex}
    \centerline{\includegraphics[width=1.2\textwidth]{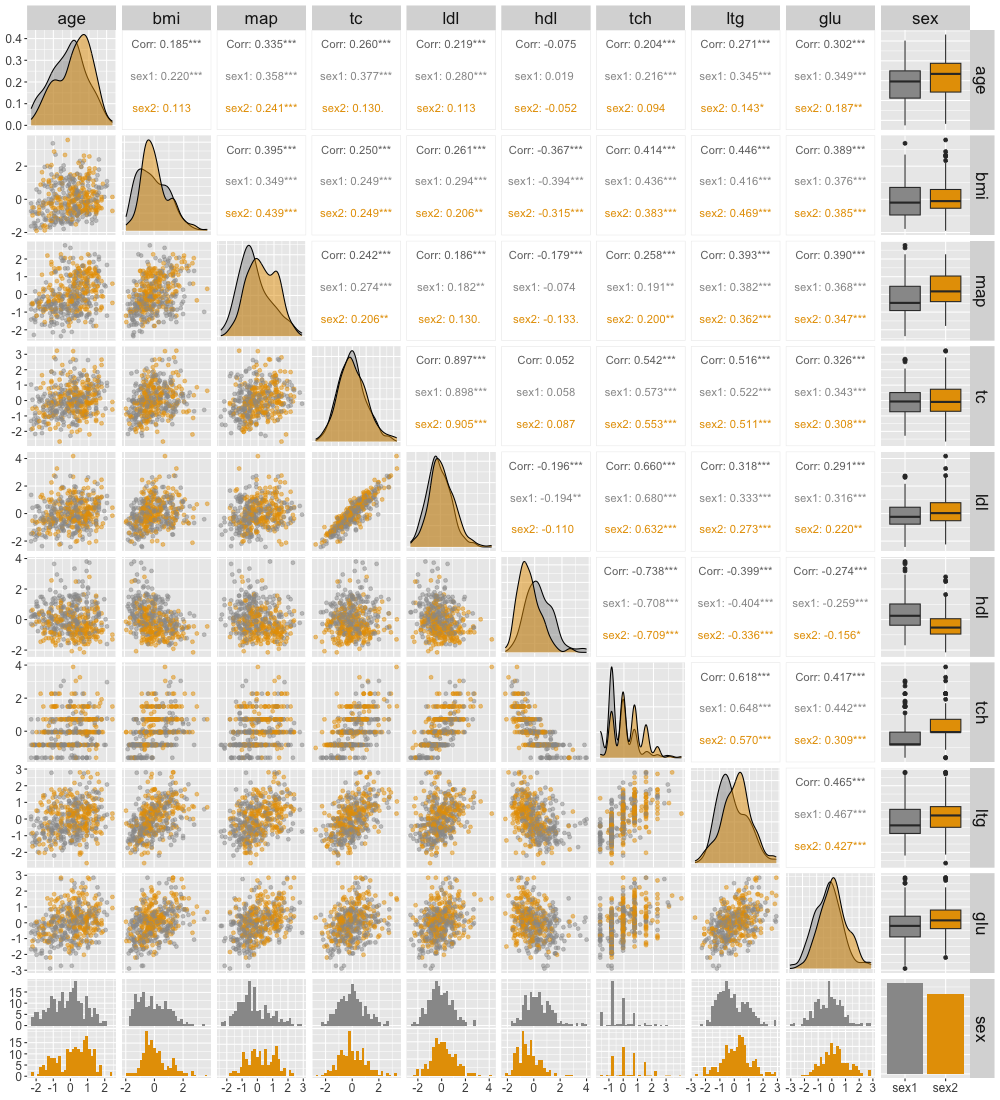}}
    \caption{{\small Pairwise scatter plots, marginal density functions, and pairwise Pearson correlation coefficients for the features in the \textbf{Diabetes data set} used in \Cref{sec:real_world_data}. The figure is grouped by \texttt{Sex}. The correlations reported in black correspond to all observations, while the colored correlations are grouped based on \texttt{Sex}.}}
    \label{fig:pairs:Diabetes}
\end{figure}

\begin{figure}[!t]
    \centering
    \vspace{-8ex}
    \centerline{\includegraphics[width=1.2\textwidth]{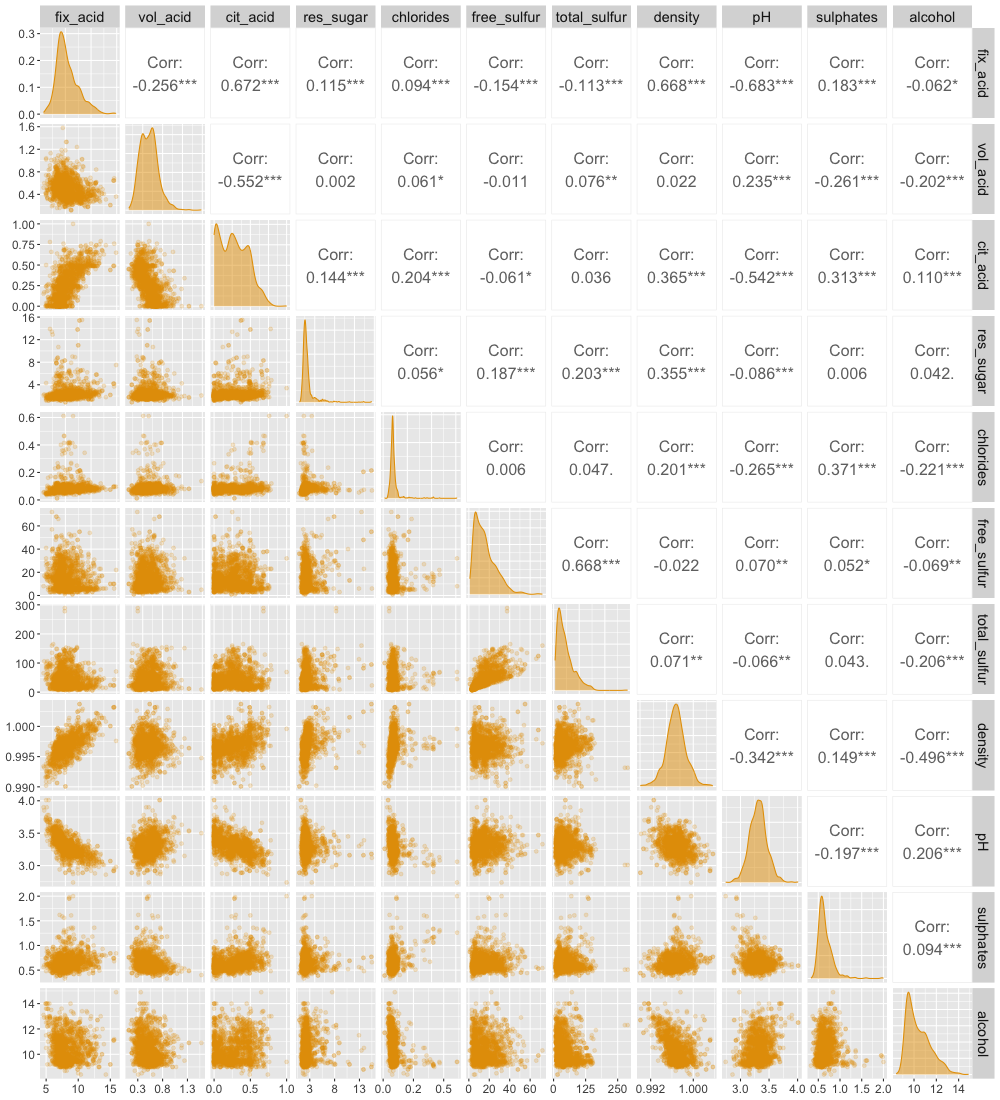}}
    \caption{{\small Pairwise scatter plots, marginal density functions, and pairwise Pearson correlation coefficients for the features in the \textbf{(Red) Wine data set} used in \Cref{sec:real_world_data}.}}
    \label{fig:pairs:Wine}
\end{figure}

In this section, we provide pairwise scatter plots, marginal density functions, and pairwise Pearson correlation coefficients (for the continuous features) between the features in some of the data sets we have used in this article. In \Cref{fig:pairs:Gaussian,fig:pairs:Burr1,fig:pairs:Burr3}, we include figures for a few of the simulated Gaussian and Burr distributed data sets from the numerical simulation studies in \Cref{sec:Numerical_Simulation_Studies} and \Cref{Appendix:ExtendedSimulations}, respectively. Further, plots for the Abalone, Diabetes, and Wine data sets from \Cref{sec:real_world_data} are provided in \Cref{fig:pairs:Abalone,fig:pairs:Diabetes,fig:pairs:Wine}, respectively. We omit the Adult dataset as it is chaotic and difficult to interpret due to the large number of levels in the categorical features. However, the pairwise Pearson correlation coefficients for the continuous features are all close to zero. 

In \Cref{fig:pairs:Gaussian,fig:pairs:Burr1,fig:pairs:Burr3}, we include plots for the Gaussian distribution with $\rho = 0.9$ and Burr distribution with $\kappa \in {1, 3}$, respectively. For the Gaussian features, the pairwise correlation between features $i$ and $j$ is given by $\rho^{|i-j|}$, i.e., it decreases from $0.9$ to $0.48$ in the $M=8$ situation. For the Burr distribution, the marginals become more right-skewed, and the correlation approximately doubles when we decrease $\kappa$ from $3$ to $1$. The average correlations are $0.28$ and $0.61$ for $\kappa = 3$ and $\kappa = 1$, respectively.

For the Abalone data set (\Cref{fig:pairs:Abalone}), there is a clear nonlinearity and heteroscedasticity among the pairs of features and a significant pairwise correlation between the features. All continuous features have a pairwise correlation above $0.775$, or $0.531$ when grouped by the categorical feature \texttt{Sex}. There is a clear distinction between infants and females/males. All marginals are right-skewed, except for the features \texttt{Length} and \tm{Diameter}, which is left-skewed.

The Diabetes data set (\Cref{fig:pairs:Diabetes}) shows a fairly strong correlation between many features. For example, the correlation between \texttt{S1} and \tm{S2} is $0.90$. On average, the \tm{Age} feature is the least correlated feature with the other features. Most scatter plots and marginal distributions display structures and marginals somewhat similar to Gaussian distribution, except the \tm{S4} feature, which has a multi-modal marginal.

In contrast, for the (Red) Wine dataset (\Cref{fig:pairs:Wine}), most scatter plots and marginal density functions display structures and marginals far from the Gaussian distribution, as most marginals are right-skewed. The largest correlation in absolute value is $0.683$ (between \texttt{pH} and \tm{fix\_acid}), while most other pairs of features have no to moderate correlation.

\clearpage

\section{Real-World Data Experiments: Computation Time}
\label{Appendix:RealWorld}
\Cref{tab:SimStudyTimeAbaloneCont,tab:SimStudyTimeAbaloneAll,tab:SimStudyTimeDiabetes,tab:SimStudyTimeWine,tab:SimStudyTimeAdult} present the decomposed CPU times for the $\text{Abalone}$, Diabetes, Wine, and Adult experiments from \Cref{sec:real_world_data}, respectively. The tables also include the $\operatorname{MSE}_v$ scores for the different methods. The total CPU times are decomposed into the same three categories as in \Cref{Numerical_Simulation_Studies:Time}: \textit{training}, \textit{generating}, and \textit{predicting}. Recall that we ran the three first experiments on the MAC system specified in \Cref{Numerical_Simulation_Studies:Time}, while the Adult experiment was run on a shared computer server described in \Cref{sec:real_world_data}. The CPU times will differ from computer to computer. 

Here, we include some methods in addition to those used in \Cref{sec:real_world_data}, such as the \tm{PCR separate} approach (\Cref{Appendix:Methods:Regression:pcr}), which performs similar to the best methods for the Diabetes experiment. Recall that the predictive model $f$ in the Diabetes experiment is a PCR model. Hence, this supports our findings in the main text that we should use a \separate\ method with the same form as $f$ for accurate Shapley value estimates. Additionally, we include the \tm{RF-def separate} method (\Cref{Appendix:Methods:Regression:rf}) with default hyperparameters to illustrate the need for conducting cross-validation to tune the hyperparameters. We also include versions of the \vaeac, \tm{NN-Olsen surrogate}, and \tm{NN-Frye surrogate} approaches with default hyperparameters but with different numbers of training epochs. It should also be noted that the \tm{PPR-fixed separate} method with a fixed number of terms $L = |\s|$ produces almost as good results as the CV-alternative \tm{PPR separate} in a fraction of the time.


As in \Cref{Numerical_Simulation_Studies:Time}, the training step is the most time-consuming step for the \separate\ and \surrogate\ methods, while the predicting step often takes only a couple of seconds. For the Abalone and Diabetes datasets, creating the augmented training data set takes approximately 2 seconds, while it takes approximately 14 seconds for the Wine data set. The time increase is due to larger a $M$ (compared to Abalone) and $N_\text{train}$ (compared to Diabetes). In comparison, the Monte Carlo-based methods use most of their time generating the Monte Carlo samples in the $\text{Abalone}$ and Diabetes experiments. This contrasts with the timings in \Cref{tab:SimStudyTimeCol}, where the predicting step was the slowest. The time difference is caused by the GAM model in the \tm{gam\tu more\tu interactions} experiment being more computationally expensive to call than the PPR and PCR models in $\text{Abalone}$ and Diabetes experiments, respectively. For the Wine experiment, the predicting step is the most expensive, which is caused by the RF model being more computationally costly to call, and we have more calls due to a larger $M$ (compared to Abalone) and $N_\text{test}$ (compared to Diabetes). 

When excluding the training time, which is only done once and can be considered an upfront time cost, it is evident that the regression-based methods are superior with respect to computation time. For example, consider the best Monte Carlo and regression-based methods for the $\text{Abalone}_\text{cont}$ experiment, i.e., the \tm{VAEAC-10000} and \tm{PPR separate} methods, respectively. The \tm{VAEAC-10000} approach uses $551.9$ seconds to explain $1044$ predictions, an average of $0.53$ seconds per explanation. In contrast, the \tm{PPR separate} method explains all the $1044$ predictions in $0.5$ seconds. Thus, there is a speed difference of a factor of $1104$, which is essential when the number of predictions to explain is large. 

If training time is not a limiting factor, we can use more time to train the \tm{NN-Olsen surrogate} and \tm{NN-Frye surrogate} methods, as these methods are slow to train but fast in the predicting step. In contrast to the numerical simulation studies in \Cref{sec:Numerical_Simulation_Studies}, the validation errors for some of the complex real-world experiments were still decreasing for these approaches, indicating that more training would be beneficial. \Cref{tab:SimStudyTimeAbaloneCont} shows that the $\operatorname{MSE}_v$ scores for the different versions of the \tm{NN-Olsen surrogate} approach decrease when we increase the number of epochs leading it to share the first place with the \tm{PPR separate} approach. However, recall that we use the network at the epoch with the lowest validation error, which was the $6457$th epoch for the \tm{NN-Olsen-20000 surrogate} approach. This means that the increased performance was not due to the additional number of training epochs, but rather to better random initialization values which caused the network parameters to converge to a better local optimum. The same tendency also holds for the other real-world experiments. For example, there is essentially no difference in the $\operatorname{MSE}_v$ scores of the \tm{NN-Olsen-500 surrogate} and \tm{NN-Olsen-10000 surrogate} methods for the Diabetes data set in \Cref{tab:SimStudyTimeDiabetes}. Furthermore, the validation data is randomly extracted and removed from the training data for each \tm{NN surrogate} method. Thus, it might be that for some \tm{NN surrogate} methods and data sets, we were unlucky in that the training and validation data were not representative of the test data. This is more likely to happen for small data sets.


\Cref{tab:SimStudyTimeDiabetes} shows that the \tm{RF-def separate} approach with default hyperparameters provides almost as low $\operatorname{MSE}_v$ score as the cross-validated counter-version \tm{RF separate}, whose training time is approximately $140$ times longer. For the \surrogate\ version, we see that the \tm{RF-def surrogate} method outperforms the \tm{RF surrogate} even though the default hyperparameters are an option in the cross-validation procedure. 




\subsection{Analysis of the \tm{NN-Olsen surrogate} method for the Abalone Data Set}
\label{Appendix:RealWorld:NN}
In this section, we look closer at the effect of initialization values and hyperparameters for the \tm{NN-Olsen surrogate} method for the two experiments on the Abalone data set.

For the $\text{Abalone}_\text{all}$ experiment, we fitted ten versions of the \tm{NN-Olsen-10000 surrogate} method with different idealizations seeds. These networks were trained on the shared computer server described in \Cref{sec:real_world_data} and had an average training time of 14:02:46.6. In comparison, the training CPU time for the \tm{NN-Olsen-10000 surrogate} method in \Cref{tab:SimStudyTimeAbaloneAll} was almost $3.5$ times higher when trained on the MAC operating system described in \Cref{Numerical_Simulation_Studies:Time}. The average $\operatorname{MSE}_v$ was $1.214$, with a standard deviation of $0.0085$. Furthermore, on average, the best epoch was the $5509$th, with a standard deviation of $1660$. The large standard deviation means there is a large variability when the networks reach their minimum validation error. Seven versions reached their minimum before epoch $5000$, while the slowest reached its minimum at the $8920$th epoch. This means that if $\tm{num\_epochs} = 5000$, the performance of seven of the networks would not be influenced by the reduced number of epochs. In contrast, the precision of the three remaining versions would decrease. This highlights the need for good network initialization values when \tm{num\_epochs} is limited or for choosing a large value for \tm{num\_epochs}.

For the \tm{NN-Olsen-10000 surrogate} method, we wanted to investigate the effect of the hyperparameters. We considered the same hyperparameter grid as in \Cref{Appendix:Implementation}, i.e., $\tm{lr} \in \{0.01, 0.001, 0.0001\}$ and $\tm{width} \in \{32, 64, 128\}$ and we call the corresponding method for \tm{NN-Olsen-CV-10000 surrogate}. We fit ten versions of the \tm{NN-Olsen-CV-10000 surrogate} method with different initialized network weights. The average best epoch was the $7706$th, with a standard deviation of $2496$, while the average $\operatorname{MSE}_v$ was $1.208$, with a standard deviation of $0.0236$. This score is nearly identical to the average score we obtained for the \tm{NN-Olsen-10000 surrogate} method above with default hyperparameters: $\tm{lr} = 0.001$ and $\tm{width} = 64$. The default hyperparameters were chosen two times, but $\tm{lr} = 0.0001$ and $\tm{width} = 128$ was the best combination five times. For these five repetitions, we obtained an average $\operatorname{MSE}_v$ of $1.198$, with a standard deviation of $0.0053$. However, the average best epoch was then $9456$, with a standard deviation of $473$, which is close to the maximum number of epochs. Thus, we might see further improvements for this hyperparameter combination by increasing \tm{num\_epochs}. We ran one network with $\tm{num\_epochs} = 20\,000$, which obtained its best validation score after $15\,518$ epochs. The method's training time was 1:08:07:48.1, and it got an $\operatorname{MSE}_v$ score of $1.214$, which is at the same level as previous versions.

We repeated the investigations for the $\text{Abalone}_\text{cont}$ experiment. For the ten versions of the \tm{NN-Olsen-10000 surrogate} approach, we obtain an average $\operatorname{MSE}_v$ score of $1.178$, with a standard deviation of $0.0068$. This score is lower than the one reported in \Cref{tab:real-world-data,tab:SimStudyTimeAbaloneCont}. Thus, it is likely that that version had poorly initialized network parameters, or that the training and validation data sets were not representative. The average best epoch was the $5410$th, with a standard deviation of $1842$, and the average training time was 10:45:13.1. The best of the ten versions obtained an $\operatorname{MSE}_v$ score of $1.167$, beating all other methods. We also fitted ten versions of the \tm{NN-Olsen-CV-10000 surrogate} method. They obtained an average $\operatorname{MSE}_v$ score of $1.77$, with a standard deviation of $0.0067$. That is, there is minimal improvement in conducting cross-validation. The default hyperparameters were never the best hyperparameter combination. The $\tm{lr} = 0.0001$ and $\tm{width} = 128$ combination was the best five times, while $\tm{lr} = 0.001$ and $\tm{width} = 128$ was best four times. For the former combination, the average $\operatorname{MSE}_v$ score is $1.171$, with a standard deviation of $0.0015$. The average best number of epochs is $9253$; meaning that we should consider increasing \tm{num\_epochs}. We ran one network with $\tm{num\_epochs} = 20\,000$, which obtained its best validation score after $11\,523$ epochs. The method's training time was 1:02:02:24.5, and it got an $\operatorname{MSE}_v$ score of $1.169$, equal to the two best methods; \tm{NN-Olsen-20000 surrogate} and \tm{PPR separate}. However, the latter is approximately $700$ times faster to train.

Several methods exist to stabilize and robustify neural networks: we can regularize the network parameters, apply drop-out during training, or create an ensemble model of several networks. One can also initialize several networks and only continue to train the best-performing one after a fixed number of epochs. The latter is done in the \tm{NN-Frye-ES} and \tm{NN-Olsen-ES surrogate} methods, but we do not see a systematic improvement. In the \textsc{R} package \tm{torch} \parencite{rtorch}, the weights and biases in each layer are uniformly initialized from $\mathcal{U}(-\sqrt{N_\text{in}}, \sqrt{N_\text{in}})$, where $N_\text{in}$ is the number of inputs to the linear layers in the network. Other initialization schemes exist, such as Xavier initialization \parencite{glorot2010understanding} and Kaiming initialization \parencite{he2015delving}, where the latter considers the rectifier nonlinearities to initialize the network parameters robustly. Discovering better procedure and ensuring representative training, validation, and test data should be of focus and is mentioned as further work in the conclusion in \Cref{sec:Conclusion}.

\begin{table}[!t]
\centering
\vspace{-8ex}
\begin{adjustbox}{max width=1\textwidth, max totalheight=21.75cm}
\begin{tabular}{lrrrrrrrrrr}
\toprule
Method & Training & Generating $\x_\s^{(k)}$ & Predicting $v(\s)$ & Total CPU Time & $\operatorname{MSE}_v$ \\ 
\midrule
\rowcolor{col_ind!10} \tm{Independence} & 39.4 & 2:03.2 & 1:15.5 & 3:58.1 & 8.672\\
\rowcolor{col_ind!10} \tm{Independence}$^*$ & 0.0 & 7.6 & 1:16.6 & 1:24.2 & 8.679\\ 
\rowcolor{col_emp!10} \tm{Empirical} & 35.3 & 2:10.3 & 57.6 & 3:43.2 & 1.540 \\ 
\rowcolor{col_par!10} \tm{Gaussian} & 0.0 & 2:21.1 & 1:22.9 & 3:44.0 & 1.349 \\ 
\rowcolor{col_par!10} \tm{Copula} & 0.0 & 13:01.0 & 2:04.5 & 15:05.5 & 1.223 \\ 
\rowcolor{col_par!10} \tm{GH} & 4:20.2 & 2:53.9 & 1:25.6 & 8:39.7 & 1.292 \\ 
\rowcolor{col_par!10} \tm{Burr} & 2:56.3 & 1:00.1 & 1:25.9 & 5:22.3 & 5.640 \\ 
\rowcolor{col_gen!10} \tm{Ctree} & 9.6 & 7:19.4 & 11.8 & 7:40.8 & 1.393 \\ 
\rowcolor{col_gen!10} \tm{VAEAC-200} & 3:03.2 & 6:52.1 & 1:30.8 & 11:26.1 & 1.340 \\ 
\rowcolor{col_gen!10} \tm{VAEAC-1000} & 14:52.5 & 7:22.2 & 1:25.0 & 23:39.7 & 1.217 \\ 
\rowcolor{col_gen!10} \tm{VAEAC-10000} & 2:24:51.7 & 7:43.1 & 1:28.8 & 2:34:03.6 & 1.182 \\ 
\rowcolor{col_gen!10} \tm{VAEAC-20000} & 4:56:25.7 & 7:49.5 & 1:20.9 & 5:05:36.1 & 1.195 \\ 
\rowcolor{col_gen!10} \tm{VAEAC-40000} & 10:25:31.1 & 7:48.4 & 1:31.2 & 10:34:50.7 & 1.193 \\ 
\rowcolor{col_gen!10} \tm{VAEAC-f-indir-200} & 3:27.7 & 9:42.7 & 1:27.7 & 14:38.1 & 1.412 \\ 
\rowcolor{col_gen!10} \tm{VAEAC-f-indir-1000} & 16:23.3 & 7:59.2 & 1:27.4 & 25:49.9 & 1.255 \\ 
\rowcolor{col_gen!10} \tm{VAEAC-f-indir-10000} & 2:36:23.3 & 8:23.1 & 1:25.8 & 2:46:12.2 & 1.197 \\ 
\rowcolor{col_gen!10} \tm{VAEAC-f-indir-20000} & 5:10:17.5 & 8:17.9 & 1:19.0 & 5:19:54.4 & 1.184 \\ 
\rowcolor{col_gen!10} \tm{VAEAC-f-indir-40000} & 10:32:42.9 & 9:08.1 & 1:30.6 & 10:43:21.6 & 1.181 \\ 
\rowcolor{col_gen!10} \tm{VAEAC-f-dir-200} & 3:27.7 & 9:42.7 & 0.0 & 13:10.4 & 1.686 \\ 
\rowcolor{col_gen!10} \tm{VAEAC-f-dir-1000} & 16:23.3 & 7:59.2 & 0.0 & 24:22.5 & 1.310 \\ 
\rowcolor{col_gen!10} \tm{VAEAC-f-dir-10000} & 2:36:23.3 & 8:23.1 & 0.0 & 2:44:46.4 & 1.228 \\ 
\rowcolor{col_gen!10} \tm{VAEAC-f-dir-20000} & 5:10:17.5 & 8:17.9 & 0.0 & 5:18:35.4 & 1.231 \\ 
  \rowcolor{col_gen!10} \tm{VAEAC-f-dir-40000} & 10:32:42.9 & 9:08.1 & 0.1 & 10:41:51.1 & 1.196 \\ 
\rowcolor{col_sep!10} \tm{LM sep.} & 0.2 & --- & 0.1 & 0.3 & 1.684 \\ 
\rowcolor{col_sep!10} \tm{Poly-2 sep.} & 1.5 & --- & 0.2 & 1.7 & 1.350 \\ 
\rowcolor{col_sep!10} \tm{Poly-3 sep.} & 1.4 & --- & 0.2 & 1.6 & 1.320 \\ 
\rowcolor{col_sep!10} \tm{LM-inter-2 sep.} & 0.3 & --- & 0.1 & 0.4 & 1.423 \\ 
\rowcolor{col_sep!10} \tm{LM-inter-3 sep.} & 0.4 & --- & 0.1 & 0.5 & 1.389 \\ 
\rowcolor{col_sep!10} \tm{Poly-inter-2 sep.} & 1.3 & --- & 0.4 & 1.7 & 1.320 \\ 
\rowcolor{col_sep!10} \tm{Poly-inter-3 sep.} & 2.4 & --- & 0.4 & 2.8 & 1.394 \\ 
\rowcolor{col_sep!10} \tm{Lasso sep.} & 8.1 & --- & 0.2 & 8.3 & 1.696 \\ 
\rowcolor{col_sep!10} \tm{Ridge sep.} & 10.2 & --- & 0.1 & 10.3 & 2.027 \\ 
\rowcolor{col_sep!10} \tm{Elastic sep.} & 9.2 & --- & 0.1 & 9.3 & 1.706 \\ 
\rowcolor{col_sep!10} \tm{GAM sep.} & 27.4 & --- & 6.3 & 33.7 & 1.298 \\ 
\rowcolor{col_sep!10} \tm{GAM-5 sep.} & 11.4 & --- & 1.0 & 12.4 & 1.306 \\ 
\rowcolor{col_sep!10} \tm{GAM-10 sep.} & 19.9 & --- & 1.0 & 20.9 & 1.294 \\ 
\rowcolor{col_sep!10} \tm{GAM-CV, sep.} & 5:55.3 & --- & 1.1 & 5:56.4 & 1.303 \\ 
\rowcolor{col_sep!10} \tm{PCR sep.} & 4.8 & --- & 0.1 & 4.9 & 1.719 \\ 
\rowcolor{col_sep!10} \tm{PLS sep.} & 3.2 & --- & 0.1 & 3.3 & 1.717 \\ 
\rowcolor{col_sep!10} \tm{PCR sep.} & 4.6 & --- & 0.1 & 4.7  & 1.721 \\ 
\rowcolor{col_sep!10} \tm{PPR sep.} & 2:14.6 & --- & 0.5 & 2:15.1 & \B1.169 \\ 
\rowcolor{col_sep!10} \tm{PPR-fixed sep.} & 7.1 & --- & 0.4 & 7.5 & 1.270 \\ 
\rowcolor{col_sep!10} \tm{SVM sep.} & 47.5 & --- & 3.7 & 51.2 & 1.260 \\
\rowcolor{col_sep!10} \tm{KNN sep.} & 30.2 & --- & 4.0 & 34.2 & 1.330 \\ 
\rowcolor{col_sep!10} \tm{Tree sep.} & 3.4 & --- & 0.2 & 3.6 & 1.553 \\ 
\rowcolor{col_sep!10} \tm{RF sep.} & 1:09:06.1 & --- & 9.7 & 1:09:15.8 & 1.239 \\ 
\rowcolor{col_sep!10} \tm{RF-def sep.} & 39.7 & --- & 4.5 & 44.2 & 1.312 \\ 
\rowcolor{col_sep!10} \tm{CatBoost sep.} & 6:16.9 & --- & 0.2 & 6:17.1 & 1.190 \\ 
\rowcolor{col_sur!10} \tm{LM sur.} & 3.2 & --- & 0.5 & 3.7 & 2.912 \\ 
\rowcolor{col_sur!10} \tm{Poly-2 sur.} & 4.3 & --- & 0.8 & 5.1 & 2.664 \\ 
\rowcolor{col_sur!10} \tm{Poly-3 sur.} & 4.1 & --- & 0.9 & 5.0 & 2.628 \\ 
\rowcolor{col_sur!10} \tm{LM-inter-2 sur.} & 7.2 & --- & 0.9 & 8.1 & 1.775 \\ 
\rowcolor{col_sur!10} \tm{Poly-inter-2 sur.} & 5.9 & --- & 1.2 & 7.1 & 2.447 \\ 
\rowcolor{col_sur!10} \tm{Poly-inter-3 sur.} & 15.2 & --- & 1.7 & 16.9 & 1.930 \\ 
\rowcolor{col_sur!10} \tm{Lasso sur.} & 7.2 & --- & 0.2 & 7.4 & 2.912 \\ 
\rowcolor{col_sur!10} \tm{Ridge sur.} & 7.4 & --- & 0.1 & 7.5 & 2.998 \\ 
\rowcolor{col_sur!10} \tm{Elastic sur.} & 7.4 & --- & 0.2 & 7.6 & 2.912 \\ 
\rowcolor{col_sur!10} \tm{GAM sur}. & 28.5 & --- & 12.6 & 41.1 & 2.611 \\ 
\rowcolor{col_sur!10} \tm{GAM-5 sur.} & 18.3 & --- & 1.0 & 19.3 & 2.612 \\ 
\rowcolor{col_sur!10} \tm{GAM-10 sur.} & 1:01.4 & --- & 0.7 & 1:02.1 & 2.605 \\ 
\rowcolor{col_sur!10} \tm{PCR sur.} & 17.8 & --- & 0.6 & 18.4 & 2.912 \\ 
\rowcolor{col_sur!10} \tm{PLS sur.} & 17.2 & --- & 0.7 & 17.9 & 2.912 \\ 
\rowcolor{col_sur!10} \tm{PPR sur.} & 14:56.2 & --- & 1.3 & 14:57.5 & 1.548 \\ 
\rowcolor{col_sur!10} \tm{KNN sur.} & 0.2 & --- & 0.0 & 0.2 & 13.081 \\ 
\rowcolor{col_sur!10} \tm{Tree sur.} & 11.2 & --- & 0.6 & 11.8 & 2.839 \\ 
\rowcolor{col_sur!10} \tm{RF sur.} &  1:14:15.8 & --- & 15.0 & 1:14:30.8 & 1.281 \\ 
\rowcolor{col_sur!10} \tm{RF-def sur.} & 1:45.1 & --- & 6.0 & 1:51.1 & 1.448 \\ 
\rowcolor{col_sur!10} \tm{XGBoost sur.} & 2:46:14.9 & --- & 0.6 & 2:46:15.5 & 1.536 \\ 
\rowcolor{col_sur!10} \tm{XGBoost-def sur.} & 1:05.5 & --- & 0.6 & 1:06.1 & 1.437 \\
\rowcolor{col_sur!10} \tm{CatBoost sur.} & 9:09.2 & --- & 1.6 & 9:10.8 & 1.298 \\ 
\rowcolor{col_sur!10} \tm{NN-Frye-3000 sur.} & 5:47:58.1 & --- & 1.9 & 5:48:00.0 & 1.625 \\ 
\rowcolor{col_sur!10} \tm{NN-Frye-6000 sur.} & 11:34:15.4 & --- & 1.7 & 11:34:17.1 & 1.433 \\ 
\rowcolor{col_sur!10} \tm{NN-Frye-15000 sur.} & 1:03:05:38.7 & --- & 2.3 & 1:03:05:41.0 & 1.310 \\ 
\rowcolor{col_sur!10} \tm{NN-Frye-40000 sur.} & 3:01:48:38.7 & --- & 2.1 & 3:01:48:40.8 & 1.244 \\ 
\rowcolor{col_sur!10} \tm{NN-Frye-ES sur.} & 6:10:13.2 & --- & 2.1 & 6:10:15.3 & 1.374 \\ 
\rowcolor{col_sur!10} \tm{NN-Olsen-500 sur.}  & 2:24:15.7 & --- & 1.7 & 2:24:17.4 & 1.248 \\ 
\rowcolor{col_sur!10} \tm{NN-Olsen-2500 sur.} & 12:10:18.6 & --- & 1.4 & 12:10:20.0 & 1.201 \\ 
\rowcolor{col_sur!10} \tm{NN-Olsen-10000 sur.} & 2:00:40:20.3 & --- & 2.0 & 2:00:40:22.3 & 1.194 \\ 
\rowcolor{col_sur!10} \tm{NN-Olsen-20000 sur.} & 3:22:23:46.0 & --- & 2.0 & 3:22:23:48.0 & \B1.169 \\ 
\rowcolor{col_sur!10} \tm{NN-Olsen-ES sur.} & 8:28:19.1 & --- & 1.8 & 8:28:20.9 & 1.191 \\ 
\bottomrule
\end{tabular}
\end{adjustbox}
\caption{{\small The $\text{Abalone}_\text{cont}$ data set experiment: $M = 7$, $N_\text{train} = 3133$, and $N_\text{test} = 1044$. 
}}
\label{tab:SimStudyTimeAbaloneCont}
\end{table}

\begin{table}[ht]
\centering
\vspace{-8ex}
\begin{adjustbox}{max width=1\textwidth, max totalheight=21.75cm}
\begin{tabular}{lrrrrrrrrrr}
\toprule
Method & Training & Generating $\x_\s^{(k)}$ & Predicting $v(\s)$ & Total CPU Time & $\operatorname{MSE}_v$ \\ 
\midrule
\rowcolor{col_ind!10} \tm{Independence}$^*$ & 0.0 & 18.6 & 3:32.9 & 3:51.5 & 9.144 \\ 
\rowcolor{col_gen!10} \tm{Ctree} & 18.1 & 18:26.6 & 29.4 & 19:14.1 & 1.424 \\ 
\rowcolor{col_gen!10} \tm{VAEAC-200} & 4:13.4 & 15:47.1 & 5:16.6 & 25:17.1 & 1.467 \\ 
\rowcolor{col_gen!10} \tm{VAEAC-1000} & 16:01.1 & 18:03.6 & 4:25.6 & 38:30.3 & 1.230 \\ 
\rowcolor{col_gen!10} \tm{VAEAC-10000} & 2:41:54.8 & 15:58.2 & 5:31.9 & 3:03:24.9 & 1.194 \\ 
\rowcolor{col_gen!10} \tm{VAEAC-20000} & 5:32:11.4 & 16:46.2 & 5:26.1 & 5:54:23.7 & 1.193 \\ 
\rowcolor{col_gen!10} \tm{VAEAC-40000} & 11:19:30.9 & 17:24.9 & 11:26.3 & 11:48:22.1 & \B1.180 \\ 
\rowcolor{col_gen!10} \tm{VAEAC-f-indir-200} & 4:16.5 & 17:02.6 & 5:21.7 & 26:40.8 & 1.457 \\ 
\rowcolor{col_gen!10} \tm{VAEAC-f-indir-1000} & 18:42.3 & 18:40.6 & 3:57.5 & 41:20.4 & 1.270 \\ 
\rowcolor{col_gen!10} \tm{VAEAC-f-indir-10000} & 2:48:12.4 & 17:13.9 & 5:33.1 & 3:10:59.4 & 1.234 \\ 
\rowcolor{col_gen!10} \tm{VAEAC-f-indir-20000} & 5:44:35.2 & 17:49.1 & 5:27.1 & 6:07:51.4 & 1.216 \\ 
  \rowcolor{col_gen!10} \tm{VAEAC-f-indir-40000} & 12:51:32.9 & 17:42.9 & 7:38.7 & 13:16:54.5 & 1.220 \\ 
\rowcolor{col_gen!10} \tm{VAEAC-f-dir-200} & 4:16.5 & 17:02.6 & 0.0 & 21:19.1 & 1.761 \\ 
\rowcolor{col_gen!10} \tm{VAEAC-f-dir-1000} & 18:42.3 & 18:40.6 & 0.0 & 37:22.9 & 1.322 \\ 
\rowcolor{col_gen!10} \tm{VAEAC-f-dir-10000} & 2:48:12.4 & 17:13.9 & 0.0 & 3:05:26.3 & 1.268 \\ 
\rowcolor{col_gen!10} \tm{VAEAC-f-dir-20000} & 5:44:35.2 & 17:49.1 & 0.0 & 6:02:24.3 & 1.239 \\ 
 \rowcolor{col_gen!10} \tm{VAEAC-f-dir-40000} & 12:51:32.9 & 17:42.9 & 0.3 & 13:09:16.1 & 1.264 \\ 
\rowcolor{col_sep!10} \tm{LM sep.} & 0.6 & --- & 0.3 & 0.9 & 1.581 \\ 
\rowcolor{col_sep!10} \tm{Poly-2 sep.} & 2.7 & --- & 0.4 & 3.1 & 1.338 \\ 
\rowcolor{col_sep!10} \tm{Poly-3 sep.} & 3.2 & --- & 0.5 & 3.7 & 1.314 \\ 
\rowcolor{col_sep!10} \tm{Poly-4 sep.} & 3.7 & --- & 0.8 & 4.5 & 1.306 \\ 
\rowcolor{col_sep!10} \tm{LM-inter-2 sep.} & 0.9 & --- & 0.3 & 1.2 & 1.381 \\ 
\rowcolor{col_sep!10} \tm{LM-inter-3 sep.} & 1.4 & --- & 0.3 & 1.7 & 1.357 \\ 
\rowcolor{col_sep!10} \tm{LM-inter-4 sep.} & 2.1 & --- & 0.4 & 2.5 & 1.360 \\ 
\rowcolor{col_sep!10} \tm{Poly-inter-2 sep.} & 2.6 & --- & 0.8 & 3.4 & 1.310 \\ 
\rowcolor{col_sep!10} \tm{Poly-inter-3 sep.} & 4.5 & --- & 0.1 & 4.6 & 1.512 \\ 
\rowcolor{col_sep!10} \tm{Poly-inter-4 sep.} & 10.4 & --- & 1.4 & 11.8 & 9.314 \\
\rowcolor{col_sep!10} \tm{GAM sep.} & 56.3 & --- & 13.4 & 1:09.7 & 1.299 \\ 
\rowcolor{col_sep!10} \tm{GAM-5 sep.} & 22.7 & --- & 2.1 & 24.8 & 1.302 \\ 
\rowcolor{col_sep!10} \tm{GAM-10 sep.} & 36.2 & --- & 2.2 & 38.4 & 1.297 \\ 
\rowcolor{col_sep!10} \tm{GAM-CV, sep.} & 12:10.8 & --- & 2.1 & 12:12.9 & 1.299 \\ 
\rowcolor{col_sep!10} \tm{PCR sep.} & 11.6 & ---  & 0.4 & 12.0 & 1.788 \\ 
\rowcolor{col_sep!10} \tm{PLS sep.} & 7.1 & --- & 0.3 & 7.4 & 1.630 \\ 
\rowcolor{col_sep!10} \tm{PPR sep.} & 3:33.5 & ---  & 1.1 & 3:34.6 &  1.185 \\ 
\rowcolor{col_sep!10} \tm{PPR-fixed sep.} & 15.4 & ---  & 0.8 & 16.2 & 1.198 \\ 
\rowcolor{col_sep!10} \tm{KNN sep.} & 1:02.8 & --- & 8.7 & 1:11.5 & 1.366 \\ 
\rowcolor{col_sep!10} \tm{Tree sep.} & 6.9 & --- & 0.3 & 7.2 & 1.559 \\ 
\rowcolor{col_sep!10} \tm{RF sep.} & 2:30:50.4 & ---  & 19.3 & 2:31:09.7 & 1.259 \\ 
\rowcolor{col_sep!10} \tm{RF-def sep.} & 1:21.4 & ---  & 8.5 & 1:29.9 & 1.344 \\ 
\rowcolor{col_sep!10} \tm{CatBoost sep.} & 18:24.7 & ---  & 0.3 & 18:25.0 & 1.213 \\ 
\rowcolor{col_sur!10} \tm{LM sur.} & 5.9 & ---  & 1.3 & 7.2 & 2.770 \\ 
\rowcolor{col_sur!10} \tm{Poly-2 sur.} & 7.8 & --- & 1.5 & 9.3 & 2.625 \\ 
\rowcolor{col_sur!10} \tm{Poly-3 sur.} & 7.7 & --- & 1.5 & 9.2 & 2.577 \\ 
\rowcolor{col_sur!10} \tm{LM-inter-2 sur.} & 27.4 & --- & 1.9 & 29.3 & 1.705 \\ 
\rowcolor{col_sur!10} \tm{LM-inter-3 sur.} & 9:11.8 & --- & 5.2 & 9:17.0 & 1.443 \\ 
\rowcolor{col_sur!10} \tm{Poly-inter-2 sur.} & 9.1 & --- & 1.7 & 10.8 & 2.435 \\ 
\rowcolor{col_sur!10} \tm{Poly-inter-3 sur.} & 29.5 & --- & 3.3 & 32.8 & 1.929 \\ 
\rowcolor{col_sur!10} \tm{GAM sur.} & 58.4 & --- & 26.2 & 1:24.6 & 2.557 \\ 
\rowcolor{col_sur!10} \tm{GAM-5 sur.} & 30.3 & --- & 2.1 & 32.4 & 2.556 \\ 
\rowcolor{col_sur!10} \tm{GAM-10 sur.} & 45.8 & --- & 2.0 & 47.8 & 2.553 \\
\rowcolor{col_sur!10} \tm{PCR sur.} & 46.2 & ---  & 0.6 & 46.8 & 2.818 \\ 
\rowcolor{col_sur!10} \tm{PLS sur.} & 38.4 & --- & 1.3 & 39.7 & 2.770 \\ 
\rowcolor{col_sur!10} \tm{PPR sur.} & 55:28.4 & ---  & 3.2 & 55:31.6 & 1.538 \\ 
\rowcolor{col_sur!10} \tm{KNN sur.} & 0.5 & --- & 0.0 & 0.5 & 9.305 \\ 
\rowcolor{col_sur!10} \tm{Tree sur.} & 25.7 & --- & 0.6 & 26.3 & 2.914 \\ 
\rowcolor{col_sur!10} \tm{RF sur.} & 3:45:58.8 & ---  & 35.3 & 3:46:34.1 & 1.311 \\ 
\rowcolor{col_sur!10} \tm{RF-def sur.} & 4:25.1 & ---  & 12.0 & 4:37.1 & 1.473 \\ 
\rowcolor{col_sur!10} \tm{CatBoost sur.} & 29:26.0 & ---  & 2.6 & 29:28.6 & 1.348 \\ 
\rowcolor{col_sur!10} \tm{NN-Frye-3000 sur.} & 6:08:53.8 & ---  & 4.1 & 6:08:57.9 & 2.049 \\ 
\rowcolor{col_sur!10} \tm{NN-Frye-6000 sur.} & 12:06:33.8 & ---  & 4.1 & 12:06:37.9 & 1.742 \\ 
\rowcolor{col_sur!10} \tm{NN-Frye-15000 sur.} & 1:09:15:16.7 & ---  & 3.9 & 1:09:15:20.6 & 1.445 \\ 
\rowcolor{col_sur!10} \tm{NN-Frye-40000 sur.} & 3:16:01:07.5 & ---  & 4.0 & 3:16:01:11.5 & 1.320 \\ 
\rowcolor{col_sur!10} \tm{NN-Frye-ES sur.} & 4:53:28.7 & --- & 6.4 & 4:53:35.1 & 1.973 \\ 
\rowcolor{col_sur!10} \tm{NN-Olsen-500 sur.} & 2:21:43.4 & ---  & 3.2 & 2:21:46.6 & 1.282 \\ 
\rowcolor{col_sur!10} \tm{NN-Olsen-2500 sur.} & 12:02:38.3 & ---  & 3.5 & 12:02:41.8 & 1.216 \\ 
\rowcolor{col_sur!10} \tm{NN-Olsen-10000 sur.} & 2:01:07:27.4 & ---  & 2.9 & 2:01:07:30.3 & 1.192 \\ 
\rowcolor{col_sur!10} \tm{NN-Olsen-20000 sur.} & 4:04:25:26.1 & --- & 2.9 & 4:04:25:29.0 & 1.210 \\ 
\rowcolor{col_sur!10} \tm{NN-Olsen-ES sur.} & 2:52:53.4 & --- & 3.8 & 2:52:57.2 & 1.269 \\ 
\bottomrule
\end{tabular}
\end{adjustbox}
\caption{{\small The $\text{Abalone}_\text{all}$ data set experiment: $M = 8$, $N_\text{train} = 3133$, and $N_\text{test} = 1044$.
}}
\label{tab:SimStudyTimeAbaloneAll}
\end{table}

\begin{table}[ht]
\centering
\vspace{-8ex}
\begin{adjustbox}{max width=1\textwidth, max totalheight=21.75cm}
\begin{tabular}{lrrrrrr}
\toprule
Method & Training & Generating $\x_\s^{(k)}$ & Predicting $v(\s)$ & Total CPU Time & $\operatorname{MSE}_v$\\ 
\midrule
\rowcolor{col_ind!10} \tm{Independence} & 2.6 & 23.3 & 26.4 & 52.3 & 0.203 \\ 
\rowcolor{col_ind!10} \tm{Independence}$^*$ & 0.0 & 7.8 & 30.6 & 38.4 & 0.196 \\ 
\rowcolor{col_emp!10} \tm{Empirical} & 2.6 & 11.1 & 1.4 & 15.1 & 0.143 \\ 
\rowcolor{col_par!10} \tm{Gaussian} & 0.0 & 2:06.1 & 29.0 & 2:35.1 & 0.127 \\ 
\rowcolor{col_par!10} \tm{Copula} & 0.0 & 10:17.2 & 37.3 & 10:54.5 & 0.127 \\ 
\rowcolor{col_par!10} \tm{GH} & 4:34.8 & 2:29.6 & 26.9 & 7:31.3 & 0.133 \\ 
\rowcolor{col_gen!10} \tm{Ctree} & 9.6 & 6:35.7 & 1.6 & 6:46.9 & 0.158 \\ 
\rowcolor{col_gen!10} \tm{VAEAC-200} & 48.5 & 7:44.9 & 2:02.7 & 10:36.1 & 0.134 \\ 
\rowcolor{col_gen!10} \tm{VAEAC-1000} & 2:35.1 & 7:04.5 & 1:41.1 & 11:20.7 & 0.131 \\ 
\rowcolor{col_gen!10} \tm{VAEAC-5000} & 12:45.5 & 7:28.0 & 1:44.3 & 21:57.8 & 0.128 \\ 
\rowcolor{col_gen!10} \tm{VAEAC-10000} & 21:37.3 & 8:38.8 & 1:23.8 & 31:39.9 & 0.128 \\ 
\rowcolor{col_gen!10} \tm{VAEAC-20000} & 43:17.2 & 8:36.1 & 1:17.2 & 53:10.5 & 0.129 \\ 
\rowcolor{col_gen!10} \tm{VAEAC-f-indir-200} & 41.0 & 7:55.9 & 1:41.8 & 10:18.7 & 0.137 \\ 
\rowcolor{col_gen!10} \tm{VAEAC-f-indir-1000} & 2:41.6 & 7:33.1 & 1:44.7 & 11:59.4 & 0.137 \\ 
\rowcolor{col_gen!10} \tm{VAEAC-f-indir-5000} & 13:33.1 & 7:53.8 & 1:44.1 & 23:11.0 & 0.133 \\ 
\rowcolor{col_gen!10} \tm{VAEAC-f-indir-10000} & 22:53.1 & 9:17.1 & 1:29.4 & 33:39.6 & 0.134 \\ 
\rowcolor{col_gen!10} \tm{VAEAC-f-indir-20000} & 44:32.1 & 8:58.6 & 1:09.5 & 54:40.2 & 0.130 \\ 
\rowcolor{col_gen!10} \tm{VAEAC-f-dir-200} & 41.0 & 7:55.9 & 0.0 & 8:36.9 & 0.149 \\ 
\rowcolor{col_gen!10} \tm{VAEAC-f-dir-1000} & 2:41.6 & 7:33.1 & 0.0 & 10:14.7 & 0.145 \\ 
\rowcolor{col_gen!10} \tm{VAEAC-f-dir-5000} & 13:33.1 & 7:53.8 & 0.0 & 21:26.9 & 0.137 \\ 
\rowcolor{col_gen!10} \tm{VAEAC-f-dir-10000} & 22:53.1 & 9:17.1 & 0.0 & 32:10.2 & 0.143 \\ 
\rowcolor{col_gen!10} \tm{VAEAC-f-dir-20000} & 44:32.1 & 8:58.6 & 0.0 & 53:30.7 & 0.133 \\ 
\rowcolor{col_sep!10} \tm{LM sep.} & 1.3 & --- & 0.6 & 1.9 & \B0.126 \\ 
\rowcolor{col_sep!10} \tm{LM-inter-2 sep.} & 1.5 & --- & 0.6 & 2.1 & 0.127 \\ 
\rowcolor{col_sep!10} \tm{LM-inter-3 sep.} & 1.9 & --- & 0.8 & 2.7 & 0.134 \\ 
\rowcolor{col_sep!10} \tm{LM-inter-4 sep.} & 2.5 & --- & 0.6 & 3.1 & 0.157 \\ 
\rowcolor{col_sep!10} \tm{Lasso sep.} & 44.4 & --- & 0.6 & 45.0 & \B0.126 \\ 
\rowcolor{col_sep!10} \tm{Ridge sep.} & 49.1 & --- & 0.5 & 49.6 & 0.128 \\ 
\rowcolor{col_sep!10} \tm{Elastic sep.} & 45.1 & --- & 0.6 & 45.7 & \B0.126 \\ 
\rowcolor{col_sep!10} \tm{GAM sep.} & 59.2 & --- & 4.4 & 1:03.6 & \B0.126 \\ 
\rowcolor{col_sep!10} \tm{PCR sep.} & 8.1 & --- & 0.8 & 8.9 & \B0.126 \\ 
\rowcolor{col_sep!10} \tm{PLS sep.} & 7.0 & --- & 0.7 & 7.7 & \B0.126 \\ 
\rowcolor{col_sep!10} \tm{PPR sep.} & 5:21.9 & --- & 0.5 & 5:22.4 & \B0.126 \\ 
\rowcolor{col_sep!10} \tm{PPR-fixed sep.} & 8.2 & --- & 0.1 & 8.3 & 0.145 \\ 
\rowcolor{col_sep!10} \tm{SVM sep.} & 8.6 & --- & 1.1 & 9.7 & 0.139 \\ 
\rowcolor{col_sep!10} \tm{KNN sep.} & 17.3 & --- & 3.0 & 20.3 & 0.150 \\ 
\rowcolor{col_sep!10} \tm{Tree sep.} & 5.9 & --- & 0.9 & 6.8 & 0.189 \\ 
\rowcolor{col_sep!10} \tm{RF sep.} & 1:00:14.2 & --- & 9.4 & 1:00:23.6 & 0.143 \\ 
\rowcolor{col_sep!10} \tm{RF-def sep.} & 28.8 & --- & 5.8 & 34.6 & 0.155 \\ 
\rowcolor{col_sep!10} \tm{XGBoost sep.} & 1:10:10.8 & --- & 0.9 & 1:10:11.7 & 0.137 \\ 
\rowcolor{col_sep!10} \tm{XGBoost-def sep.} & 1:01.9 & --- & 1.1 & 1:03.0 & 0.182 \\ 
\rowcolor{col_sep!10} \tm{CatBoost sep.} & 18:40.3 & --- & 0.3 & 18:40.6 & 0.135 \\ 
\rowcolor{col_sur!10} \tm{LM sur.} & 3.6 & --- & 0.8 & 4.4 & 0.165 \\ 
\rowcolor{col_sur!10} \tm{LM-inter-2 sur.} & 19.3 & --- & 1.1 & 20.4 & 0.134 \\ 
\rowcolor{col_sur!10} \tm{LM-inter-3 sur.} & 11:51.0 &  & 2.7 & 11:53.7 & 0.128 \\
\rowcolor{col_sur!10} \tm{Lasso sur.} & 6.9 & --- & 0.2 & 7.1 & 0.165 \\ 
\rowcolor{col_sur!10} \tm{Ridge sur.} & 7.9 & --- & 0.2 & 8.1 & 0.165 \\ 
\rowcolor{col_sur!10} \tm{Elastic sur.} & 5.6 & --- & 0.4 & 6.0 & 0.165 \\ 
\rowcolor{col_sur!10} \tm{GAM sur.} & 18.7 & --- & 2.6 & 21.3 & 0.168 \\ 
\rowcolor{col_sur!10} \tm{PCR sur.} & 22.7 & --- & 0.7 & 23.4 & 0.165 \\ 
\rowcolor{col_sur!10} \tm{PLS sur.} & 20.8 & --- & 0.6 & 21.4 & 0.165 \\ 
\rowcolor{col_sur!10} \tm{PPR sur.} & 4:11.9 & --- & 1.1 & 4:13.0 & 0.136 \\ 
\rowcolor{col_sur!10} \tm{KNN sur.} & 7.6 & --- & 0.1 & 7.7 & 0.671 \\ 
\rowcolor{col_sur!10} \tm{Tree sur.} & 15.3 & --- & 0.8 & 16.1 & 0.254 \\  
\rowcolor{col_sur!10} \tm{RF sur.} & 1:33:37.6 & --- & 13.7 & 1:33:51.3 & 0.143 \\ 
\rowcolor{col_sur!10} \tm{RF-def sur.} & 2:28.8 & --- & 7.6 & 2:36.4 & 0.159 \\ 
\rowcolor{col_sur!10} \tm{XGBoost sur.} & 3:30:18.6 & --- & 0.5 & 3:30:19.1 & 0.167 \\ 
\rowcolor{col_sur!10} \tm{XGBoost-def sur.} & 1:11.1 & --- & 0.6 & 1:11.7 & 0.202 \\ 
\rowcolor{col_sur!10} \tm{CatBoost sur.} & 52.7 & --- & 0.7 & 53.4 & 0.140 \\ 
\rowcolor{col_sur!10} \tm{NN-Frye-3000 sur.} & 55:43.4 & --- & 2.0 & 55:45.4 & 0.177 \\ 
\rowcolor{col_sur!10} \tm{NN-Frye-6000 sur.} & 1:52:42.1 & --- & 0.8 & 1:52:42.9 & 0.178 \\ 
\rowcolor{col_sur!10} \tm{NN-Frye-10000 sur.} & 3:11:37.4 & --- & 2.0 & 3:11:39.4 & 0.154 \\ 
\rowcolor{col_sur!10} \tm{NN-Frye-15000 sur.} & 4:28:14.0 & --- & 1.9 & 4:28:15.9 & 0.153 \\ 
\rowcolor{col_sur!10} \tm{NN-Frye-ES sur.} & 35:12.8 & --- & 1.6 & 35:14.4 & 0.191 \\ 
\rowcolor{col_sur!10} \tm{NN-Olsen-500 sur.} & 17:39.8 & --- & 1.2 & 17:41.0 & 0.136 \\ 
\rowcolor{col_sur!10} \tm{NN-Olsen-2500 sur.} & 1:28:53.3 & --- & 1.5 & 1:28:54.8 & 0.135 \\ 
\rowcolor{col_sur!10} \tm{NN-Olsen-7500 sur.} & 4:27:26.2 & --- & 1.5 & 4:27:27.7 & 0.139 \\ 
\rowcolor{col_sur!10} \tm{NN-Olsen-10000 sur.} & 5:42:27.4 & --- & 1.5 & 5:42:28.9 & 0.134 \\ 
\rowcolor{col_sur!10} \tm{NN-Olsen-ES sur.} & 27:24.8 & --- & 1.7 & 27:26.5 & 0.137 \\ 
\bottomrule
\end{tabular}
\end{adjustbox}
\caption{{\small The Diabetes data set experiment: $M = 10$, $N_\text{train} = 332$, and $N_\text{test} = 110$. 
}}
\label{tab:SimStudyTimeDiabetes}
\end{table}

\begin{table}[ht]
\centering
\vspace{-8ex}
\begin{adjustbox}{max width=1\textwidth, max totalheight=21.75cm}
\begin{tabular}{lrrrrrr}
\toprule
Method & Training & Generating $\x_\s^{(k)}$ & Predicting $v(\s)$ & Total CPU Time & $\operatorname{MSE}_v$\\ 
\midrule
\rowcolor{col_ind!10} \tm{Independence} & 1:50.6 & 4:49.7 & 3:50:15.6 & 3:56:55.9 & 0.145 \\ 
\rowcolor{col_ind!10} \tm{Independence}$^*$ & 0.0 & 1:39.1 & 3:59:14.4 & 4:00:53.5 & 0.145 \\ 
\rowcolor{col_emp!10} \tm{Empirical} & 1:54.3 & 4:22.9 & 2:27:01.3 & 2:33:18.5 & 0.088 \\ 
\rowcolor{col_par!10} \tm{Gaussian} & 0.0 & 11:06.4 & 3:57:22.2 & 4:08:28.6 & 0.118 \\ 
\rowcolor{col_par!10} \tm{Copula} & 0.0 & 54:01.1 & 3:59:35.5 & 4:53:36.6 & 0.107 \\ 
\rowcolor{col_par!10} \tm{GH} & 10:46.1 & 12:46.2 & 4:00:07.6 & 4:23:39.9 & 0.109 \\ 
\rowcolor{col_par!10} \tm{Burr} & 2:22.1 & 5:19.1 & 3:44:51.5 & 3:52:32.7 & 0.202 \\ 
\rowcolor{col_gen!10} \tm{Ctree} & 1:06.3 & 33:35.5 & 27:59.7 & 1:02:41.5 & 0.102 \\ 
\rowcolor{col_gen!10} \tm{VAEAC-200} & 1:51.6 & 36:22.8 & 3:59:54.5 & 4:38:08.9 & 0.103 \\ 
\rowcolor{col_gen!10} \tm{VAEAC-1000} & 8:33.3 & 36:38.4 & 3:53:15.3 & 4:38:27.0 & 0.097 \\ 
\rowcolor{col_gen!10} \tm{VAEAC-10000} & 1:34:59.6 & 40:42.5 & 3:53:55.8 & 6:09:37.9 & 0.093 \\ 
\rowcolor{col_gen!10} \tm{VAEAC-20000} & 3:08:28.3 & 37:06.7 & 3:53:31.3 & 7:37:06.3 & 0.093 \\ 
\rowcolor{col_gen!10} \tm{VAEAC-f-indir-200} & 2:10.4 & 40:23.6 & 3:59:21.4 & 4:41:55.4 & 0.105 \\ 
\rowcolor{col_gen!10} \tm{VAEAC-f-indir-1000} & 9:55.4 & 40:43.6 & 3:49:50.5 & 4:40:29.5 & 0.096 \\ 
\rowcolor{col_gen!10} \tm{VAEAC-f-indir-10000} & 1:33:53.4 & 43:21.2 & 4:00:27.9 & 6:17:42.5 & 0.098 \\ 
\rowcolor{col_gen!10} \tm{VAEAC-f-indir-20000} & 3:17:31.3 & 43:09.3 & 4:00:22.9 & 8:01:03.5 & 0.097 \\ 
\rowcolor{col_gen!10} \tm{VAEAC-f-dir-200} & 2:10.4 & 40:23.6 & 0.0 & 42:34.0 & 0.139 \\ 
\rowcolor{col_gen!10} \tm{VAEAC-f-dir-1000} & 9:55.4 & 40:43.6 & 0.0 & 50:39.0 & 0.120 \\ 
\rowcolor{col_gen!10} \tm{VAEAC-f-dir-10000} & 1:33:53.4 & 43:21.2 & 0.0 & 2:17:14.6 & 0.122 \\ 
\rowcolor{col_gen!10} \tm{VAEAC-f-dir-20000} & 3:17:31.3 & 43:09.3 & 0.5 & 4:00:41.1 & 0.123 \\ 
\rowcolor{col_sep!10} \tm{LM sep.} & 3.4 & --- & 1.2 & 4.6 & 0.146 \\ 
\rowcolor{col_sep!10} \tm{Poly-2 sep.} & 11.6 & --- & 2.2 & 13.8 & 0.137 \\ 
\rowcolor{col_sep!10} \tm{Poly-3 sep.} & 18.2 & --- & 2.9 & 21.1 & 0.128 \\ 
\rowcolor{col_sep!10} \tm{Poly-4 sep.} & 25.1 & --- & 2.8 & 27.9 & 0.127 \\ 
\rowcolor{col_sep!10} \tm{LM-inter-2 sep.} & 6.4 & --- & 2.1 & 8.5 & 0.134 \\ 
\rowcolor{col_sep!10} \tm{LM-inter-3 sep.} & 10.1 & --- & 1.9 & 12.0 & 0.138 \\ 
\rowcolor{col_sep!10} \tm{LM-inter-4 sep.} & 22.7 & --- & 2.1 & 24.8 & 0.150 \\ 
\rowcolor{col_sep!10} \tm{Poly-inter-2 sep.} & 17.4 & --- & 6.1 & 23.5 & 0.128 \\ 
\rowcolor{col_sep!10} \tm{Poly-inter-3 sep.} & 1:03.2 & --- & 15.7 & 1:18.9 & 0.131 \\ 
\rowcolor{col_sep!10} \tm{Poly-inter-4 sep.} & 4:43.3 & --- & 1:21.1 & 6:04.4 & 1.142 \\ 
\rowcolor{col_sep!10} \tm{Lasso sep.} & 2:00.1 & --- & 1.3 & 2:01.4 & 0.146 \\ 
\rowcolor{col_sep!10} \tm{Ridge sep.} & 2:44.1 & --- & 1.5 & 2:45.6 & 0.147 \\ 
\rowcolor{col_sep!10} \tm{Elastic sep.} & 2:33.0 & --- & 1.7 & 2:34.7 & 0.146 \\ 
\rowcolor{col_sep!10} \tm{GAM sep.} & 3:12.0 & --- & 12.6 & 3:24.6 & 0.124 \\ 
\rowcolor{col_sep!10} \tm{GAM-5 sep.} & 40.8 & --- & 19.6 & 1:00.4 & 0.124 \\ 
\rowcolor{col_sep!10} \tm{GAM-10 sep.} & 41.2 & --- & 17.9 & 59.1 & 0.122 \\ 
\rowcolor{col_sep!10} \tm{GAM-CV, sep.} & 31:29.5 & --- & 16.9 & 31:46.4 & 0.124 \\ 
\rowcolor{col_sep!10} \tm{PCR sep.} & 1:04.2 & --- & 3.2 & 1:07.4 & 0.146 \\ 
\rowcolor{col_sep!10} \tm{PLS sep.} & 53.3 & --- & 2.5 & 55.8 & 0.146 \\ 
\rowcolor{col_sep!10} \tm{PPR sep.} & 25:17.3 & --- & 2.1 & 25:19.4 & 0.129 \\ 
\rowcolor{col_sep!10} \tm{PPR sep.} & 25:17.3 & --- & 2.1 & 25:19.4 & 0.129 \\ 
\rowcolor{col_sep!10} \tm{PPR-fixed sep.} & 54.3 & --- & 3.2 & 57.5 & 0.130 \\ 
\rowcolor{col_sep!10} \tm{SVM sep.} & 3:42.3 & --- & 12.2 & 3:54.5 & 0.109 \\ 
\rowcolor{col_sep!10} \tm{KNN sep.} & 3:37.1 & --- & 21.3 & 3:58.4 & 0.121 \\ 
\rowcolor{col_sep!10} \tm{Tree sep.} & 48.7 & --- & 2.7 & 51.4 & 0.132 \\ 
\rowcolor{col_sep!10} \tm{RF sep.} & 9:20:25.2 & --- & 47.7 & 9:21:12.9 & \B0.071 \\ 
\rowcolor{col_sep!10} \tm{RF-def sep.} & 4:00.4 & --- & 23.3 & 4:23.7 & 0.072 \\ 
\rowcolor{col_sep!10} \tm{XGBoost-def sep.} & 4:32.5 & --- & 3.2 & 4:35.7 & 0.103 \\ 
\rowcolor{col_sep!10} \tm{CatBoost sep.} & 1:41:31.6 & --- & 1.1 & 1:41:32.7 & 0.082 \\ 
\rowcolor{col_sur!10} \tm{LM sur.} & 22.5 & --- & 3.4 & 25.9 & 0.162 \\ 
\rowcolor{col_sur!10} \tm{Poly-2 sur.} & 39.0 & --- & 4.7 & 43.7 & 0.155 \\ 
\rowcolor{col_sur!10} \tm{Poly-3 sur.} & 43.7 & --- & 6.2 & 49.9 & 0.147 \\ 
\rowcolor{col_sur!10} \tm{Lasso sur.} & 1:11.2 & --- & 0.4 & 1:11.6 & 0.167 \\ 
\rowcolor{col_sur!10} \tm{Ridge sur.} & 1:30.2 & --- & 1.0 & 1:31.2 & 0.182 \\ 
\rowcolor{col_sur!10} \tm{Elastic sur.} & 1:23.7 & --- & 1.3 & 1:25.0 & 0.166 \\ 
\rowcolor{col_sur!10} \tm{GAM sur.} & 3:38.5 & --- & 22.9 & 4:01.4 & 0.145 \\ 
\rowcolor{col_sur!10} \tm{GAM-5 sur.} & 11:29.9 & --- & 10.3 & 11:40.2 & 0.144 \\ 
\rowcolor{col_sur!10} \tm{GAM-10 sur.} & 11:15.0 & --- & 13.8 & 11:28.8 & 0.142 \\ 
\rowcolor{col_sur!10} \tm{PCR sur.} & 4:37.8 & --- & 4.3 & 4:42.1 & 0.162 \\ 
\rowcolor{col_sur!10} \tm{PLS sur.} & 4:34.8 & --- & 5.7 & 4:40.5 & 0.162 \\ 
\rowcolor{col_sur!10} \tm{PPR sur.} & 1:20:09.6 & --- & 5.9 & 1:20:15.5 & 0.149 \\ 
\rowcolor{col_sur!10} \tm{KNN sur.} & 31.0 & --- & 0.2 & 31.2 & 0.369 \\ 
\rowcolor{col_sur!10} \tm{Tree sur.} & 3:40.8 & --- & 3.7 & 3:44.5 & 0.214 \\ 
\rowcolor{col_sur!10} \tm{RF sur.} & 1:15:41:45.0 & --- & 1:01.3 & 1:15:42:46.3 & 0.085 \\ 
\rowcolor{col_sur!10} \tm{RF-def sur.} & 27:59.0 & --- & 42.0 & 28:41.0 & 0.075 \\ 
\rowcolor{col_sur!10} \tm{XGBoost-def sur.} & 15:53.3 & --- & 3.6 & 15:56.9 & 0.113 \\ 
\rowcolor{col_sur!10} \tm{CatBoost sur.} & 29:01.9 & --- & 4.5 & 29:06.4 & 0.108 \\ 
\rowcolor{col_sur!10} \tm{NN-Frye-3000 sur.} & 1:43:10.9 & --- & 6.3 & 1:43:17.2 & 0.227 \\ 
\rowcolor{col_sur!10} \tm{NN-Frye-6000 sur.} & 3:22:15.2 & --- & 6.2 & 3:22:21.4 & 0.210 \\ 
\rowcolor{col_sur!10} \tm{NN-Frye-15000 sur.} & 17:35:44.9 & --- & 8.2 & 17:35:53.1 & 0.190 \\ 
\rowcolor{col_sur!10} \tm{NN-Frye-40000 sur.} & 1:10:53:20.6 & --- & 6.8 & 1:10:53:27.4 & 0.170 \\ 
\rowcolor{col_sur!10} \tm{NN-Frye-ES sur.} & 1:39:33.7 & --- & 8.6 & 1:39:42.3 & 0.272 \\ 
\rowcolor{col_sur!10} \tm{NN-Olsen-500 sur.} & 35:04.3 & --- & 5.1 & 35:09.4 & 0.145 \\ 
\rowcolor{col_sur!10} \tm{NN-Olsen-2500 sur.} & 2:55:22.9 & --- & 5.4 & 2:55:28.3 & 0.137 \\ 
\rowcolor{col_sur!10} \tm{NN-Olsen-10000 sur.} & 23:56:39.8 & --- & 7.2 & 23:56:47.0 & 0.130 \\ 
\rowcolor{col_sur!10} \tm{NN-Olsen-20000 sur.} & 1:23:54:17.3 & --- & 6.2 & 1:23:54:23.5 & 0.132 \\  
\rowcolor{col_sur!10} \tm{NN-Olsen-ES sur.} & 2:42:45.9 & --- & 9.0 & 2:42:54.9 & 0.116 \\ 
\bottomrule
\end{tabular}
\end{adjustbox}
\caption{{\small The (Red) Wine data set experiment: $M = 11$, $N_\text{train} = 1349$, and $N_\text{test} = 250$.
}}
\label{tab:SimStudyTimeWine}
\end{table}

\begin{table}[ht]
\centering
\vspace{-8ex}
\begin{adjustbox}{max width=1\textwidth, max totalheight=21.5cm}
\begin{tabular}{lrrrrrr}
\toprule
Method & Training & Generating $\x_\s^{(k)}$ & Predicting $v(\s)$ & Total CPU Time & $\operatorname{MSE}_v$\\ 
\midrule
\rowcolor{col_ind!10} \tm{Independence*}       &           0.0 &      34:31.9 &   42:14.4 &     1:16:46.3 & 0.041 \\ 
\rowcolor{col_gen!10} \tm{VAEAC-200}           &  1:05:55:35.9 & 4:05:53:10.4 &   42:17.4 &  5:12:31:03.7 & 0.027 \\ 
\rowcolor{col_gen!10} \tm{VAEAC-1000}          &  6:01:10:13.0 & 3:23:34:03.1 &   51:35.8 & 10:01:35:51.9 & 0.027 \\ 
\rowcolor{col_gen!10} \tm{VAEAC-f-indir-200}   &  1:10:59:08.6 & 5:19:33:00.1 & 1:01:06.4 &  7:07:33:15.1 & 0.027 \\ 
\rowcolor{col_gen!10} \tm{VAEAC-f-indir-1000}  &  6:17:32:29.9 & 4:22:48:45.9 &   53:42.3 & 11:17:14:58.1 & 0.027 \\ 
\rowcolor{col_gen!10} \tm{VAEAC-f-dir-200}     &  1:10:59:08.6 & 5:19:33:00.1 &       0.2 &  7:06:32:08.9 & 0.033 \\ 
\rowcolor{col_gen!10} \tm{VAEAC-f-dir-1000}    &  6:17:32:29.9 & 4:22:48:45.9 &       0.3 & 11:16:21:16.1 & 0.032 \\ 
\rowcolor{col_sep!10} \tm{LM sep.}             &     8:38:57.2 &          --- &   21:16.3 &     9:00:13.5 & 0.043 \\
\rowcolor{col_sep!10} \tm{Poly-2 sep.}         &    13:44:03.3 &          --- &   27:29.3 &    14:11:32.6 & 0.037 \\ 
\rowcolor{col_sep!10} \tm{Poly-3 sep.}         &    17:00:32.0 &          --- &   29:08.1 &    17:29:40.1 & 0.037 \\ 
\rowcolor{col_sep!10} \tm{Poly-4 sep.}         &    19:25:17.4 &          --- &   28:26.1 &    19:53:43.5 & 0.036 \\ 
\rowcolor{col_sep!10} \tm{Poly-inter-2 sep.}   &    11:53:01.3 &          --- &   29:00.7 &    12:22:02.0 & 0.036 \\ 
\rowcolor{col_sep!10} \tm{GAM sep.}            &     2:37:14.8 &          --- &    1:37.8 &     2:38:52.6 & 0.033 \\
\rowcolor{col_sep!10} \tm{GAM-5 sep.}          &  7:19:07:43.1 &          --- & 2:12:59.1 &  7:21:20:42.2 & 0.034 \\ 
\rowcolor{col_sep!10} \tm{GAM-10 sep.}         & 75:22:18:27.1 &          --- & 2:05:52.8 & 76:00:24:19.9 & 0.033 \\ 
\rowcolor{col_sep!10} \tm{PLS sep.}            & 25:04:43:30.2 &          --- &   17:21.1 & 25:05:00:51.3 & 0.046 \\ 
\rowcolor{col_sep!10} \tm{PPR sep.}            & 14:12:16:08.6 &          --- &    3:39.0 & 14:12:19:47.6 & 0.032 \\
\rowcolor{col_sep!10} \tm{PPR-fixed sep.}      &  1:15:44:05.5 &          --- &    3:41.7 &  1:15:47:47.2 & 0.032 \\ 
\rowcolor{col_sep!10} \tm{Tree sep.}           &     5:44:54.7 &          --- &      46.0 &     5:45:40.7 & 0.031 \\ 
\rowcolor{col_sep!10} \tm{RF sep.}             & 98:13:17:15.8 &          --- &   16:11.6 & 98:13:33:27.4 & 0.027 \\ 
\rowcolor{col_sep!10} \tm{RF-def sep.}         &    12:49:17.0 &          --- &    8:39.0 &    12:57:56.0 & 0.028 \\ 
\rowcolor{col_sep!10} \tm{CatBoost sep.}       & 35:09:59:20.8 &          --- &      38.3 & 35:09:59:59.1 &\B0.026\\ 
\rowcolor{col_sur!10} \tm{NN-Frye-3000 sur.}   &  3:16:10:45.4 &          --- &    3:04.9 &  3:16:13:50.3 & 0.085 \\ 
\rowcolor{col_sur!10} \tm{NN-Frye-15000 sur.}  & 81:00:01:50.5 &          --- &    3:02.3 & 81:00:04:52.8 & 0.098 \\
\rowcolor{col_sur!10} \tm{NN-Frye-ES sur.}     &     9:42:32.0 &          --- &    2:24.8 &     9:44:56.8 & 0.101 \\ 
\rowcolor{col_sur!10} \tm{NN-Olsen-500 sur.}   &  3:11:31:47.6 &          --- &    2:10.3 &  3:11:33:57.9 & 0.045 \\ 
\rowcolor{col_sur!10} \tm{NN-Olsen-2500 sur.}  & 14:17:59:53.4 &          --- &    2:19.0 & 14:18:02:12.4 & 0.065 \\ 
\rowcolor{col_sur!10} \tm{NN-Olsen-10000 sur.} & 82:11:56:11.3 &          --- &    2:36.9 & 82:11:58:48.2 & 0.037 \\ 
\rowcolor{col_sur!10} \tm{NN-Olsen-ES sur.}    &  1:13:15:31.9 &          --- &    1:58.4 &  1:13:17:30.3 & 0.030 \\ 
\bottomrule
\end{tabular}
\end{adjustbox}
\caption{{\small The Adult data set experiment: $M = 14$, $N_\text{train} = 30\,000$, and $N_\text{test} = 162$. 
Creating the augmented test data on the form in \eqref{eq:surrogate_augmented_data} takes approximately two minutes and is part of the predicting time for the \tm{NN surrogate} approaches.
}}
\label{tab:SimStudyTimeAdult}
\end{table}

\clearpage
\newpage
\section{Schematic Overview of Conditional Shapley Values in XAI}
\label{Appendix:SchematicOverviewMethods}
\begin{figure}[!t]
    \centering
    \centerline{\includegraphics[width=1.2\textwidth]{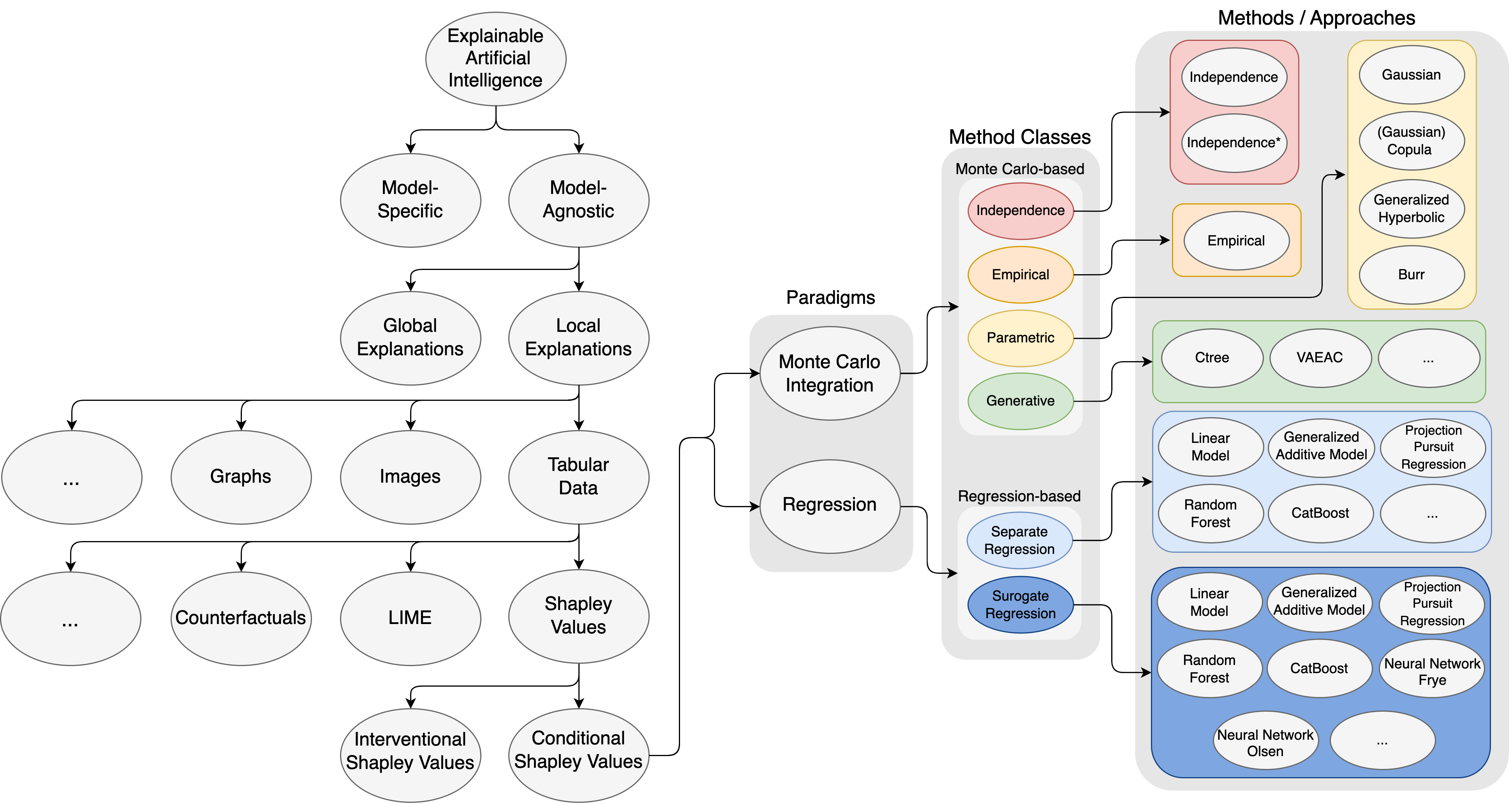}}
    \caption{{\small Schematic overview of conditional Shapley values in XAI. See \Cref{Appendix:SchematicOverviewMethods} for details. \vspace{-1em}}}
    \label{fig:SchematicOverviewMethods}
\end{figure}

\Cref{fig:SchematicOverviewMethods} provides a schematic overview of this article's method classes and methods for computing conditional Shapley value explanations. Furthermore, the figure also shows conditional Shapley values' place within the explainable artificial intelligence field as a model-agnostic explanation framework with local explanations. Note that the ellipses represent the additional methods introduced in \Cref{Appendix:Methods}. Furthermore, LIME is an explanation framework developed by \textcite{ribeiro2016should}, and see, e.g., \textcite{molnar2019}[Section 9.3] and \textcite{redelmeier2021mcce} for more on counterfactual explanations.

In this article, we used Shapley values to provide local explanations for models fitted to tabular data, but different Shapley value-based frameworks are developed for other settings. For example, \textcite{lundberg2020local} develop model-specific local and global Shapley value explanations for tree ensemble models. \textcite{covert2020understanding} introduce a model-agnostic global explanation framework based on Shapley values. \textcite{SurvShap} propose time-dependent Shapley value-based explanations for machine learning survival models. \textcite{bento2021timeshap} extend Shapley values to the sequential domain and introduce a model-agnostic Shapley value explanation framework for sequential decision-making models, such as recurrent neural networks. \textcite{chen2022explaining} explain a series of models by propagating Shapley values. \textcite{fastshap} use Shapley values to explain classifications made by image classifiers. \textcite{mastropietro2022edgeshaper,duval2021graphsvx} introduce new methodologies based on Shapley values to explain graph neural network models. \textcite{wang2021shapley} include knowledge about a causal graph between the features when creating Shapley value-based explanations. \textcite{heskes2020causal} also introduce a causal Shapley value methodology that exploits causal knowledge. 

\clearpage
\newpage
\printbibliography

\end{document}